# AI-driven ionic liquid discovery with unified chemical intelligence

Yuqi Yin[1#], Yibo Fu[2#], Siyuan Wang[3#], Lizhen Huo[4,5#], Peng Sun[1], Hongyu Wang[4], Xiaohui Wang[4], Chengli Pan[1], Zhiyong Li[2*], Zhirong Liu[1*], Jianji Wang[2*], and Zhaoxi Sun[4*]

*[1]College of Chemistry and Molecular Engineering, Peking University, Beijing 100871, China*

*[2]Key Laboratory of Green Chemical Media and Reactions, Ministry of Education, Henan Key Laboratory for Green Media and Reactions, Henan International Joint Laboratory of Green Chemistry, Collaborative Innovation Center of Henan Province for Green Manufacturing of Fine Chemicals, School of Chemistry and Chemical Engineering, Henan Normal University, Xinxiang, Henan 453007, P. R. China*

*[3]School of Computer Science, Shanghai Jiao Tong University, Shanghai 201100, China*

*[4]Faculty of Synthetic Biology, Shenzhen University of Advanced Technology, Shenzhen 518107, China*

*[5]Department of Biomedical Engineering, Southern University of Science and Technology, Shenzhen 518055, China*

*To whom correspondence should be addressed:

Zhiyong Li yli@htu.edu.cn

Zhirong Liu LiuZhiRong@pku.edu.cn

Jianji Wang jwang@htu.edu.cn

Zhaoxi Sun z.sun@suat-sz.edu.cn

## Abstract

The rational discovery of functional solvents remains challenging because macroscopic solvent behaviors emerge from complex many-body interactions across vast chemical spaces. Ionic Liquids (ILs) constitute one of the most chemically diverse solvent families, yet their exceptional structural tunability[1] remains only partially explored owing to the limitations of existing experimental[2,3] and computational[4-10] approaches. Here we introduce AIonopedia, a large-language-model-orchestrated agentic framework for IL research. Built around a domain-specific multimodal foundation model, it integrates information retrieval, chemical reasoning and autonomous tool execution within a unified architecture. Pretrained on large-scale unlabeled corpora and fine-tuned using over 80,000 curated measurements, AIonopedia learns expressive joint representations of multicomponent ionic systems, enabling accurate prediction across diverse physicochemical properties. Through a series of literature-grounded evaluations, AIonopedia exhibits robust out-of-distribution generalization, while many-body-expansion-based analysis provides physically grounded explanations of structure-property relationships. By coupling molecular screening and optimization with wet-lab validation in a closed-loop workflow, AIonopedia identifies high-performance ILs and reveals transferable design principles for the capture of various volatile organic compounds. This work establishes an end-to-end AI-driven paradigm for understanding and exploring complex ionic systems, demonstrating the promise of chemically specialized foundation models as a generalizable strategy for scientific discovery.

The rational discovery of functional solvents remains a central challenge in modern chemistry because their properties emerge from complex many-body interactions across vast chemical spaces. Ionic Liquids (ILs), one of the most chemically diverse solvent families, offer exceptional design flexibility through variation of cations and anions.[1] This versatility has enabled applications spanning catalysis, electrochemistry, biomass processing and separations.[11-14] However, this key advantage also complicates their discovery: even among synthetically accessible ions, possible cation-anion combinations exceed millions, rendering exhaustive experimental exploration impractical.[2,3] Consequently, most IL discovery efforts remain guided by empirical intuition. These heuristics typically involve selecting specific ion families to meet targeted property requirements or introducing customized modifications to the corresponding ion structures. Nevertheless, such knowledge-driven strategies explore only a minute fraction of chemical space and remain strongly constrained by existing prior knowledge.

Computational approaches have been developed to accelerate IL discovery, yet each faces intrinsic limitations. Physics-based simulations, including quantum chemistry and all-atom Molecular Dynamics (MD), provide general frameworks for property prediction.[4,5] However, their computational cost becomes prohibitive for large-scale screening, particularly for multicomponent ion-solute systems. Semi-empirical approaches (e.g., the Abraham model) reduce resource demands but often compromise predictive accuracy.[6,7] Moreover, both physics-based and semi-empirical approaches rely heavily on method-specific approximations or system-specific parameterizations, limiting their transferability across diverse IL systems. In recent years, machine-learning models have demonstrated encouraging performance for predicting individual IL properties.[15-17] Nevertheless, systematic IL discovery remains hindered by three major challenges: the scarcity of reliable experimental data[8], inadequate representations of IL chemistry[9], and the lack of a unified full-cycle pipeline integrating retrieval, prediction, and optimization[10].

Recent advances in foundation models offer a promising route to addressing these issues. Self-supervised pretraining on vast collections of unlabeled data has enabled various molecular foundation models to learn transferable chemical knowledge.[18,19] Building on this, multimodal alignment across molecular sequences, graphs, and 3D structures yields multi-view representations capturing complementary structural information.[20,21] Furthermore, coupling Large Language Models (LLMs) with external tools enables autonomous reasoning and workflow execution, markedly reducing manual research effort.[22,23] Despite their considerable potential, such capabilities remain largely unexplored in functional-solvent discovery, leaving opportunities for paradigm innovation.

Here we introduce AIonopedia, an LLM-orchestrated agentic framework tailored to IL research, built around a domain-specific multimodal foundation model (Fig. 1A). AIonopedia is developed using one of the largest manually curated IL datasets reported to date, comprising >80,000 experimentally validated measurements covering bulk physicochemical properties and solute-solvent thermodynamic quantities. To better represent IL chemistry, the core model aligns and fuses sequence and graph representations through multimodal pretraining on large-scale unlabeled corpora, with physicochemical descriptors providing auxiliary supervision, followed by task-specific fine-tuning for property prediction. Its agentic layer further orchestrates specialized tools from literature retrieval to candidate searching, enabling an end-to-end workflow for IL discovery.

AIonopedia consistently outperforms existing approaches in diverse property-prediction tasks, supports robust Out-Of-Distribution (OOD) generalization across heterogeneous datasets, and extends effectively to the interpretation of literature-reported phenomena across various scenarios. Beyond prediction, the framework provides physically informed explanations of model outputs, enables autonomous discovery of previously unexplored ILs with experimentally validated State-Of-The-Art (SOTA) performance, and uncovers transferable solvent-design principles corroborated by intermolecular interaction analyses. Collectively, these results demonstrate how data-driven foundation models can deepen physicochemical insight into ionic-solvent behavior while accelerating the discovery of new ionic solvents.

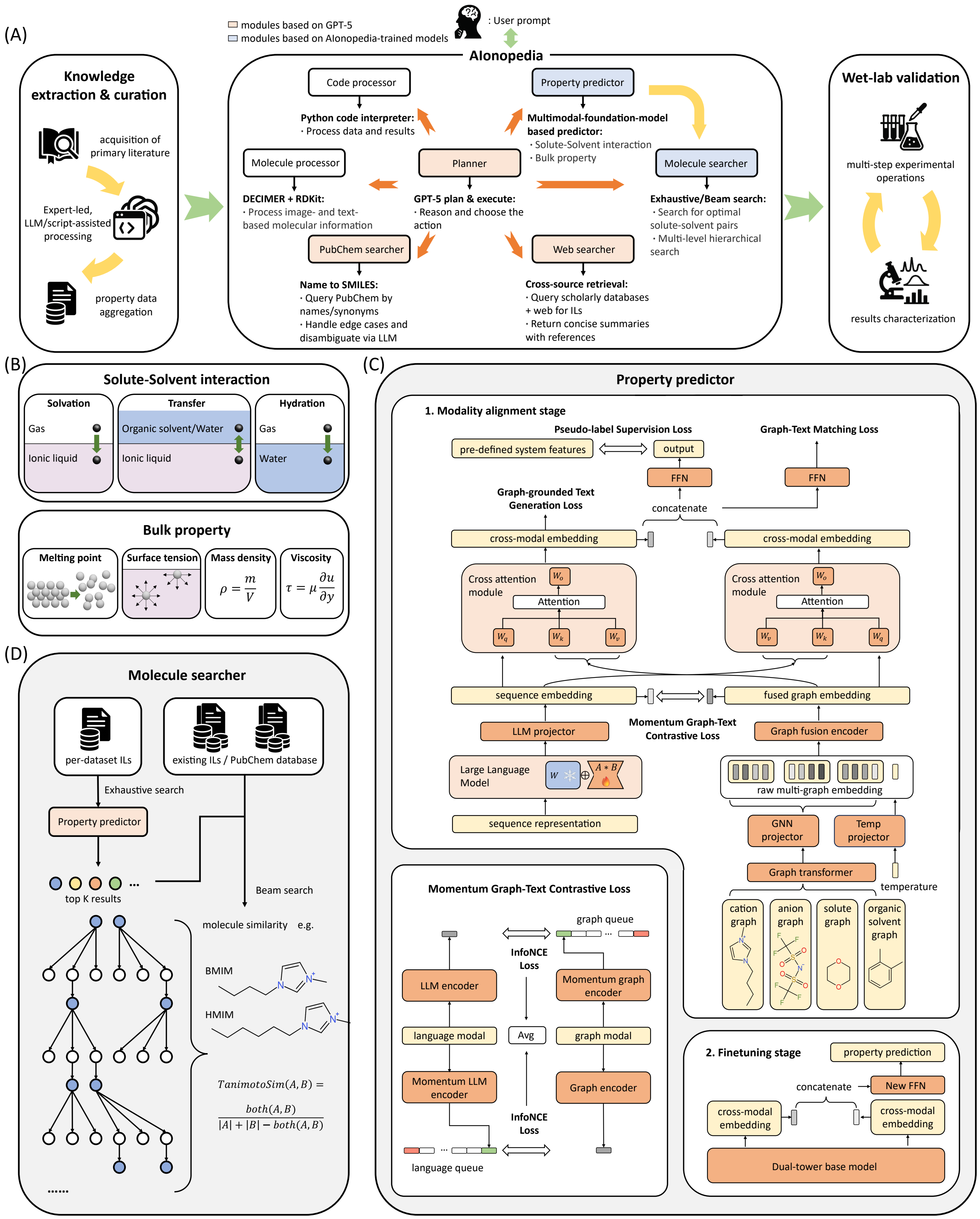

**Fig. 1. Overview of AIonopedia, an LLM-orchestrated agentic framework for IL discovery, built around a domain-specific multimodal foundation model.** (A) Schematic of the AIonopedia-centric closed-loop IL discovery pipeline, in which the framework bridges upstream knowledge extraction and curation with

downstream wet-lab validation. Within AIonopedia, an LLM-based planner coordinates specialized modules for data processing, information retrieval, property prediction, and molecule screening. (B) Representative application scenarios supported by the curated IL-specific knowledge base, including solute-solvent interaction tasks (solvation, biphasic transfer and hydration) and bulk property prediction tasks (melting point, surface tension, mass density, and viscosity). (C) Architecture of the multimodal property predictor. Embeddings from the molecular-language and graph branches are aligned before bidirectional cross-modal fusion, with the model jointly optimized under four multimodal pretraining objectives. The model is then fine-tuned in the same dual-tower framework under task-specific supervision to enable accurate prediction across various IL properties. (D) Architecture of the molecule searcher, which combines exhaustive database screening with hierarchical beam search guided by the property predictor and molecular similarity (Tanimoto score) to identify promising candidates satisfying user-defined criteria.

**Results.**

**Building an AI foundation for IL discovery.**

To establish the data foundation for AIonopedia, we curated >80,000 experimentally validated measurements spanning solute-solvent thermodynamics and key bulk IL properties across diverse temperatures, molecular compositions and literature sources (Fig. 1B). Notably, the solute-solvent interaction dataset includes ~30,000 free energy entries, which represents the largest data source to the best of our knowledge. We also included organic solvent-water transfer data to enable predictions of IL-organic transfer free energies through thermodynamic cycles. Meanwhile, the bulk-property datasets cover ~6,000 distinct ILs, approximately twice the number of IL species covered in the widely used ILThermo database.[24] To further exploit abundant unlabeled molecular information, we generated 2.8 million IL-centric synthetic systems for modality-alignment pretraining through similarity-guided molecular expansion and combinatorial sampling (Section S1 and Fig. S1).

At the core of AIonopedia, we developed a dual-tower multimodal foundation model that jointly represents complete IL systems through molecular language and graph modalities (Fig. 1C). Unlike conventional models that encode each molecule independently, the model treats the entire multicomponent system (e.g., 1-butanol in $[BMPyrr]^{+}[B(CN)_4]^{-}$ at 298.15 K) as a unified learning object, preserving the intermolecular and thermodynamic context that governs chemical behavior. Using the synthetic corpus, the two modalities are first aligned in a shared latent space through a momentum contrastive objective[25] and subsequently fused via bidirectional cross-modal attention. Multimodal pretraining further combines two complementary cross-modal objectives[26] with descriptor-based auxiliary supervision to preserve structural consistency and physicochemical information, followed by task-specific fine-tuning on experimental datasets (Sections S2-S3).

Beyond predictive modeling, we embedded the foundation model within an agentic framework that translates complex research queries into executable multistep workflows. A central GPT-5-based planner[27] iteratively reasons over intermediate results, executes appropriate external tools, and incorporates their outputs into subsequent decisions until task completion[28]. This architecture grounds its reasoning in verified molecular identities and literature evidence before prediction, while allowing predicted properties to guide candidate screening and refinement (Fig. 1D). Details of the agent architecture and tool interfaces are provided in Section S4. Together with the data resources and core model, this agentic framework completes the versatile AIonopedia system that underpins the downstream analyses and experimental discovery presented below.

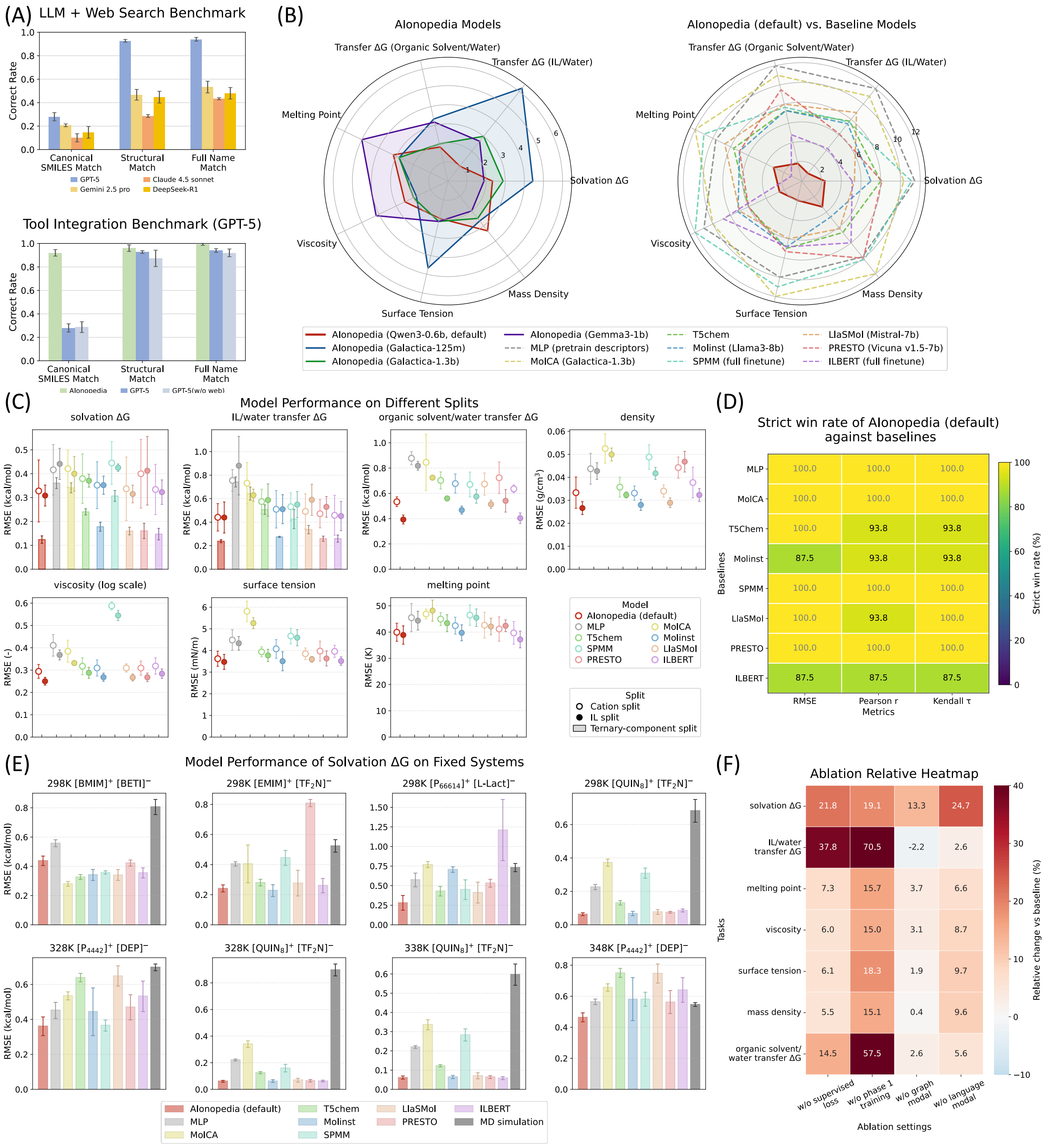

**Fig. 2. Accurate chemical grounding and property prediction with AIonopedia.** (A) Accuracy of LLM-based grounding of IL-forming species. Top, performance of four web-augmented LLMs in interpreting IL abbreviations under three increasingly lenient criteria: exact canonical-SMILES recovery, structure-level SMILES equivalence, and correct full-name identification. Bottom, comparison of GPT-5-based frameworks with different levels of tool integration. (B) Model performance comparison based on rankings averaged across metrics and split schemes. Left, average performance ranks of AIonopedia variants using different LLM

backbones. Right, comparison of the default AIonopedia model (Qwen3-0.6b-based[29]) with deep learning baselines. The original RMSE (↓), Pearson $r$ (↑), and Kendall $\tau$ (↑) values used to derive these rankings are provided in Tables S9-S11. (C) Generalization performance under single-component (cation or organic solvent), pairwise (ion pair or organic solvent-solute pair), and ternary-component (solute-cation-anion triplet) splits. (D) Strict win rates of the default AIonopedia model against baseline methods. Performance differences below predefined thresholds (baseline-relative RMSE < 1%, Pearson $r$ < 0.001, and Kendall $\tau$ < 0.005) were treated as ties. (E) Performance on system-specific IL solvation free energy datasets, with the numerical metrics provided in Table S13. Across all test sets, AIonopedia achieved RMSEs well below the chemical-accuracy threshold of 1 kcal $mol^{-1}$. (F) Ablation analysis of the smallest AIonopedia variant (Galactica-125m-based[30]). Values indicate the relative RMSE increase after removing individual training objectives or architectural components.

**Chemical grounding enables reliable agent reasoning.**

A fundamental challenge for IL informatics is the inconsistent representation of ionic species across the literature. The same ion is frequently described using multiple abbreviations, synonyms or non-standard nomenclature, making reliable chemical grounding difficult for conventional pipelines (e.g., N-octyltrimethylammonium may be abbreviated as both $[C_8TA]^+$ and $[N_{1118}]^+$).[31] Such ambiguity can propagate into downstream property prediction and mechanistic analysis. We therefore benchmarked the ion-recognition capability of AIonopedia using a challenging ion-abbreviation interpretation dataset comprising 50 commonly used abbreviations (Tables S7-S8). In Fig. 2A, GPT-5 achieved the highest accuracy among LLMs and was adopted as the planner of AIonopedia, while the complete agentic framework further increased accuracy to 92.0% under the strictest evaluation criterion through coordination of domain-specific modules.

**Accurate property prediction with AIonopedia.**

Accurate property prediction is the prerequisite for rational IL discovery. We therefore benchmarked AIonopedia on bulk properties and ternary solute-solvent interactions, using multi-level 5-fold cross-validation schemes designed to assess OOD generalization (Section S6). The cation- and IL-based splits were used for the bulk properties, while an additional split based on the cation-anion-solute ternary combination was included for the solute-solvent interaction datasets. For organic-solvent/water transfer free energies, analogous solvent-based and solvent-solute-pair-based splits were adopted. These protocols assess model applicability to unseen molecular components and increasingly complex component combinations.

Hyperparameters were tuned on the strictest cation-based split and fixed across all remaining protocols to avoid overfitting (Table S1).

AIonopedia variants initialized with different LLMs were compared with representative chemistry-domain baselines, including encoder- and decoder-based architectures as well as unimodal and multimodal approaches. Across datasets and evaluation schemes, AIonopedia consistently achieved the best overall performance (Fig. 2B-C), with the Qwen3-0.6b variant ranking first in 18 dataset-split-metric combinations (Tables S9-S11) and therefore being selected as the default model for subsequent analyses. Despite its LLM backbone, AIonopedia had a trainable parameter count comparable to those of fully fine-tuned encoder-based models and lower than those of several larger decoder-based baselines. The strict win rates in Fig. 2D summarize comparisons across all datasets and metrics, with AIonopedia maintaining win rates above 85% against every baseline.

Further, we evaluated AIonopedia on system-specific OOD test sets (Fig. S2) containing temperature-dependent measurements for chemically diverse ILs and compared it with physics-based MD simulations and other deep-learning models (Fig. 2E). Notably, although MD simulations (detailed in Section S5) provided chemically reasonable predictions with stable errors across systems (RMSE = 0.524~0.897 kcal $mol^{-1}$), their computational cost limits large-scale exploration of IL chemical space. AIonopedia achieved substantially lower errors on most systems (RMSE = 0.060~0.464 kcal $mol^{-1}$), including challenging OOD cases in which one or both ions were absent from the training set. The largest improvements were observed for phosphonium-based ILs, for which several baseline models exhibited pronounced performance degradation, while AIonopedia retained accurate predictions. Additional benchmarks were also performed for mass density and hydration free energy, where AIonopedia achieved the best or comparable performance among all approaches (Table S14). Taken together, these results demonstrate the robust generalization of AIonopedia to previously unseen ionic systems and its applicability across diverse IL-related thermodynamic properties.

To understand the contribution of different components within the multimodal framework, we performed ablation analyses using the smallest AIonopedia model (Fig. 2F). Removing individual modalities or the descriptor-supervised objective reduced prediction accuracy in nearly all cases, showing that topological, semantic, and physicochemical information contributes in complementary ways. Among the tested ablations, omitting the initial modality-alignment stage caused the largest performance degradation, increasing the RMSE on IL/water transfer free energy from 0.47 to 0.81 kcal $mol^{-1}$. Together with the descriptor-based MultiLayer Perceptron (MLP) baseline, these controls show that the gains of AIonopedia arise from integrated multimodal representation learning rather than descriptor supervision or either molecular modality alone (Table S15).

| reported | predicted | reported | predicted |
|---|---|---|---|

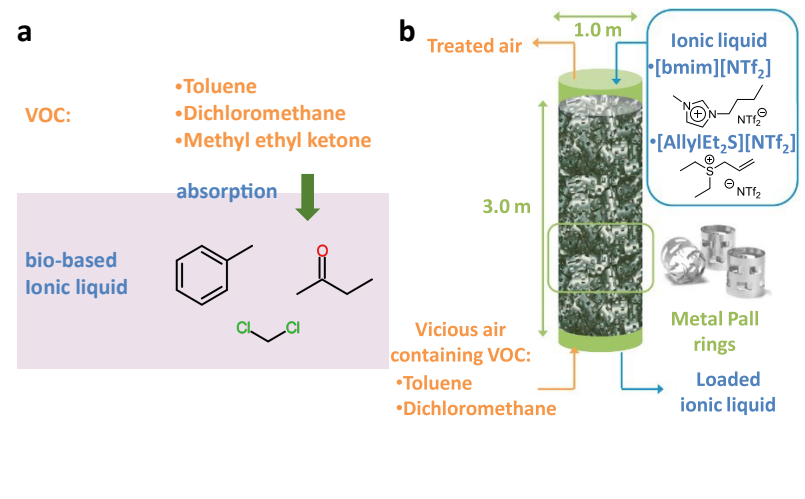

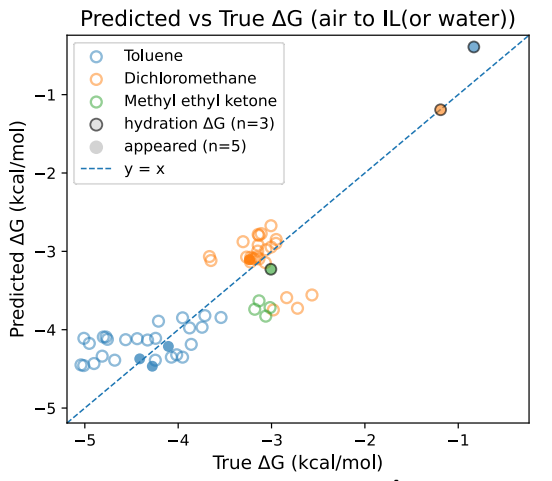

RMSE = 0.430 kcal/mol

Max |pred-true| = 1.004 kcal/mol

**(A)** Different Volatile Organic Compounds (VOCs) absorbed with (a) bio-based / (b) traditional hydrophobic ILs

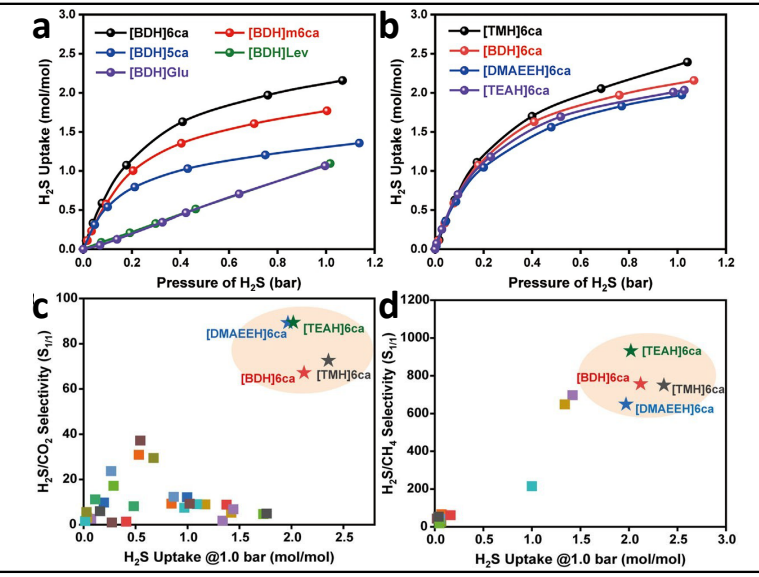

ΔG (solv, $H_2S$/[BDH]$^+$[6ca]$^-$) = **-3.664 kcal/mol**

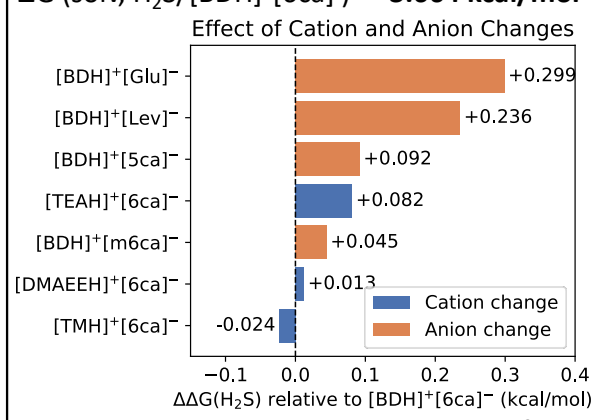

Mean |ΔG($H_2S$)-ΔG($CO_2$)| = **1.940 kcal/mol**

Mean |ΔG($H_2S$)-ΔG($CH_4$)| = **3.616 kcal/mol**

**$S_{1/1}(H_2S/CO_2) < S_{1/1}(H_2S/CH_4)$**

**(B)** (a,b) $H_2S$ absorption isotherms at 303.2 K in [BDH]$^+$-derived and [6ca]$^-$-derived ILs (c,d) Plot of $H_2S/CO_2$ and $H_2S/CH_4$ ideal selectivities ($S_{1/1}$) against H2S uptake at 1.0 bar in different absorbents

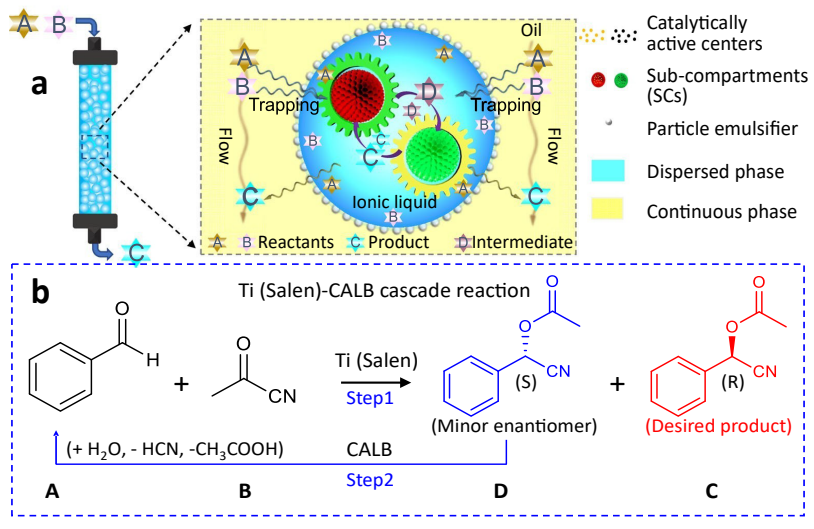

Benzaldehyde (A):
ΔG (IL to oil) = 0.277 kcal/mol

Acetyl cyanide (B):
ΔG (IL to oil) = 2.153 kcal/mol

Desired product (C) & Enantiomer (D):
ΔG (IL to oil) = 0.872 kcal/mol

**all ΔG > 0**

**RMSE = 0.702 kcal/mol**

**(C)** (a) Schematics for the construction of biomimetic microreactors for continuous flow cascade reactions (b) Cascade reaction for the synthesis of chiral O-acylated cyanohydrin by cascading Ti(Salen) and CALB

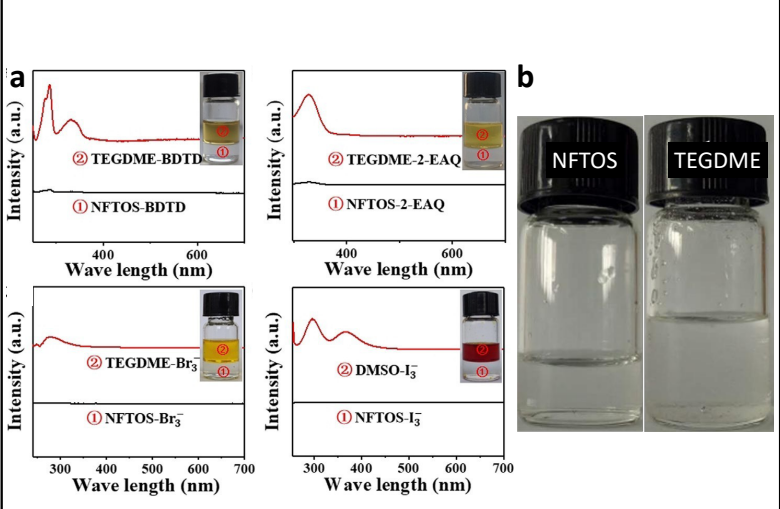

(a) BDTD: ΔG (TEGDME to NFTOS)
= 0.502 kcal/mol

2-EAQ: ΔG (TEGDME to NFTOS)
= 0.128 kcal/mol

[$Br_3$]$^-$: ΔG (TEGDME to NFTOS)
= 0.251 kcal/mol

[$I_3$]$^-$: ΔG (DMSO to NFTOS)
= 0.215 kcal/mol

(b) [Li]$^+$ [$TF_2N$]$^-$: ΔG (TEGDME to NFTOS)
= 0.936 kcal/mol

**all ΔG > 0**

**(D)** (a) The distribution of redox molecules in the TEGDME/DMSO and NFTOS immiscible electrolyte system (b) Images of NFTOS and TEGDME systems at maximum [Li]$^+$ [$TF_2N$]$^-$ solubilities (1.6 and 6.0 M, respectively)

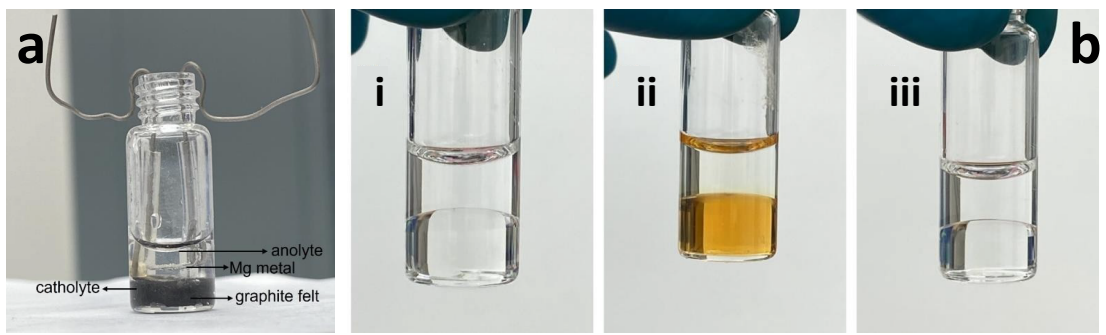

c

| | C3-PTZ | C8-PTZ | C18-PTZ | PEG3-PTZ |
|---|---|---|---|---|
| WSEIL (anolyte) | NS | NS | NS | PS |
| CH2Cl2 (catholyte) | 1.2 M | 0.8 M | 0.5 M | 1 M |

| PEG12-PTZ | PEG3-TTF | Tri-TEMPO | CP | TEMPO | PEG12-AQ | Fc |
|---|---|---|---|---|---|---|
| PS | PS | NS | NS | NS | PS | PS |
| 1 M | 1 M | 0.7 M | 0.6 M | 1.8 M | 0.5 M | 1.4 M |

NS: Not soluble, PS: Poor solubility (solubility < 0.1 M), S: Soluble (solubility > 1 M)

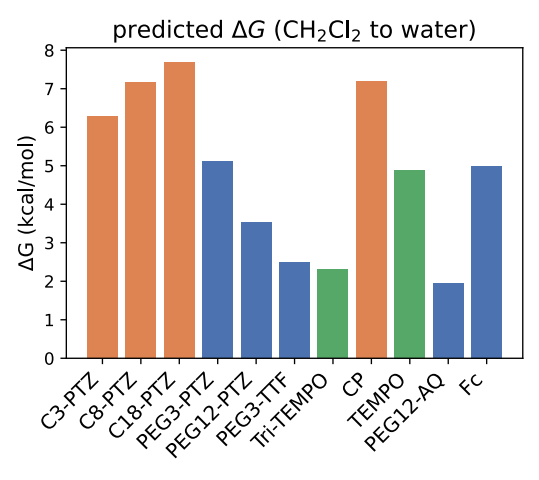

Mean ΔG ($CH_2Cl_2$ to water, PS)
= 3.612 kcal/mol

Mean ΔG ($CH_2Cl_2$ to water, NS, with radicals)
= 5.9230 kcal/mol

Mean ΔG ($CH_2Cl_2$ to water, NS, w/o radicals)
= 7.0863 kcal/mol

**Mean ΔG(PS) < Mean ΔG (NS)**

**(E)** (a) Image of Mg biphasic battery under static conditions (b) Images of (i) biphasic system with (ii) TEMPO and (iii) C3-PTZ catholytes (c) Screening of cathode materials for CH2Cl2 catholyte of the biphasic system

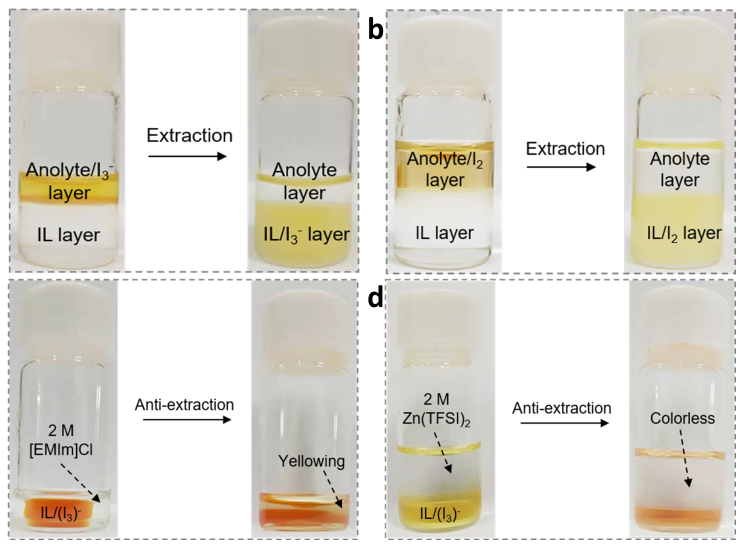

(a) [$I_3$]$^-$ : ΔG (IL to water)
= **1.577 kcal/mol > 0**

(b) [$I_2$] : ΔG (IL to water)
= **1.470 kcal/mol > 0**

(c) [$I_3$]$^-$ :
ΔG ([EMIM]$^+$[$TF_2N$]$^-$ to [EMIM]$^+$[Cl]$^-$ )
= **-0.498 kcal/mol < 0**

(d) [$I_3$]$^-$ :
ΔG ([EMIM]$^+$[$TF_2N$]$^-$ to [Zn]$^{+2}$[$TF_2N$]$^-_2$ )
= **0.556 kcal/mol > 0**

**(F)** Extraction effect study of IL toward (a) iodide and (b) iodine species, and anti-extraction effect study of (c) [EMIM]$^+$ cations and (d) [$TF_2N$]$^-$ anions for iodide species in IL/I2 mixture

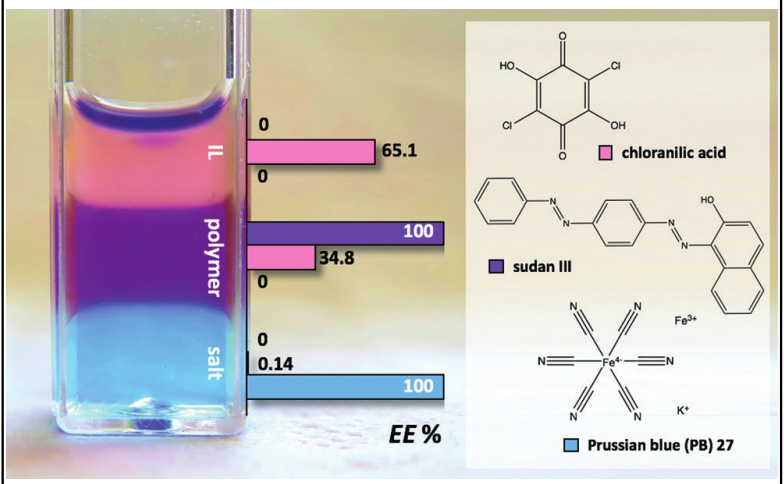

Chloranilic acid:
ΔG (IL to water) = **1.504 kcal/mol**
ΔG (IL to polymer) = **0.160 kcal/mol**

Sudan III:
ΔG (polymer to water) = **4.119 kcal/mol**
ΔG (polymer to IL) = **0.408 kcal/mol**

PB27: insoluble in water and decomposes in alkaline solution

**(G)** Dyes mixture separation in aqueous MultiPhase Systems (MuPSs) composed of [$N_{111(2OH)}$]$^+$[Ace]$^-$ based IL/PEG 600/$K_3PO_4$/$H_2O$

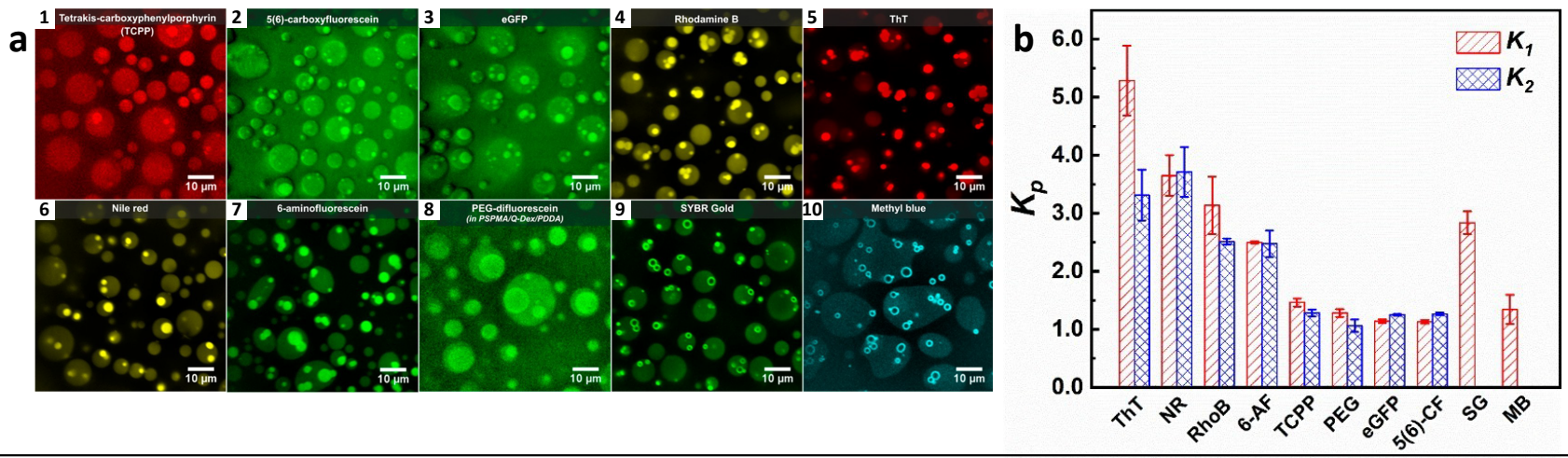

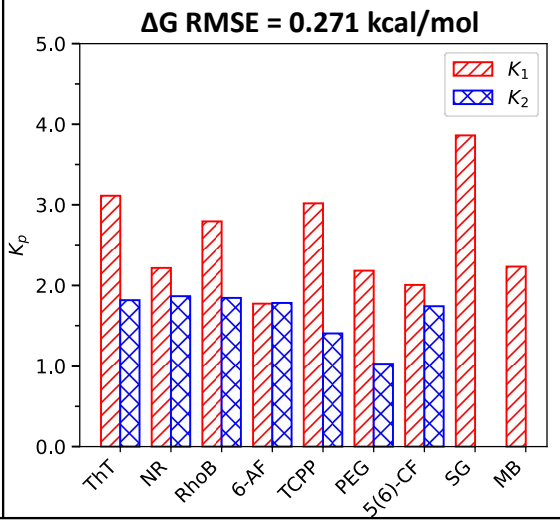

ThT, NR, RhoB, 6-AF, TCPP, 5(6)-CF:
**Mean K1 > Mean K2 > 1**
Prefer inner coacervate phase (PAH/PSPMA)

PEG-difluorescein:
**K1 > K2 ≳ 1**
Slightly prefer inner coacervate phase (PDDA/PSPMA)

eGFP: excluded (no direct molecular representation in current model)

SG/MB:
**Mean K1 > 1**, no K2 is provided

**(H)** (a) Partitioning of guest molecules in PSPMA/PAH(Q-Dex)/PDDA multiphase droplets visualized by confocal fluorescence microscopy (b) Partitioning coefficients of guest molecules in the outer coacervate phase (K1) and between the core coacervate and outer coacervate phase (K2)

**Fig. 3. Mechanism elucidation with AIonopedia.** (A-H) Case studies spanning increasingly challenging multiphase systems (i.e., from comparatively familiar to extremely OOD cases), selected from high-impact chemistry and materials journals.[32-40] For each example, the upper left panel shows the experimental observations reported in the original study, the upper right summarizes the thermodynamic quantities predicted by AIonopedia, and the lower panel provides a schematic description of the corresponding experimental behavior.

**Mechanistic interpretation across ILs and multiphase systems.**

Extrapolative performance beyond the benchmark splits was further evaluated using a selective literature panel spanning diverse topics and progressively greater OOD difficulty. Rather than serving as a comprehensive external benchmark, this panel was designed to bridge dataset-based evaluation and real wet-lab scenarios. The studies were manually curated under human-expert guidance from high-impact chemistry and materials science journals, with detailed selection criteria provided in Section S7. The final set comprised approximately 150 data points from 13 studies, all published from 2017 onward. Free-energy-related cases are presented in Fig. 3 in order of increasing OOD difficulty, with several non-IL multiphase systems included as complementary tests.

The first examples include two Volatile Organic Compound (VOC) absorption studies (Fig. 3A).[32,33] These species extend beyond the benchmark splits, with entire ion families absent from our training set. Nevertheless, AIonopedia-predicted solvation/hydration free energies remained consistent with benchmark performance, with essentially all errors within chemical accuracy. Another challenging case involved selective $H_2S/CO_2$ absorption (Fig. 3B), which represents a strongly OOD situation with $H_2S$ appearing only once in the experimental dataset and differing markedly from other sulfur-containing compounds in both physicochemical properties and chemical representation.[34] Despite the difference between saturated absorption capacity and infinite-dilution solvation free energy, AIonopedia-predicted free energy values largely reproduced the experimental absorption-capacity ranking across the controlled series, with $[TEAH]^+[6ca]^-$ as the only mismatch. This near-complete agreement in ranking highlights free-energy-based optimization as a reliable strategy for improving absorption capacity. Meanwhile, free energy differences within each system also supported the conclusion that $H_2S/CO_2$ separation is more difficult than $H_2S/CH_4$ separation.

Evaluation was further extended to more complex mixture settings. In Fig. 3C, AIonopedia accurately predicted IL/oil transfer free energies of reactants and products in biomimetic microreactors for continuous flow cascade reactions, with the largest error (1.058 kcal $mol^{-1}$) observed for two relatively complex chiral products.[35] Similar behavior was observed in several battery-related systems (Fig. 3D-F). A TEGDME(DMSO)/NFTOS biphasic system was used to test phase preference prediction for several complex electrolytes, including three ionic solutes.[36] Quantitative predictions remained reasonable for the solute $[Li]^+[TF_2N]^-$, whereas BDTD showed a larger deviation (1.740 kcal $mol^{-1}$). AIonopedia also distinguished insoluble from poorly soluble electrolytes in a $CH_2Cl_2$/water system.[37] Only TEMPO and Tri-TEMPO were misclassified, likely because they are ambient-stable radicals that lie far outside the training distribution. Finally, a biphasic $I_2/[I_3]^-$ extraction system was analyzed, in which we successfully identified $[EMIM]^+$ as the primary driver of extraction rather than the anion.[38]

The most challenging scenarios explored were those involving partitioning into polymer phases. For the dye distribution in an IL/polymer/salt triphasic system, AIonopedia correctly predicted the phase preferences of two dye classes using an oligomeric approximation (Fig. 3G).[39] PB27 was the only mismatch, likely because it is normally water-insoluble and decomposes under the alkaline experimental conditions, making its partitioning inconsistent with a standard dissolution process. In another extremely OOD example, we examined fluorophore partitioning in multiphase complex coacervate droplets (Fig. 3H).[40] Through the same approximation, AIonopedia successfully reproduced the overall phase preferences of nearly all dyes with an RMSE of 0.271 kcal $mol^{-1}$. The only exception was the protein eGFP, which lies outside the scope of our framework. AIonopedia also captured the inversion of core/outer phase identity between the PAH-based and Q-Dex-based systems, while preserving PEG-difluorescein preference for the inner phase. More broadly, AIonopedia identified the higher-order trend that dilute/outer partitioning is stronger than outer/core partitioning.

Additional tests of AIonopedia on literature-grounded bulk-property cases are provided in Section S8 and Fig. S3, with greater emphasis on quantitative agreement because these properties tend to relate to experimental phenomena more indirectly. Collectively, these examples suggest that AIonopedia retains meaningful predictive capability across various chemical environments, extending from conventional ILs to highly heterogeneous multiphase systems. Although several of the most complex examples necessarily rely on simplified representations of experimental systems, the ability to recover experimentally established trends across progressively more challenging OOD regimes suggests that AIonopedia captures generalizable physicochemical knowledge beyond the memorization of specific chemical combinations.

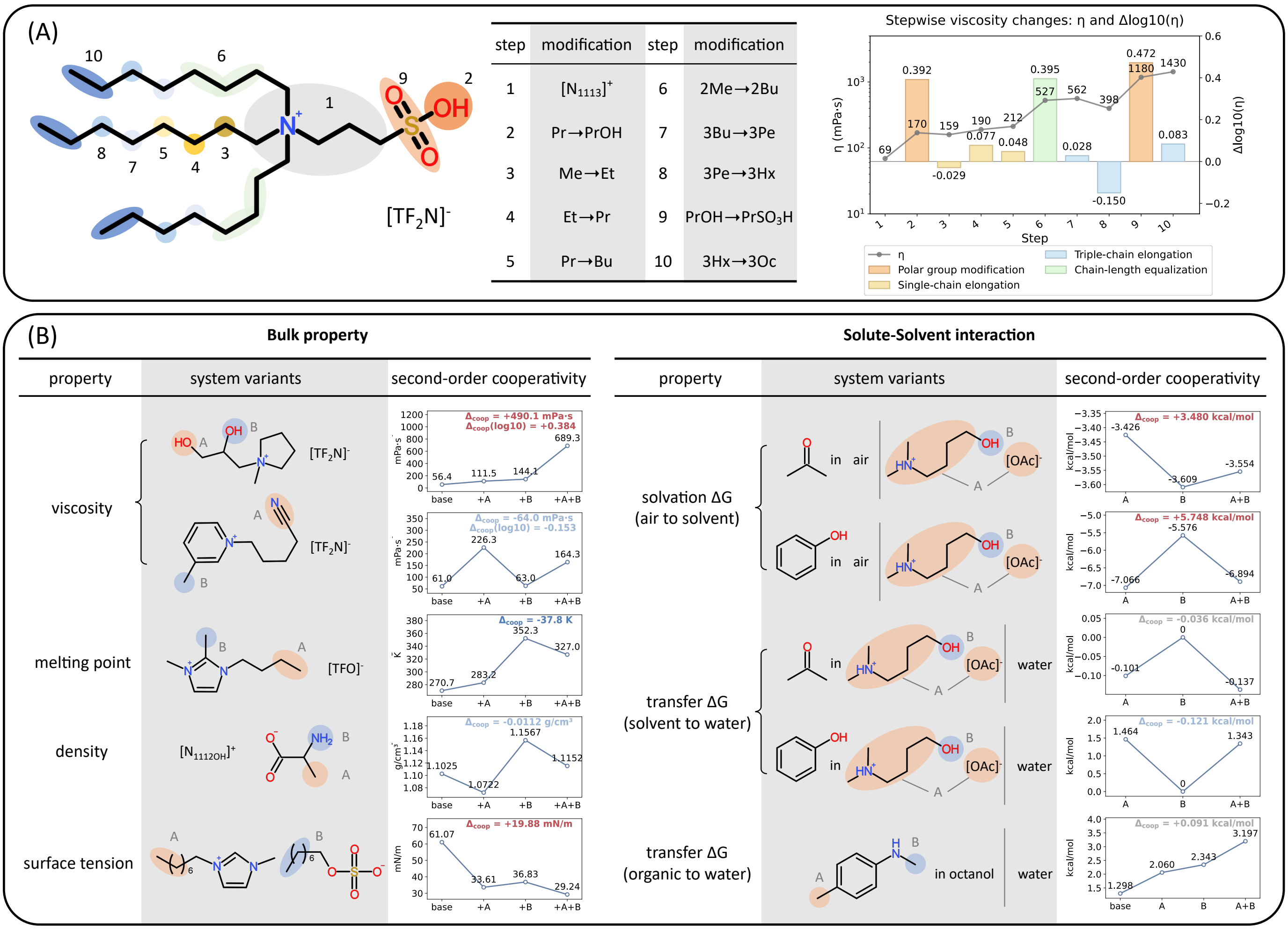

**Fig. 4. XAI characterization with AIonopedia.** (A) Sequential, knowledge-guided molecular perturbations trace how controlled structural modifications affect model predictions. Changes in viscosity following each modification are quantified by the predicted $\eta$ values (gray line) and the corresponding stepwise contributions to log-viscosity $\Delta\log10(\eta)$ (colored bars). (B) Higher-order interactions between molecular fragments were quantified using second-order finite differences, enabling the identification of synergistic, antagonistic, and additive effects across multiple prediction tasks. Red, blue, and gray denote positive, negative, and negligible interaction effects, respectively, while darker shades roughly indicate larger relative magnitudes. Literature sources supporting these qualitative assignments are provided in Section S9.

**Explainable AI through many-body expansion.**

To understand how the model derives its predictions beyond numerical accuracy, we introduced an explainable AI (XAI) framework inspired by many-body expansion, enabling physicochemically interpretable decomposition of molecular substructure contributions. Unlike conventional perturbation methods, which estimate feature importance by deleting fragments from complete molecules, our framework builds each target incrementally while maintaining chemically valid representations at each step. This strategy avoids the ambiguity inherent to deletion-based attribution (e.g., removing any carbon atom from the three alkyl chains of the cation shown in Fig. 4A produces identical responses) while naturally accounting for the evolving local environment during molecular assembly.

Taking viscosity as an illustrative example, first-order expansion quantified the contribution of individual structural modifications to the predicted property. The largest increases in $\Delta\log10(\eta)$ were associated with hydroxyl and sulfonic acid substitution, as well as equalization of the three side chains at butyl length, consistent with established experimental observations. More importantly, AIonopedia reproduced the subtle non-monotonic viscosity variations observed during both single- and triple-chain elongation, likely reflecting the competition among electrostatic interactions, hydrogen bonding, and van der Waals interactions that collectively govern IL dynamics.[41,42]

Building upon the first-order expansion, we further introduced a second-order many-body term to quantify cooperativity between fragment pairs (Fig. 4B), thereby identifying cases in which the combined effect of two structural perturbations deviates from simple additivity:

$$\Delta_{\text{coop}}O(A_p, B_q \mid S) = O(S + A_p + B_q) - O(S + B_q) - O(S + A_p) + O(S)$$

Here $S$ denotes the base chemical environment, and $A_p$ and $B_q$ denote the introduction of fragments $A$ and $B$ at positions $p$ and $q$, respectively. The full derivation of the framework is provided in Section S9.

Representative cooperative effects were consistently observed across all quantities predictable by the model. As a representative case of bulk properties, viscosity exhibited pronounced positive cooperativity under double hydroxyl substitution of the cation, whereas the meta carbon in pyridinium cations partially offset the increase induced by cyano substitution.

The same framework further revealed hidden physicochemical mechanisms governing transfer thermodynamics. Since free-energy-related quantities are not inherently positive, a positive $\Delta_{\text{coop}}$ here indicates a second-order effect that thermodynamically disfavors phase transfer. In the $[N_{114OH}]^+[OAc]^-$ IL system, hydroxyl substitution only weakly altered hydrogen-bond donor ability α but substantially reduced acceptor ability β.[43,44] This asymmetry reflects strong hydrogen-bond acceptance by $[OAc]^-$: interionic hydrogen bonding masks the hydroxyl-induced increase in donor character while weakening the exposed acceptor character of the anion. Although this effect was largely obscured by the dominant gas-to-IL solvation contribution, it became evident in the IL-to-water transfer process. Accordingly, hydroxyl substitution produced negligible cooperativity for acetone, a hydrogen-bond-accepting solute, because α remained nearly unchanged, whereas the reduced β favored transfer of phenol, a hydrogen-bond-donating solute, into water.

To explain numerical predictions in an explicit analytical form, we also derived an AIonopedia-informed symbolic correction for IL density (Section S10 and Fig. S4). Collectively, these analyses reveal context-dependent, chemically intuitive structure-property relationships beyond simple fragment-level attribution or statistical correlation, demonstrating that AIonopedia complements strong predictive performance with scientifically grounded explanations of its predictions.

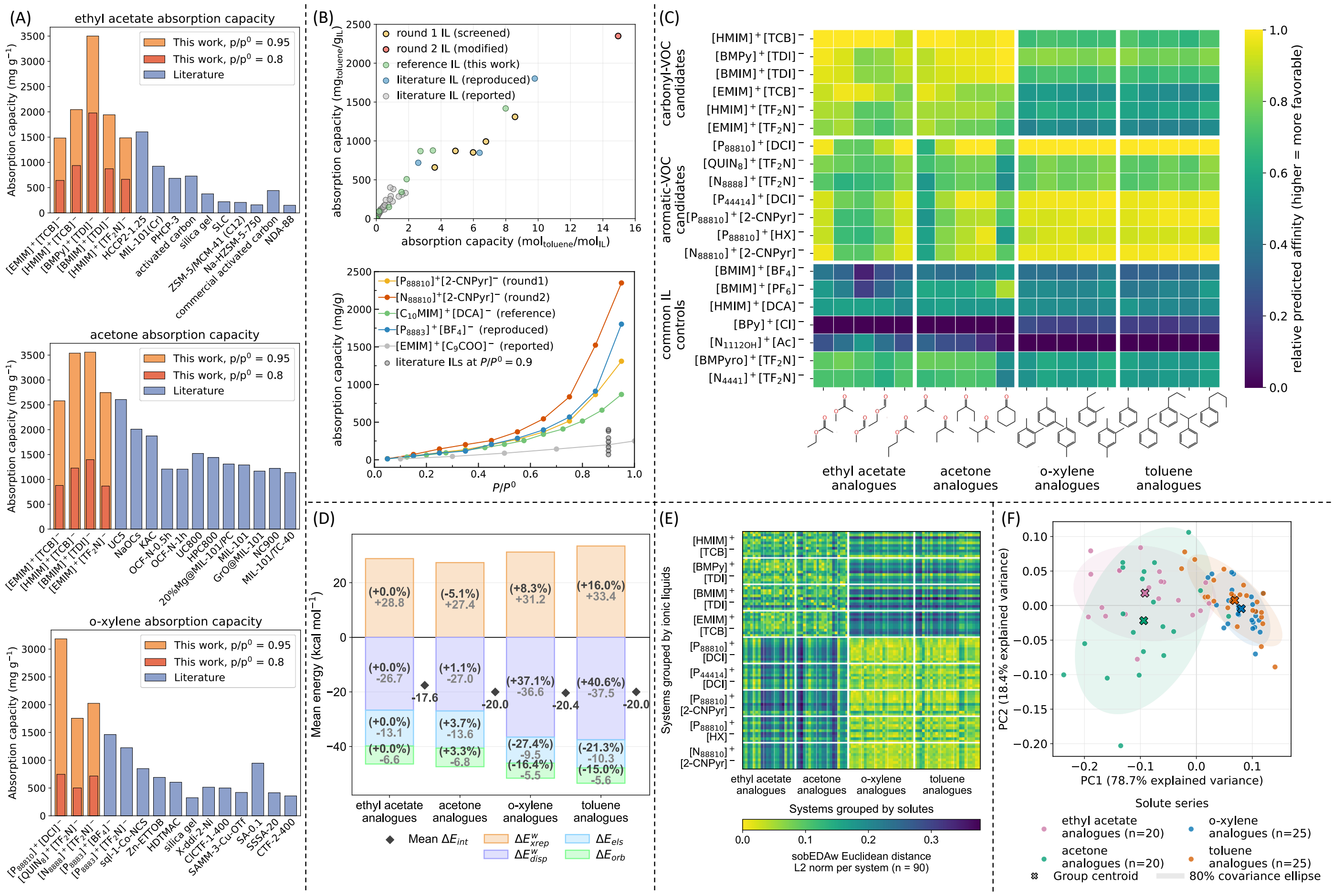

**Fig. 5. De novo design of ionic solvents for VOC capture with AIonopedia.** (A) Absorption capacities of ILs and other absorbent materials for three representative VOCs (ethyl acetate, acetone, and o-xylene). (B) Toluene absorption capacities expressed on mass- and mole-based scales, together with absorption isotherms as a function of partial pressure ($p/p^0$) for representative ILs and literature systems. (C) AIonopedia-predicted relative affinity map for prioritized IL candidates and common controls across target-centered VOC analogue series. Colors represent column-wise min-max-normalized $-\Delta G_{solvation}$ values, with higher values indicating more favorable solvation. (D) Mean EDA energy components for IL-VOC systems spanning four VOC

analogue series, with positive and negative contributions stacked separately. Diamonds show mean total interaction energies; percentages indicate component-magnitude changes relative to ethyl acetate analogues. Details of the 90 IL-VOC systems are provided in Section S12. (E) Pairwise Euclidean distances between per-system L2-normalized four-component EDA interaction fingerprints. Systems are grouped by ILs along rows and VOC analogue series along columns; lower values indicate more similar fingerprints. (F) Principal Component Analysis (PCA) projection of the same L2-normalized EDA interaction fingerprints, colored by VOC analogue series. Crosses and shaded regions indicate group centroids and 80% covariance ellipses, respectively.

**AIonopedia enables autonomous IL discovery.**

We applied AIonopedia in practical IL discovery settings using VOC absorption tasks as representative wet-lab scenarios (Fig. S5). Experimentally reported absorption capacities remain fragmented and heterogeneous due to inconsistent protocols and operating conditions, thereby hindering standardized data processing and systematic optimization. The agentic framework therefore prioritizes candidate ILs according to the solvation thermodynamics of VOCs. The discovery workflow combines two complementary search modes. When little application-specific knowledge is available, AIonopedia performs large-scale exploration using the multimodal foundation model together with similarity-guided beam search, enabling efficient navigation of the vast combinatorial ion-pair space. Conversely, when prior knowledge exists, the agent iteratively refines known ILs through reasoning-guided structural optimization, evaluating chemically meaningful design hypotheses against predicted thermodynamic performance.

We first validated the general screening strategy (i.e., no prior knowledge) using three representative VOCs, namely ethyl acetate, acetone, and o-xylene (Fig. 5A). The highest-ranked candidates were synthesized and experimentally evaluated, with their performance compared with that of representative literature-reported materials, including previously reported ILs, metal-organic frameworks (MOFs), carbon-based materials, and silica gels. Across all three systems, AIonopedia successfully identified ILs exhibiting SOTA absorption capacities (Tables S18-S20), demonstrating that free-energy-based screening provides an effective general strategy for practical IL discovery.

The prior-structure-guided mode was also validated in two additional cases. Following common IL design practice, the workflow structurally modifies cations, which are more commonly tailored, while treating the less diverse anions mainly through direct substitution.[45-48] For $CO_2$ absorption, starting from $[EMIM]^+[SCN]^-$ with the cation fixed, AIonopedia identified $[TCB]^-$ within ten iterations, with predicted solvation free energies matching the reported uptake trend with increasing cyano substitution. In another zero-shot $NH_3$-absorption case, starting from the nearly inactive $[P_{4442}]^+[DEP]^-$, AIonopedia proposed $[P_{4442COOH}]^+[DEP]^-$, which achieved a 23-fold increase in uptake to 74.1 mg $g^{-1}$. (Table S21, Fig. S6 and Section S13)

Next, we present toluene as an example illustrating the agentic feedback-guided discovery workflow combining both search modes. Toluene was selected as a stringent benchmark because numerous high-performance IL absorbents have previously been reported. The discovery process therefore proceeded in a closed-loop manner, where experimentally validated candidates from the first screening round served as starting points for subsequent reasoning-guided optimization. In Fig. 5B, AIonopedia first identified $[P_{88810}]^+$[2-CNPyr]$^-$ as one of the highest-performing candidates from de novo discovery. Guided by subsequent structural refinement, the agent further proposed $[N_{88810}]^+$[2-CNPyr]$^-$, which achieved a toluene uptake of 2350 mg $g^{-1}$ at 25 °C and a relative toluene vapor pressure of $p/p^0 = 0.95$ (Table S22), exceeding the strongest experimentally reproduced IL benchmark by approximately 30% (53% on a molar basis). Under saturated static absorption conditions used in the benchmark literature, the uptake further increased to 4137 mg $g^{-1}$ without reaching a clear plateau (Fig. S6), establishing a new SOTA IL absorbent for toluene capture.

Together, these results demonstrate that AIonopedia functions not merely as a property predictor but as an autonomous discovery platform that couples chemical reasoning with molecular screening and optimization, enabling closed-loop high-performance IL discovery when integrated with experimental validation.

**AIonopedia reveals transferable design principles.**

Beyond identifying individual high-performing ILs, we examined whether the solvent recommendations generated by AIonopedia reflect systematic preferences across related VOCs. To probe this class-level consistency, we extended the predictive evaluation from the experimentally validated targets to their surrounding chemical neighborhoods.

After removing duplicate candidates from independent screening runs, we compared the predicted solvation free energy landscape across the four VOC analogue series (Fig. 5C). The resulting matrix exhibited a pronounced block-diagonal structure, indicating that ILs prioritized for a given target tended to retain high relative predicted affinities both within individual analogue series and across chemically related series. For example, ILs containing [TCB]$^-$ consistently ranked highly across carbonyl-containing species, including ketones and esters, while showing weaker relative affinities toward aromatic hydrocarbons such as ethylbenzene and xylene isomers.

To investigate the physical basis of these solvent recommendations, we selected nine AIonopedia-prioritized ILs from Fig. 5C and paired them with VOCs from their preferred chemical families, yielding 90 IL-VOC systems. These systems were then analyzed using an Energy Decomposition Analysis (EDA)[49] method based on Density Functional Theory (DFT) (Sections S12). Despite broadly comparable mean total interaction energies, the four series exhibited distinct balances among the underlying energy components (Fig. 5D). Dispersion dominated attraction throughout but was particularly pronounced in aromatic systems and partly offset by greater exchange repulsion, whereas carbonyl systems showed stronger electrostatic and orbital interactions. These family-dependent shifts in the interaction-energy balance are consistent with established chemical expectations, providing chemically interpretable EDA interaction fingerprints for assessing AIonopedia-derived solvent recommendations.

At the individual-system level, we represented each system using a four-component EDA interaction fingerprint. For the comparisons in Figs. 5E and 5F, these fingerprints were L2-normalized to emphasize the relative balance among their components. Pairwise distances between these fingerprints revealed a broad separation between carbonyl- and aromatic-associated systems (Fig. 5E), qualitatively consistent with the family-level organization of the predicted affinity landscape in Fig. 5C. PCA provided a complementary view of this organization, with the first two principal components accounting for 97.1% of the variance (Fig. 5F). Aromatic-associated systems formed more compact groupings, whereas carbonyl-associated systems occupied broader, more diffuse regions.

Overall, AIonopedia uncovers class-specific yet transferable design principles supported by DFT-derived interaction energetics, establishing a framework with a clear physical basis for accelerating solvent discovery.

**Discussion.**

Traditional discovery of functional solvents relies heavily on costly iterative cycles of synthesis and experimental characterization, guided by existing chemical intuition. This limitation reflects a fundamental challenge in molecular science: how to move beyond isolated-molecule descriptions towards accurate, efficient and unified modelling of multicomponent systems whose properties emerge from collective interactions within a thermodynamic context. AIonopedia offers a potential avenue for pursuing this goal, demonstrating that such complex chemical spaces can be represented, interpreted and adaptively explored within a knowledge-grounded AI-driven framework.

A central implication of this work is that chemically specialized foundation models drive a shift toward a new paradigm for molecular research. Combining carefully curated, application-oriented experimental resources with large-scale unlabeled IL-system corpora forms the cornerstone of this data-centric methodology. With all molecular constituents jointly encoded through complementary modalities, AIonopedia achieves leading performance with robust OOD generalization across various datasets and extends effectively to literature-grounded scenarios and, more fundamentally, provides a universal computational scheme spanning heterogeneous observables and IL chemistries.

Importantly, predictive generality alone does not resolve the opacity of numeric outputs. Through an explainable approach based on many-body expansion, our framework reveals how molecular modifications and higher-order cooperative effects collectively give rise to bulk and solvent behaviors. In this way, AIonopedia provides a physicochemically motivated language for converting model predictions into experimentally testable hypotheses about key structure-property relationships.

In realistic wet-lab applications, AIonopedia reframes IL discovery from individual candidate evaluation to strategic and adaptive exploration. Through an LLM-orchestrated agent, AIonopedia unifies high-throughput exploration with prior-guided refinement. Ultimately, integration with experimental validation establishes a closed-loop workflow for identifying SOTA VOC absorbents and accumulating reusable knowledge for subsequent domain research. Furthermore, the correspondence between learned affinity landscapes and quantum-chemical interaction signatures bridges data-driven discovery with first-principles understanding, providing molecular-level guidance for solvent design beyond the originally validated targets.

The advances demonstrated through AIonopedia in the IL domain may facilitate progress in neighboring material systems governed by intricate coupled interactions, including electrolytes, supramolecular assemblies and functional soft materials. Still, its broader deployment remains constrained by the need for richer experimental coverage, expanded property spaces, and tighter integration with robotic wet-lab synthesis and characterization platforms. Addressing these limitations could transform future AI systems from predictive tools into self-evolving scientific collaborators capable of learning from experience, formulating hypotheses, and autonomously interrogating them, with AIonopedia representing an initial step towards this vision.

**Acknowledgements.**

We acknowledge the Polestar High-Performance Computing Platform at Peking University for providing computational resources used in part of this work.

**Funding.**

This work was supported by the National Natural Science Foundation of China (Grant Nos. 22273016, 22473003 and 22673072), the Natural Science Foundation of Henan Province (Grant No. 252300421043), the Plan for Henan Province University Science and Technology Innovation Team (Grant No. 25IRTSTHN002), and the Shanghai Municipal Science and Technology Commission (Grant No. 25DX2800500).

**CRediT Authorship Contribution Statement.**

Yuqi Yin: Conceptualization, Data curation, Formal analysis, Investigation, Methodology, Project administration, Resources, Software, Validation, Visualization, Writing - original draft, Writing - review & editing.

Yibo Fu: Data curation, Formal analysis, Investigation, Methodology, Resources, Software, Validation, Visualization, Writing - original draft, Writing - review & editing.

Siyuan Wang: Formal analysis, Investigation, Methodology, Resources, Software, Visualization, Writing - review & editing.

Lizhen Huo: Formal analysis, Investigation, Methodology, Resources, Software, Visualization, Writing - original draft, Writing - review & editing.

Peng Sun: Data curation, Formal analysis, Investigation, Methodology, Resources, Software, Validation, Visualization, Writing - review & editing.

Hongyu Wang: Data curation, Formal analysis, Investigation, Resources, Software, Validation, Visualization, Writing - review & editing.

Xiaohui Wang: Data curation, Investigation, Resources, Software, Visualization, Writing - review & editing.

Chengli Pan: Formal analysis, Investigation, Methodology, Validation, Writing - review & editing.

Zhiyong Li: Conceptualization, Data curation, Formal analysis, Funding acquisition, Investigation, Methodology, Project administration, Resources, Software, Supervision, Validation, Visualization, Writing - original draft, Writing - review & editing.

Zhirong Liu: Conceptualization, Data curation, Formal analysis, Funding acquisition, Investigation,

Methodology, Project administration, Resources, Software, Supervision, Validation, Visualization, Writing - review & editing.

Jianji Wang: Conceptualization, Data curation, Formal analysis, Funding acquisition, Investigation, Methodology, Project administration, Resources, Software, Supervision, Validation, Visualization, Writing - review & editing.

Zhaoxi Sun: Conceptualization, Data curation, Formal analysis, Funding acquisition, Investigation, Methodology, Project administration, Resources, Software, Supervision, Validation, Visualization, Writing - original draft, Writing - review & editing.

**Supporting Information Description.**

The Supplementary Information includes the construction of experimental datasets and multimodal pretraining corpora (Section S1); model architecture, training procedures and the agentic framework (Sections S2-S4); molecular simulation methods (Section S5); detailed chemical-grounding benchmarks, cross-validation benchmarks for property prediction, system-specific OOD and additional benchmarks, and ablation analyses (Section S6); selection criteria and details for literature case studies of biphasic transitions and bulk properties (Sections S7-S8); many-body expansion analysis and symbolic correction analysis (Sections S9-S10); wet-lab experimental procedures (Section S11); atomistic modelling of selective VOC absorption (Section S12); and examples of AIonopedia agent task execution and online deployment (Section S13).

**Code availability.**

The source code associated with AIonopedia, including data-processing, multimodal-pretraining, property-model fine-tuning, model-definition, and reference agent and molecular-search workflows, is publicly available at https://github.com/AIonopedia/AIonopedia-public. The pretrained AIonopedia model is available at https://huggingface.co/AIonopedia/AIonopedia. The property models used in this study can additionally be accessed for prediction through the freely available AIonopedia online service at https://ai4gs.suat-sz.edu.cn/zh (Section S13).

**Conflicts of interest.**

There are no conflicts of interest to declare.

# Supporting Information

## S1. Construction of experimental datasets and pretraining corpora.

To establish a comprehensive data foundation for AIonopedia, we curated large-scale resources covering both experimentally measured IL properties and synthetic systems. The experimental datasets were designed to support supervised learning of physicochemical properties, whereas the synthetic corpora were constructed to facilitate large-scale modality alignment between the sequence (molecular language) and graph representations.

Fig. S1A summarizes the size and chemical diversity of all datasets. To minimize the dataset bias, all datasets were curated with an emphasis on maximizing chemical diversity and avoiding excessive representation of any single molecular family. Although the organic solvent/water transfer free energy dataset does not contain IL species, it was included because it enables the derivation of IL/organic transfer free energies when combined with IL-water transfer free energies through thermodynamic cycles.

### S1.1. Experimental datasets.

The experimental datasets encompass both solute-solvent thermodynamic properties and bulk physicochemical properties of ILs, extracted from a vast body of literature. To ensure broad coverage and data reliability, we developed these datasets using a semi-automated literature-mining workflow. Approximately 10,000 papers were initially retrieved through automated scripts, followed by Large-Language-Model-(LLM)-assisted data extraction involving text retrieval and optical character recognition when necessary. The extracted records were subsequently manually verified.

Solvation-related properties are of utmost importance for the greener solvent ILs. However, no extensive database with a wide coverage of chemical space is currently available. In order to enable our AIonopedia model to solve practical problems like gas absorption (the physicochemical process where a solute becomes solvated in the greener solvent), we curated ~30,000 experimental datapoints of multi-temperature solvation free energies and biphasic transfer free energies involving approximately 80 ionic solvents and 150 solutes. To the best of our knowledge, this (i.e., the cation-anion-solute triple) represents the largest curated dataset currently available for these properties. Owing to the thermodynamic relationship among solvation free energy, transfer free energy, and hydration free energy, hydration behavior can be inferred directly from the collected datasets.

In addition to solute-solvent interactions, we collected datasets also for several crucial bulk properties, including mass density, viscosity, surface tension, and melting point. Compared with the solute-solvent datasets, the bulk-property datasets span substantially broader ion-pair combination space, comprising ~6,000 ILs formed by approximately 1,200 cations and 300 anions.

In Fig. S1B, we present the 1D distributions for all properties, along with the temperature-expanded 2D distributions when applicable. As melting points are intrinsic thermodynamic properties and transfer free energies are almost exclusively reported at 298 K, these two properties do not have 2D distributions.

To characterize the chemical diversity of the experimental datasets, we computed four complementary molecular fingerprints (ECFP, MACCS, Atom Pair, and PubChem fingerprints) using RDKit and chemfp[1], followed by hierarchical clustering. Fig. S1C presents the resulting molecular similarity (Tanimoto similarity[2]) matrix for all species, including 1,159 cations, 287 anions, and 328 neutral molecules (solutes and organic solvents). The low overall similarity indicated by the predominantly dark matrix with sparse highlights underscores the breadth of our chemical coverage.

We further analyzed the distribution of chemical space using t-SNE projections derived from both our descriptor set and traditional molecular similarity metrics (Fig. S1D). While both approaches distinguish ionic species from neutral molecules, the descriptor-based representation yields substantially clearer separation, indicating that it captures chemically meaningful distinctions relevant to IL systems. As a local consistency check, two structurally related cations, 1-butyl-3-methylimidazolium ($[BMIM]^+$) and 1-hexyl-3-methylimidazolium ($[HMIM]^+$), remain closely clustered in descriptor space, consistent with their underlying structural similarity.

**S1.2. Construction of the multimodal pretraining corpora.**

Although the experimental datasets provide high-quality supervision, the coverage of chemical space and the scale remain insufficient for training multimodal foundation models with a good generalization behavior. We therefore constructed substantially larger synthetic corpora for modality-alignment pretraining.

Unlike existing deep learning frameworks on ionic solvents[3-5] that directly leverage ultra-large molecular databases such as ZINC20[6] for pretraining, we adopted an IL-centric strategy. Up to now, merely <10,000 ILs have been reported and characterized experimentally, and the largest dedicated resource named ILThermo[7] contains only approximately 3,000 IL species. Such a sample size is obviously too small and the coverage of chemical space is limited. Consequently, direct pretraining on chemically broad ultra-large databases dominated by non-IL molecules could introduce representation biases that are poorly aligned with the target domain (i.e., the IL chemical space). Further, as the language backbones employed in this work were already pretrained on large general corpora, large-scale language pretraining was unnecessary.

Instead, we prioritized domain relevance over corpus size. Starting from known IL species deposited/explored experimentally, we expanded the chemical space through similarity-based retrieval from large molecular databases and generated virtual systems by combinatorial sampling of cations, anions, solutes, and organic solvents. The resulting pretraining corpora span four molecular-system formats, corresponding to temperature-conditioned bulk IL systems, temperature-independent bulk IL systems, IL-solute systems, and organic-solvent-solute systems. For the temperature-conditioned formats, each virtually generated molecular system was paired with a fixed temperature during sampling. These procedures yielded the modality-alignment synthetic corpora, comprising approximately 2.8 million unique synthetic systems in total, the scale of which exceeds that of the experimental datasets by about two orders of magnitude.

Each synthetic system was annotated with pseudo-labels as the supervised component of modality-alignment training. The pseudo-labels consist of molecular descriptors together with system-format identifiers. Specifically, descriptor representations include the number of hydrogen bond donors, the number of hydrogen bond acceptors, the number of rotatable bonds, the polar surface area,[8] the number of atomic stereocenters, the octanol-water partition coefficient log P and molar refractivity using the Wildman-Crippen method,[9] the fraction of sp3 carbon, the number of rings, the number of heterocycles, the number of aromatic rings, the number of aromatic heterocycles, the number of spiro atoms, the molecular weight, the number of heteroatoms, the number of heavy atoms, the κ shape indices[10], the Balaban J index[11] and the Bertz complexity index[12], which provide chemically informative summaries of molecular composition and topology.

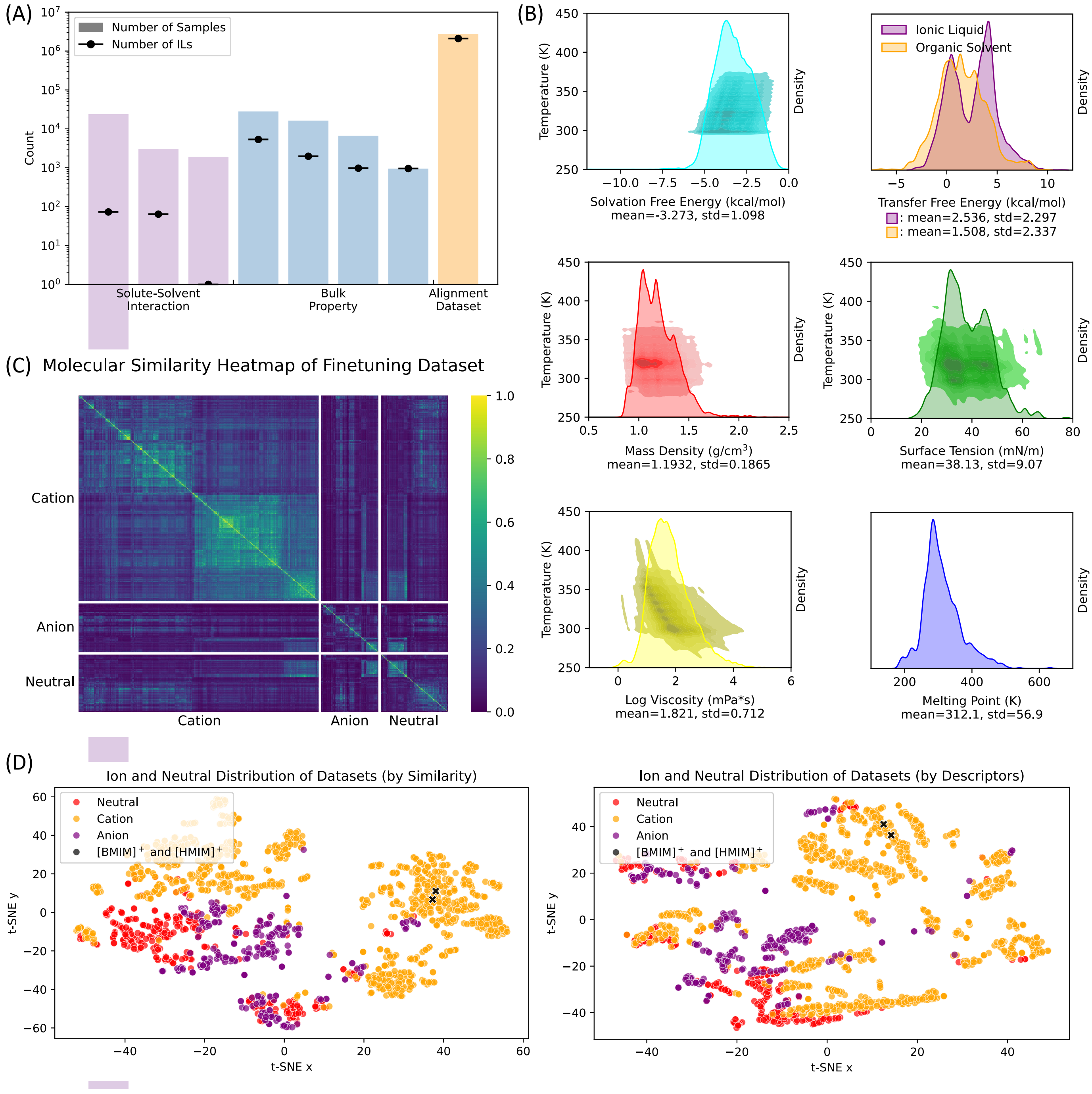

**Fig. S1. Dataset composition, property distributions, and chemical-space diversity.** (A) The sample sizes and the numbers of IL species in our dataset. From left to right: solvation free energy, transfer free energy between IL and water phases, transfer free energy between organic solvent and water phases (non-IL species; plotted as 1 on the log scale), mass density, viscosity, surface tension, melting point, and the modality-alignment dataset. (B) 1D and 2D KDE distributions for six independent training properties. Note that the training data for hydration free energy is not an independent component due to its relationship with solvation-in-IL free energy and biphasic transfer free energy. (C) Molecular similarity heatmap of molecular species in the experimental dataset, including 1159 cations, 287 anions, and 328 neutral molecules (solutes and organic solvents). (D) t-SNE visualization of chemical-space distribution of ions and neutral molecules in the experimental dataset.

## S2. AIonopedia employs a multimodal foundation model for ionic solvents.

### S2.1. Overall architecture.

AIonopedia employs a dual-tower multimodal foundation model that jointly represents IL systems through complementary language and graph modalities. The language branch, built upon a pretrained LLM, captures molecular semantics together with system-level contextual information, whereas the graph branch preserves molecular topology through graph-based representations. Rather than interacting only at the prediction stage, the two modalities are first aligned in a shared representation space and subsequently integrated through bidirectional cross-modal attention, enabling complementary information exchange between sequential and structural representations.

Unlike conventional molecular foundation models that encode individual molecules independently, AIonopedia considers an entire physicochemical system as the fundamental representation unit. For example, a solute dissolved in an IL at a specified temperature is represented as a single multimolecular sample rather than a collection of independent molecular embeddings. This system-level formulation enables the model to directly capture intermolecular interactions, compositional dependencies, and thermodynamic context that collectively determine experimentally observable physicochemical properties.

The overall framework consists of four consecutive components: (i) multimodal system representation, in which molecular sequences (SMILES) and graph structures are constructed for the complete IL systems; (ii) self-supervised cross-modal alignment, where language and graph representations are projected into a common latent space through momentum contrastive learning; (iii) bidirectional cross-modal fusion, where Transformer-based cross-attention modules exchange information between the two modalities; and (iv) multi-task supervised learning, which optimizes the fused multimodal representation for downstream physicochemical prediction. An overview of the architecture is shown in Fig. 1C.

### S2.2. System representation.

To represent complex multicomponent IL-involved physicochemical systems, AIonopedia constructs each training sample at the system level. Depending on the prediction task, the system-level input may consist of an ionic solvent alone (for bulk-property prediction) or an ionic/organic solvent together with a neutral solute (for solute-solvent thermodynamics). Consequently, all relevant components and conditions are organized within a unified representation, providing a consistent basis for subsequent multimodal processing.

For the language modality, each system is serialized into a structured textual prompt containing the canonical SMILES representations of all molecular components together with the corresponding experimental conditions, including temperature when applicable. This serialized prompt provides the textual input for the language branch.

For the graph modality, each molecular component, including cations, anions, solutes, and organic solvents, is converted from its SMILES representation into an attributed molecular graph, with atom indices traversed in canonical SMILES order to ensure consistent graph construction. Atom features include atom type, degree, hybridization state, implicit valence, aromaticity, formal charge, and hydrogen-bonding characteristics, whereas bond features encode bond order, stereochemistry, conjugation, and ring membership. The graph representations of all molecular components are subsequently concatenated together with system-level tokens representing experimental conditions to construct a unified graph sequence describing the complete physicochemical system.

### S2.3. Multimodal encoding.

The system-level representations described above are independently encoded by the language and graph branches before multimodal alignment. The language branch is built upon a pretrained decoder-only LLM, whereas the graph branch employs a Graph Transformer[13] to encode molecular topology. Together, these complementary representations capture both high-level chemical semantics and detailed structural information.

For the language modality, the serialized system prompt is tokenized and processed by the pretrained LLM to generate contextual token embeddings. In particular, LLMs trained on scientific corpora possess intrinsic domain-specific priors that support chemical problem comprehension and molecular representation. The language backbone is adapted with a parameter-efficient fine-tuning protocol, yielding a token sequence that serves as the language representation of the complete physicochemical system.

In parallel, each molecular graph is encoded using a shared Graph Transformer, allowing graph representations to be learned for the same molecular components described in the language prompt, including cations, anions, solutes, and organic solvents. The resulting molecular graph embeddings are subsequently concatenated together with system-level embeddings describing experimental conditions, including temperature and property type. This unified graph sequence is further processed by a lightweight Transformer-based graph-fusion encoder, enabling information exchange among all molecular components before multimodal integration.

To facilitate interaction between the two modalities, both the language and graph representations are projected into a common hidden space through lightweight two-layer MultiLayer Perceptron (MLP) projectors, following the strategy adopted in LLaVA-1.5[14] (with the graph-side projection applied before the graph-fusion encoder). The final valid token from each modality is subsequently extracted as the global representation of the complete physicochemical system and serves as the input for the cross-modal alignment procedure.

### S2.4. Cross-modal alignment.

Although the language and graph branches describe the same physicochemical system, they encode complementary aspects of molecular information and therefore reside in different representation spaces. Before information exchange between the two modalities, we first align their global representations through self-supervised contrastive learning, encouraging the language and graph encoders to produce consistent embeddings for the same molecular system while preserving discrimination among different systems.

To improve the stability of representation learning, we adopt a momentum graph-text contrastive (GTC) framework[15] based on the InfoNCE[16] objective. Instead of constructing contrastive pairs from individual molecules, each positive pair corresponds to the complete physicochemical system (see Section S2.2), thereby preserving intermolecular context during modality alignment. Momentum encoders and first-in-first-out feature queues are incorporated following the strategy commonly adopted in momentum contrastive learning.

For the $i$-th sample in a mini-batch of size $B$, let $\mathbf{t}_i$ and $\boldsymbol{g}_i$ denote the $L_2$-normalized final-token representations from the LLM and graph branches, respectively. Their momentum-encoder counterparts are denoted as $\bar{\mathbf{t}}_i$ and $\bar{\boldsymbol{g}}_i$. To provide a large and stable set of contrastive candidates, we maintain two first-in-first-out momentum queues for the LLM and graph branches, denoted as $\mathcal{Q}_t$ and $\mathcal{Q}_g$, respectively. Each queue stored $K = 65{,}536$ $L_2$-normalized feature vectors. The momentum encoders are updated by exponential moving average

$$\boldsymbol{\theta}_m \leftarrow m\boldsymbol{\theta}_m + (1 - m)\boldsymbol{\theta} \qquad \text{(S1)},$$

where $\boldsymbol{\theta}$ and $\boldsymbol{\theta}_m$ denote the parameters of the online encoder and momentum encoder, respectively, and the momentum coefficient was set to $m = 0.995$. Using the current mini-batch together with queued momentum features as candidates, each sample has one matched positive pair and multiple unmatched negative pairs. Specifically, the text-to-graph and graph-to-text candidate sets are constructed as

$$\mathcal{G} = \{\bar{\boldsymbol{g}}_1, \dots, \bar{\boldsymbol{g}}_B\} \cup \mathcal{Q}_g, \mathcal{T} = \{\bar{\boldsymbol{t}}_1, \dots, \bar{\boldsymbol{t}}_B\} \cup \mathcal{Q}_t \qquad \text{(S2)}.$$

Thus, the candidate index $j$ runs over $B + K$ candidates. The hard contrastive target was defined as

$$y_{ij} = \begin{cases} 1, & \text{if candidate } j \text{ is the matched counterpart of sample } i \\ 0, & \text{otherwise} \end{cases} \qquad \text{(S3)}.$$

The text-to-graph and graph-to-text similarities are computed as

$$s_{ij}^{t\rightarrow g} = \frac{\mathbf{t}_i^{\mathsf{T}} \bar{\boldsymbol{g}}_j}{\tau}, s_{ij}^{g\rightarrow t} = \frac{\boldsymbol{g}_i^{\mathsf{T}} \bar{\mathbf{t}}_j}{\tau} \quad \text{(S4)},$$

where the contrastive temperature was set to $\tau = 0.5$. Instead of using only the one-hot hard target $y_{ij}$, we adopted a soft target by blending $y_{ij}$ with the similarity distribution predicted by the momentum encoders

$$q_{ij}^{t\rightarrow g} = \alpha \frac{\exp(\bar{\mathbf{t}}_i^{\mathsf{T}} \bar{\mathbf{g}}_j/\tau)}{\sum_{k=1}^{B+K} \exp(\bar{\mathbf{t}}_i^{\mathsf{T}} \bar{\mathbf{g}}_k/\tau)} + (1-\alpha) y_{ij} \quad \text{(S5)},$$

$$q_{ij}^{g\rightarrow t} = \alpha \frac{\exp(\bar{\mathbf{g}}_i^{\mathsf{T}} \bar{\mathbf{t}}_j/\tau)}{\sum_{k=1}^{B+K} \exp(\bar{\mathbf{g}}_i^{\mathsf{T}} \bar{\mathbf{t}}_k/\tau)} + (1-\alpha) y_{ij} \quad \text{(S6)},$$

where $k$ also runs over all $B+K$ candidates, and the soft-target mixing coefficient was set to $\alpha = 0.4$. After each mini-batch, the newly computed momentum features are enqueued, and the oldest features are removed from the queues. The final GTC objective is defined as

$$L_{\mathrm{GTC}} = \frac{1}{2}\left(L_{t\rightarrow g} + L_{g\rightarrow t}\right) \quad \text{(S7)},$$

where the text-to-graph and graph-to-text losses are respectively given by

$$L_{t\rightarrow g} = -\frac{1}{B} \sum_{i=1}^{B} \sum_{j=1}^{B+K} q_{ij}^{t\rightarrow g} \log \frac{\exp\left(s_{ij}^{t\rightarrow g}\right)}{\sum_{k=1}^{B+K} \exp\left(s_{ik}^{t\rightarrow g}\right)} \quad \text{(S8)},$$

$$L_{g\rightarrow t} = -\frac{1}{B} \sum_{i=1}^{B} \sum_{j=1}^{B+K} q_{ij}^{g\rightarrow t} \log \frac{\exp\left(s_{ij}^{g\rightarrow t}\right)}{\sum_{k=1}^{B+K} \exp\left(s_{ik}^{g\rightarrow t}\right)} \quad \text{(S9)}.$$

Optimizing this objective projects the two modalities into a common latent representation space, providing a well-aligned initialization for the subsequent bidirectional cross-modal fusion.

### S2.5. Bidirectional cross-modal fusion.

Following modality alignment, the two representations reside in a shared latent space but still preserve complementary information originating from their respective modalities. To enable mutual information exchange while retaining the strengths of each modality, we employ a cross-modal fusion strategy based on Transformer cross-attention.[17] Unlike conventional multimodal architectures that perform one-way feature injection, AIonopedia allows each modality to attend to the other through bidirectional cross-modal attention.

Let $\mathbf{T}_i \in \mathbb{R}^{L_t \times d}$ denote the projected LLM token sequence and $\mathbf{G}_i \in \mathbb{R}^{L_g \times d}$ denote the graph-side sequence after graph-fusion encoding, where $L_t$ and $L_g$ are the sequence lengths of the text and graph-side inputs, respectively, and $d$ is the shared hidden dimension.

In the language-query branch, the language representation (text sequence) serves as the query, while the graph representation provides the corresponding keys and values. The cross-attention operation is formulated

as

$$\text{CrossAttn}_{t\leftarrow g} = \text{softmax}\left(\frac{(\mathbf{T}_i\mathbf{W}_Q^t)(\mathbf{G}_i\mathbf{W}_K^g)^{\top}}{\sqrt{d}}\right)\mathbf{G}_i\mathbf{W}_V^g \quad \text{(S10)},$$

where $\mathbf{W}_Q^t$, $\mathbf{W}_K^g$, and $\mathbf{W}_V^g$ are learnable projection matrices for the query, key, and value representations, respectively. This operation is implemented within a Transformer decoder block, yielding the text-conditioned cross-modal representation

$$\mathbf{H}_i^t = D_t(\mathbf{T}_i, \mathbf{G}_i) \quad \text{(S11)},$$

where $D_t$ denotes the language-query Transformer decoder module, and a causal attention mask is applied only in this branch to preserve the autoregressive property of the language model. Similarly, the graph-query branch follows the same formulation in the opposite direction, $g \leftarrow t$, allowing the graph representation to attend to the language representation

$$\mathbf{H}_i^g = D_g(\mathbf{G}_i, \mathbf{T}_i) \quad \text{(S12)},$$

where $D_g$ denotes the graph-query Transformer decoder module. The final valid token representations from $\mathbf{H}_i^t$ and $\mathbf{H}_i^g$, denoted as $\mathbf{h}_i^t$ and $\mathbf{h}_i^g$, respectively, are concatenated to obtain the fused representation

$$\mathbf{z}_i = [\mathbf{h}_i^t; \mathbf{h}_i^g] \quad \text{(S13)},$$

which serves as the shared representation for all downstream learning objectives.

**S2.6. Multi-task learning objectives.**

To jointly optimize multimodal representation learning and downstream transferability, AIonopedia is trained with three additional objectives[18] complementary to the contrastive objective described above during multimodal pretraining. The contrastive objective aligns the global language and graph representations before cross-modal fusion, whereas the additional objectives supervise the fused multimodal representations from three perspectives: cross-modal correspondence, graph-conditioned language generation, and physicochemical representation learning. Collectively, they encourage the learned representations to remain structurally consistent across modalities while preserving chemically meaningful information required for downstream property prediction.

The first additional objective is Graph-Text Matching (GTM), which determines whether the language and graph representations originate from the same physicochemical system. For each positive graph-text pair, two mismatched negative pairs are generated by replacing either the language or graph modality with another sample from the same mini-batch, which results in $3B$ graph-text pairs in total. Rather than sampling these mismatched pairs uniformly, negative samples are drawn according to the cross-modal similarity distribution within the mini-batch, with the matched diagonal entries masked out, thereby emphasizing harder mismatched pairs during GTM training. The GTM objective is formulated as a binary classification task using a GTM-specific feed-forward classifier applied to the fused final-token representation

$$L_{\mathrm{GTM}} = -\frac{1}{3B}\sum_{r=1}^{3B}[y_r^{\mathrm{GTM}}\log p_r^{\mathrm{GTM}} + (1 - y_r^{\mathrm{GTM}})\log(1 - p_r^{\mathrm{GTM}})] \quad \text{(S14)},$$

where $y_r^{GTM} \in \{0,1\}$ and $p_r^{\mathrm{GTM}}$ denote the ground-truth matching label and the predicted matching probability for the $r$-th graph-text pair, respectively.

The second additional objective is Graph-grounded Text Generation (GTG), which encourages the language branch to generate chemically consistent textual representations conditioned on graph-derived structural information. Specifically, autoregressive next-token prediction is performed using the graph-conditioned language representations produced by the text-query decoder. The corresponding language-modeling objective is defined as

$$\mathcal{L}_{\mathrm{GTG}} = -\frac{\sum_{i=1}^{B}\sum_{l=1}^{L_t-1} m_{i,l+1}\log p_\theta\left(x_{i,l+1}|\mathbf{H}_{i,\le l}^{t}\right)}{\sum_{i=1}^{B}\sum_{l=1}^{L_t-1} m_{i,l+1}} \quad \text{(S15)},$$

where $x_{i,l}$ is the $l$-th token of the input prompt, this loss is evaluated only over valid (non-padding, defined by $m_{i,l} \in \{0,1\}$) language tokens, and $p_\theta(x_{i,l+1} \mid \mathbf{H}_{i,\le l}^{t})$ is the predicted next-token probability conditioned on the graph-grounded text representation.

To further inject physicochemical knowledge into the multimodal representation, we introduce a supervised pseudo-label objective. Each synthetic sample is assigned a pseudo-label consisting of a continuous molecular descriptor vector together with a categorical system-format label

$$\mathbf{y}_i^{\mathrm{sup}} = [\mathbf{d}_i; \mathrm{onehot}(c_i)], \mathbf{d}_i \in \mathbb{R}^{64}, \mathrm{onehot}(c_i) \in \mathbb{R}^{C} \quad \text{(S16)},$$

where $\mathbf{d}_i$ denotes the continuous descriptor vector, $c_i$ denotes the system-format class label, and $C$ is the number of system-format classes. A separate feed-forward prediction head $f_{\mathrm{sup}}$ maps the same fused multimodal representation $\mathbf{z}_i$ to the corresponding pseudo-label

$$\hat{\mathbf{y}}_i^{\mathrm{sup}} = f_{\mathrm{sup}}(\mathbf{z}_i) = [\hat{\mathbf{d}}_i; \hat{\mathbf{c}}_i], \hat{\mathbf{d}}_i \in \mathbb{R}^{64}, \hat{\mathbf{c}}_i \in \mathbb{R}^{C} \quad \text{(S17)},$$

where $\hat{\mathbf{d}}_i$ denotes the predicted descriptor vector and $\hat{\mathbf{c}}_i$ denotes the logits for system-format classification. The regression and classification objectives are respectively defined as

$$L_{\text{reg}} = \frac{1}{B}\sum_{i=1}^{B} \| \hat{\mathbf{d}}_i - \mathbf{d}_i \|_2^2 \quad \text{(S18)},$$

$$L_{\text{cls}} = -\frac{1}{B}\sum_{i=1}^{B} \log \frac{\exp(\hat{c}_{i,c_i})}{\sum_{k=1}^{C} \exp(\hat{c}_{i,k})} \quad \text{(S19)}.$$

The two loss functions are then combined to produce the supervised objective

$$L_{\text{sup}} = L_{\text{reg}} + 0.5L_{\text{cls}} \quad \text{(S20)}.$$

Finally, the overall pretraining objective is defined as the weighted sum of the GTM, GTC, GTG, and supervised objectives

$$\begin{aligned} L &= 2L_{\text{GTM}} + 0.2L_{\text{GTC}} + 0.2L_{\text{GTG}} + L_{\text{sup}} \\ &= 2L_{\text{GTM}} + 0.2L_{\text{GTC}} + 0.2L_{\text{GTG}} + L_{\text{reg}} + 0.5L_{\text{cls}} \quad \text{(S21)}. \end{aligned}$$

After the modality-alignment stage, the used auxiliary prediction heads are discarded and replaced by task-specific regression heads for downstream physicochemical property prediction, as described in Section S3.

## S3. Training procedures.

### S3.1. Backbone implementations.

To investigate the applicability of different pretrained LLM backbones to IL representation learning, we constructed AIonopedia using three representative LLM families, namely Galactica, Qwen3, and Gemma3. Models of comparable parameter scales were selected, including Galactica-1.3b, Qwen3-0.6b, and Gemma3-1b. Among them, Galactica and Qwen3 explicitly state that they were trained on scientific texts to enhance domain knowledge, whereas Gemma3 serves as a general-purpose LLM control without such a claim. To further investigate the scaling behavior, the substantially smaller Galactica-125m model was additionally included as a lightweight baseline.

For all backbones, the language input consisted of a unified prompt containing canonical SMILES together with system-level contextual information. Token sequences were dynamically padded to the longest sequence within each mini-batch. Rather than performing full-parameter optimization, all language backbones were adapted using Low-Rank Adaptation (LoRA)[19], substantially reducing computational overhead and GPU memory consumption while preserving the pretrained language representations. The LoRA rank was fixed to 16 with lora_alpha = 32 and a dropout rate of 0.1. Following the implementation of each backbone, LoRA layers were inserted into the attention projection layers (q_proj, k_proj, v_proj and out_proj/o_proj).

The graph branch shared the same molecular encoder architecture across all downstream tasks. Molecular graphs were encoded individually before being combined with the temperature token and property-type embedding to construct the system-level graph sequence described in Section S2. For Galactica-1.3b, Qwen3-0.6b and Gemma3-1b, the graph hidden dimension was fixed at 256, whereas the lightweight Galactica-125m model employed a hidden dimension of 128. To further match its reduced language-backbone scale, Galactica-125m was also paired with proportionally shallower graph Transformer, graph-fusion, and cross-attention modules. Missing molecular components (for example, systems without solutes or ions) were represented using empty graph placeholders, allowing a unified multimodal input format to support both IL systems and conventional molecular solvent systems without modifying the model architecture.

### S3.2. Self-supervised modality alignment.

The multimodal foundation model was first optimized using the synthetic dataset described in Section S1 through the self-supervised modality-alignment framework introduced in Section S2. During this stage, all backbone models were trained for five epochs using distributed data parallel training on eight NVIDIA A100 80GB GPUs. Each GPU processed a mini-batch of 32 samples, and gradients were accumulated over two steps, resulting in an effective global batch size of 512 per optimizer update. Under this setup, the default Qwen3-0.6b model required approximately 14 h to complete pretraining.

Optimization was performed using the AdamW optimizer under mixed-precision bfloat16 training. A cosine learning-rate schedule with 1,000 warm-up steps was adopted throughout training, together with a weight decay of 0.01 and gradient clipping using a maximum gradient norm of 2. Learning rates were selected according to model size, with $4 \times 10^{-4}$ for Galactica-125m, $2 \times 10^{-4}$ for Qwen3-0.6b, and $1.5 \times 10^{-4}$ for both Galactica-1.3b and Gemma3-1b.

Validation was performed after every training epoch, and checkpoints containing both the LoRA adapter parameters and all non-LLM modules were saved independently for subsequent downstream adaptation.

### S3.3. Property-specific adaptation.

Following modality alignment, the pretrained multimodal foundation model was adapted independently to each downstream physicochemical property. The multimodal input format remained identical to that used during pretraining, while the auxiliary prediction heads employed during the former stage were replaced by task-specific scalar regression heads. During property-specific fine-tuning, the regression heads and the remaining trainable components of the multimodal model were jointly optimized. Model optimization was then performed using L2 loss on normalized experimental property values.

To rigorously evaluate model generalization under practically relevant scenarios, we adopted multiple chemically meaningful 5-fold cross-validation protocols rather than conventional random data splitting[20]. For bulk properties, two complementary evaluation strategies were employed, namely cation-based splitting and IL-based splitting, which respectively assess extrapolation to previously unseen cation families and entirely unseen ion-pair compositions. For the organic-solvent/water transfer free energy dataset, which does not involve ionic species, analogous solvent-based and solute-solvent-pair-based splitting strategies were adopted. For ternary thermodynamic datasets involving solutes, cations and anions, an additional ternary-component split was introduced, in which each unique solute-cation-anion combination appeared exclusively in either the training or validation partition. This protocol represents a substantially more challenging OOD setting than random splitting by explicitly preventing memorization of molecular combinations.

Specifically, hydration free energy was evaluated without training an additional property-specific model. As described in Section S1, it was estimated by summing the predicted gas-to-IL solvation free energy and IL-to-water transfer free energy through IL intermediates. For each solute, ten deduplicated virtual ILs were constructed by sampling ion combinations from the solvent-solute interaction datasets after excluding known solute-IL pairs, and the final solute-level prediction was obtained by averaging the resulting thermodynamic-cycle estimates. The benchmark hydration free energy labels were experimentally collected at the solute level, and the virtual ILs were introduced only to obtain leakage-free model predictions. This protocol slightly underestimates performance variance but reduces small-sample bias and better matches real-world usage. Detailed information on the hydration dataset construction is provided in Section S5.1.

Unless otherwise specified, all hyperparameters were determined using the most stringent cation-based split and subsequently applied without further tuning to all remaining evaluation protocols. This unified evaluation strategy enabled consistent comparisons across different physicochemical properties while providing a rigorous assessment of model generalization beyond previously observed IL systems. (Table S1)

**Table S1.** Hyperparameters in supervised fine-tuning on different tasks.

| Base Model | Property | Epochs | Batch size / GPU | Warmup steps | LR: FFN ($\times 10^{-4}$) | LR: decoder ($\times 10^{-4}$) | LR: other modules ($\times 10^{-4}$) |
|---|---|---|---|---|---|---|---|
| Galactica-125m[21] | Solvation ΔG | 10 | 16 | 50 | 0.6 | 0.4 | 0.3 |
| | IL/water transfer ΔG | 10 | 16 | 20 | 0.6 | 0.4 | 0.3 |
| | Mass density | 15 | 16 | 50 | 1.8 | 1.2 | 0.9 |
| | Viscosity | 10 | 16 | 50 | 3 | 2 | 1.5 |
| | Surface tension | 20 | 16 | 20 | 0.9 | 0.6 | 0.45 |
| | Melting point | 15 | 8 | 10 | 3 | 2 | 1.5 |
| | Organic solvent/water transfer ΔG | 15 | 8 | 10 | 3 | 2 | 1.5 |
| Galactica-1.3b | Solvation ΔG | 10 | 16 | 50 | 0.45 | 0.3 | 0.225 |
| | IL/water transfer ΔG | 10 | 16 | 20 | 0.45 | 0.3 | 0.225 |
| | Mass density | 15 | 16 | 50 | 0.6 | 0.4 | 0.3 |
| | Viscosity | 10 | 16 | 50 | 2.4 | 1.6 | 1.2 |
| | Surface tension | 20 | 16 | 20 | 0.6 | 0.4 | 0.3 |
| | Melting point | 15 | 8 | 10 | 1.2 | 0.8 | 0.6 |
| | Organic solvent/water transfer ΔG | 15 | 8 | 10 | 3 | 2 | 1.5 |
| | Solvation ΔG | 10 | 16 | 50 | 0.4 | 0.3 | 0.3 |
| | IL/water transfer ΔG | 10 | 16 | 20 | 0.4 | 0.3 | 0.3 |
| | Mass density | 10 | 16 | 20 | 1.2 | 0.9 | 0.9 |
| | Viscosity | 10 | 16 | 20 | 1.2 | 0.9 | 0.9 |
| Qwen3-0.6b[22] | Surface tension | 10 | 16 | 20 | 0.6 | 0.45 | 0.45 |
| | Melting point | 15 | 8 | 10 | 0.8 | 0.6 | 0.6 |
| | Organic solvent/water transfer ΔG | 15 | 8 | 10 | 1.2 | 0.9 | 0.9 |
| | Solvation ΔG | 10 | 16 | 50 | 0.4 | 0.3 | 0.3 |
| | IL/water transfer ΔG | 10 | 16 | 20 | 0.4 | 0.3 | 0.3 |
| | Mass density | 10 | 16 | 20 | 1.6 | 1.2 | 1.2 |
| Gemma3-1b[23] | Viscosity | 10 | 16 | 20 | 1.6 | 1.2 | 1.2 |
| | Surface tension | 10 | 16 | 20 | 0.8 | 0.6 | 0.6 |
| | Melting point | 15 | 8 | 10 | 0.8 | 0.6 | 0.6 |
| | Organic solvent/water transfer ΔG | 15 | 8 | 10 | 2 | 1.5 | 1.5 |

## S4. Agentic framework.

### S4.1. Agent architecture.

AIonopedia implements an end-to-end agentic framework for IL discovery by combining LLM-based reasoning with domain-specific computational tools.[24-26] Rather than relying solely on parametric model knowledge, the agent dynamically invokes external tools for literature retrieval, chemical identification, property prediction, and molecular optimization, thereby improving the reliability and practical utility of AI-assisted IL research.

At the core of the framework is a planner implemented using GPT-5[27], which orchestrates all downstream modules following the Reasoning and Acting (ReAct)[28] paradigm. The planner iteratively alternates between Thought, Action, and Observation phases. In the Thought phase, the planner interprets the user request and performs zero-shot Chain-of-Thought (CoT)[29] reasoning to decompose complex scientific problems into executable subtasks. In the Action phase, it selects the appropriate computational tool and constructs the corresponding input arguments conditioned on the current reasoning state. The selected tool is then executed, and the returned outputs are incorporated in the Observation phase, which informs the next reasoning step. This iterative reasoning-acting cycle continues until task completion, enabling multi-step scientific workflows that require both symbolic reasoning and numerical computation.

The specific functions of each tool within this end-to-end workflow are described in the following subsections.

### S4.2. Knowledge retrieval and chemical grounding.

The knowledge-retrieval pipeline provides externally verifiable scientific evidence that complements the internal knowledge of the planner and serves as the grounding layer for all downstream reasoning.

The web searcher adopts an LLM-controlled fused-search architecture. Queries generated by the planner are first reformulated by the searcher's internal GPT-5 model and submitted to the Serper API[30] that enables access to Google Scholar. For each retrieved publication, the system further queries the Semantic Scholar API[31] to obtain enriched metadata and abstracts when available. When scholarly databases fail to provide sufficient coverage, a general web search is automatically invoked as a fallback mechanism. This hierarchical retrieval strategy ensures robustness across diverse scenarios ranging from IL literature mining to general scientific information retrieval. By grounding responses in externally verifiable sources, this Retrieval-Augmented Generation (RAG) [32] framework effectively reduces planner hallucination and improves response reliability.

In parallel, the PubChem searcher[33] is dedicated to chemical structure grounding. Molecular names, abbreviations, and synonyms are first mapped to SMILES representations using PubChem. Because ionic species are often incompletely or inconsistently represented in public databases, an internal LLM is further employed to perform reasoning-based correction, including reconstruction of chemically valid ionic forms and assignment of appropriate charge states when necessary. This reasoning-enhanced retrieval substantially improves the robustness of IL structure identification.

Furthermore, common line-angle molecular structures are processed using a DECIMER[34]-based molecule processor, which identifies molecular structures from chemical images and converts them into corresponding SMILES representations. An RDKit[35] standardization module then converts SMILES strings obtained from either image recognition or web-based retrieval into a specified canonical form, enabling the agent to handle common molecule-centered image-text reasoning tasks. In parallel, a Python shell code processor provides an executable environment for complex processing tasks such as numerical processing, intermediate computation, tabular data processing, and visualization. After preprocessing, the planner extracts the essential inputs required for property prediction, including molecular structures, system composition, temperature, and target properties, thereby establishing a unified interface between external knowledge and the multimodal foundation model.

**S4.3. Property prediction interface.**

Property prediction is performed using the multimodal foundation model described in Section S2. Following chemical standardization, the planner extracts all required system-level information, including IL composition, solute identity, solvent identity (e.g., water or organic solvents when applicable), temperature, and target property specification. These inputs are transformed into the unified multimodal representation required by the domain-specific model.

The predictor simultaneously supports bulk physicochemical properties and solute-solvent thermodynamics. The former includes mass density, viscosity, surface tension, and melting point, while the latter includes solvation free energy, biphasic transfer free energy (between IL/water and organic-solvent/water phases) and hydration free energy. Because all predictions are produced within a single unified model architecture, the agent can perform consistent multi-property inference and integrate thermodynamic quantities within a coherent framework.

Importantly, predicted properties are not only used as final outputs but also serve as intermediate signals for downstream decision-making in the agentic loop, enabling property-guided ion screening and refinement during molecular discovery.

**S4.4. IL discovery with agentic AI.**

Beyond property prediction, AIonopedia enables autonomous IL discovery through an agent-guided heuristic strategy. Instead of relying on generative molecular models, which often suffer from limited chemical validity, poor synthesizability, and weak controllability in IL chemical space, we reformulate IL design as a constrained exploration problem that combines similarity-guided candidate screening with targeted structure modification around experimentally validated hits.

The discovery process begins by selecting high-performing ILs (or solutes in solute discovery) from existing experimentally measured datasets using the property predictor. These data-derived entries serve as seed points for subsequent exploration. Candidate expansion is performed through beam search guided by similarity constraints, where both the full set of collected IL species and external molecular databases are explored using Tanimoto similarity. This allows structurally related yet previously unexplored molecules to be proposed while maintaining proximity to known valid chemical regions.

Each newly proposed candidate is evaluated using the property predictor, forming an iterative evaluation-selection cycle. This procedure enables progressive refinement of the search pool by promoting molecules with improved target properties while eliminating inferior ones. The final top-ranked hits are prioritized for experimental validation.

After the initial wet-lab validation of the prioritized candidates, experimental feedback is used to identify the most promising one, which is then returned to AIonopedia for further structure modification. This allows the agent to explore its local chemical neighborhood through targeted molecular edits in search of potentially improved IL candidates. Detailed examples of the agent-driven screening and modification procedures are provided in Section S13.

By coupling predictive modeling with similarity-guided screening, wet-lab feedback, and targeted structure modification, the agentic framework transforms IL discovery into an efficient exploration-exploitation process over a chemically constrained space, substantially accelerating the identification of experimentally promising ionic solvents.

## S5. Molecular simulations.

### S5.1. Hydration free energy calculation.

Molecular simulations are considered one of the key tools for extracting atomistic insights and thermodynamic behaviors of ionic solvents. We construct the all-atom models of solute molecules by obtaining the 3D coordinates from experimental structures (PubChem). The solute molecules are parametrized with restrained electrostatic potential (RESP)[36] charges (geometry optimization, scan of molecular electrostatic potentials and regularized two-step fitting) and GAFF2[37] parameters for the other force-field terms. For the generation of the molecular electrostatic potential, while traditionally the gas-phase HF/6-31G* level is selected as a treatment of fortuitous error cancellation, in recent years it has been criticized in various aspects, such as the small size of the basis set and the low-level selection HF. As a result, there is a tendency to elevate it to a higher-level selection (e.g., B3LYP with a larger basis set) and incorporate implicit-solvent models (e.g., PCM variants). To provide a solid reference baseline from the simulation perspective, we present an exhaustive scan on this charge-generation option by considering the traditional HF/6-31G*, 13 additional gas-phase levels in combination with a larger basis set def2-TZVPP, including (HF, B3LYP, BP86, M052X, M06HF, M062X, MN15, PBE, PW6B95, TPSSh, wB97X-D, B2PLYP, mPW2PLYP), their implicit-solvent components (IEFPCM) and an additional gas-phase HF-based selection but with an even larger basis set HF/def2-QZVPP. The selected levels of theory cover selections used in most force-field-based investigations in the computational chemistry and biophysics community, such as PW6B95 employed in the recent RESP2 paper[38] and M06 derivatives used in water studies.[39] In total, we have 28 RESP charge sets for the hydration solutes, as listed in Table S2.

**Table S2.** A list of QM levels used to parametrize the RESP charges.

| | | | |
|---|---|---|---|
| HF/6-31G* | HF/def2-TZVPP | HF/def2-TZVPP IEFPCM | HF/def2-QZVPP |
| | B3LYP/def2-TZVPP | B3LYP/def2-TZVPP IEFPCM | |
| | BP86/def2-TZVPP | BP86/def2-TZVPP IEFPCM | |
| | M052X/def2-TZVPP | M052X/def2-TZVPP IEFPCM | |
| | M062X/def2-TZVPP | M062X/def2-TZVPP IEFPCM | |
| | M06HF/def2-TZVPP | M06HF/def2-TZVPP IEFPCM | |
| | MN15/def2-TZVPP | MN15/def2-TZVPP IEFPCM | |
| | PBE/def2-TZVPP | PBE/def2-TZVPP IEFPCM | |
| | PW6B95/def2-TZVPP | PW6B95/def2-TZVPP IEFPCM | |

| | |
|---|---|
| TPSSh/def2-TZVPP | TPSSh/def2-TZVPP IEFPCM |
| wB97XD/def2-TZVPP | wB97XD/def2-TZVPP IEFPCM |
| B2PLYP/def2-TZVPP | B2PLYP/def2-TZVPP IEFPCM |
| mPW2PLYP/def2-TZVPP | mPW2PLYP/def2-TZVPP IEFPCM |

The simulation cell is constructed with a single solute in its net-neutral form and TIP3P water molecules. The solute and water molecules are uncoupled at the beginning (i.e., without non-bonded electrostatic and van der Waals [vdW] interactions), and the nonequilibrium alchemical transformation is performed in the solute-creation direction. We would not describe many technical details of nonequilibrium free energy calculations in this work, but recommend readers to read relevant references[40-43]. We use GROMACS 2020.6,[44] for molecular simulations, the velocity rescaling algorithm[45] for temperature regulation at 298 K, the Parrinello-Rahman Lagrangian[46,47] with isotropic scaling for pressure regulation at 1 atm, the smooth Particle-mesh Ewald[48] method for long-range electrostatics and a 13 Å cutoff for vdW interactions. We equilibrate the solute-solvent-decoupled box for 1000 steps of energy minimization and 2 ns of NPT simulation. The equilibrated system is then sampled for 100 ns with an interval of 1 ns, producing 100 initial conditions as equilibrium snapshots. From these configurations, we initiate 1.2 ns fast-switching simulations consisting of a 0.8 ns vdW-coupling stage followed by a 0.4 ns charging stage, with a 0.5 $fs^{-1}$ frequency for the small perturbation of the alchemical order parameter.

The nonequilibrium works accumulated along the Hamiltonian switching process are processed with the asymptotically unbiased Jarzynski's scheme[49,50] or equivalently the exponential averaging (EXP) estimator, which computes the free energy difference between equilibrium states as the exponential average of microscopic nonequilibrium works $W_{\text{growth}}$, i.e., $\Delta \text{G}_{\text{solvation}} = -\ln\langle e^{-W_{\text{growth}}}\rangle$. Note that all thermodynamic properties are dimensionless here. The statistical uncertainty is extracted with numerical bootstrap resampling, in order to avoid the intrinsic (underestimation) behavior of the analytical formula.[51]

We summarize the 84 solute molecules included in the hydration set in Table S3. Instead of listing all numerical reference values in the table, the distribution of the experimental hydration free energies is summarized in Fig. S2A to provide an overview of the coverage of the dataset. In total, the benchmark set considering 28 RESP charge sets, 84 solutes and ~220 ns per solute-charge-set combination achieves an all-atom simulation length of ~0.5 ms. Such a ms-scale calculation is, to our best knowledge, the largest benchmark of solvation free energy calculations reported to date.

**Table S3.** Solutes included in the hydration benchmark set

| Solute Molecules | | | | | |
|---|---|---|---|---|---|
| 1,4-dioxane | 2-methyl-2-butanol | dichloromethane | methyl pentanoate | 1-pentyne | isopropyl benzene |
| 1-butanol | 2-pentanone | diethyl ether | methyl propanoate | 1-propanol | methane |
| 1-butene | 2-propanol | diisopropyl ether | methyl tert-butyl ether | 2,2,2-trifluoroethanol | methanol |
| 1-heptanol | 3-methylpentane | dimethyl ether | m-xylene | 2,2,4-trimethylpentane | methyl butanoate |
| 1-heptyne | 3-pentanone | ethane | nitromethane | 2-butanol | methyl cyclohexane |
| 1-hexanol | acetone | ethanol | n-pentane | 2-butanone | methyl cyclopentane |
| 1-hexene | acetonitrile | ethene | o-xylene | 2-methyl-1-propanol | methyl hexanoate |
| 1-hexyne | aniline | ethyl acetate | pentanal | carbon tetrachloride | pyridine |
| 1-methylcyclohexene | benzaldehyde | ethylbenzene | phenol | chloroform | tert-butanol |
| 1-methylimidazole | benzamide | heptanal | propanal | cyclohexane | tert-butylbenzene |
| 1-nitropropane | benzene | heptane | propane | cyclohexanol | tetrahydrofuran |
| 1-octene | butanal | hexanal | propene | cyclohexene | thiophene |
| 1-pentanol | butane | hexane | propyl benzene | cyclopentane | toluene |
| 1-pentene | butylbenzene | hydrogen sulfide | p-xylene | cyclopentene | triethylamine |

The hydration thermodynamics computed with nonequilibrium fast-switching simulations are compared with the reference values, and RMSE, MAE, Kendall rank correlation coefficient ($\tau$)[52] and Pearlman's predictive index (PI)[53] are computed as quality metrics. The performances of all parameter sets are compared in Fig. S2B. According to MAE, Kendall $\tau$ and PI, the HF/6-31G* RESP charges achieve the top performance, while under the RMSE criterion several DFT selections achieve a performance similar to the traditional selection HF/6-31G*. Notably, none of the gas-phase DFT selections or HF with enlarged basis sets (def2-TZVPP or def2-QZVPP) outperform the gas-phase HF/6-31G*, and the DFT-derived charges computed with implicit solvent models achieve performance metrics close to HF/6-31G*. These observations emphasize the incorporation of implicit-solvent models when elevating the level of theory in charge generation (i.e., using DFT), and also highlight the success of the fortuitous error cancellation of the traditional option. Therefore, we stick to the traditional HF/6-31G*-derived RESP charges in subsequent modelling of situations involving ionic solvents.

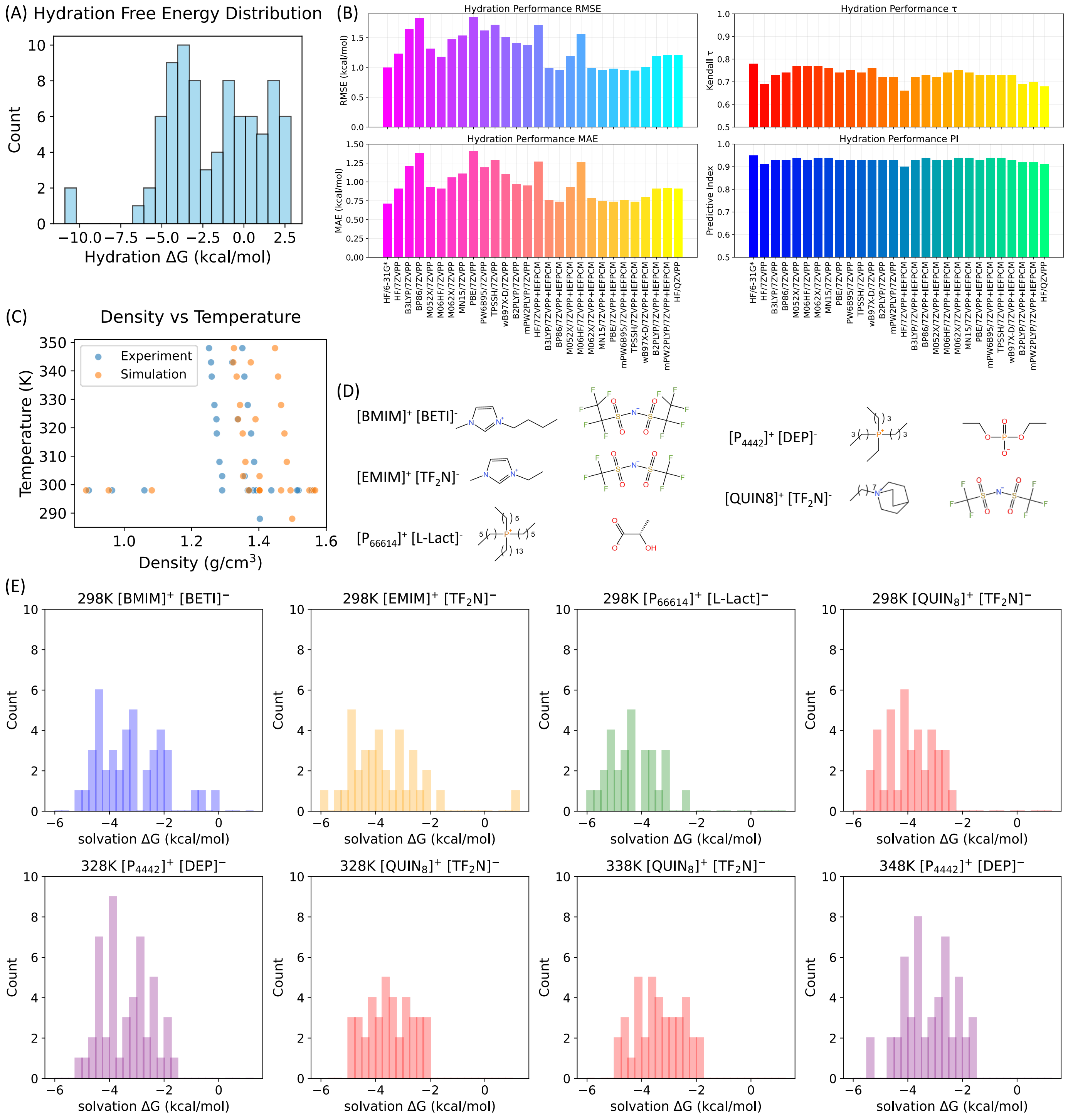

**Fig. S2. Molecular-simulation benchmarks and specific test-set characterization.** (A) Distribution of experimental hydration free energies in the benchmark set. (B) Performance of nonequilibrium alchemical free energy calculation, quantified using error metrics (RMSE and MAE) and ranking coefficients (Kendall τ and PI). (C) Temperature-density coverage of the mass-density external test set for bulk ionic solvents. Blue points denote experimental results, and orange points denote the corresponding simulation results. (D) Structures of ILs of the external test set for solvation-in-IL free energies. (E) Distributions of solvation free

energies in the external solvation-in-IL dataset at both the room temperature and elevated ones (Tables S4-S5).

### S5.2. Molecular simulations of ionic solvents.

Molecular modelling of ionic solvents is generally similar to the hydration situation detailed above. Specifically, we still use the PubChem coordinates for solutes. For ionic solvents, the PubChem coordinates are used when available, and the outputs of structure prediction algorithms (the Boltz structure prediction engine, to be specific) are employed for molecules not deposited in existing databases.

For the parametrization of fixed-charge force fields, we use the HF/6-31G* RESP charges in combination with GAFF2. A major difference is the scaling factor for atomic charges[54,55] (and sometimes also the vdW scaling factor)[56]. As the RESP charges are fitted in the absence of solvent-solvent interactions where charge transfer could be significant due to the ionic nature of these species, the atomic charges are often scaled down as an approximate treatment incorporating the charge transfer effect. As a result, the net charges of individual molecules become a fraction of the physical number (e.g., -1 and +1). The most popular treatment is uniform scaling, i.e., applying the same charge scaling factor to all atoms in the molecule, with the value of 0.8 as a common selection. Note that this charge scaling adjustment is only imposed on molecules forming the ionic solvent, while the net-neutral solutes remain untouched.

The IL box is constructed by inserting ~13000 atoms in the simulation cell. As a result, the number of IL pairs in the simulation cell exhibits a composition dependence, but the minimum number of ion pairs is 120, which is sufficient to eliminate finite-size artifacts for mass densities and solvation thermodynamics, according to our previous benchmark test[57]. The packing procedure is performed with PACKMOL[58]. Unlike the hydration case, in IL simulations we consider not only the room temperature condition (298 K), but also other temperatures, e.g., 328 K and 348 K. The bulk simulations of ionic solvents follow energy minimizations, 10 ns NPT equilibration and 100 ns NPT production, similar to the hydration case. The bulk densities are extracted from simulation trajectories every 20 ps, based on which the averaged density is computed.

As for solvation free energy calculations in ionic solvents, the nonequilibrium fast-switching procedure is generally similar to the hydration case. The only difference lies in the length of the alchemical switching process. In the solute-in-IL simulations, we extend the vdW coupling time from the previous 0.8 ns to 1.5 ns and the electrostatic coupling time from 0.4 ns to 0.5 ns, leading to a total simulation length of 2 ns. Such an extension is set according to the experience about the poor convergence behavior in IL simulations[54-56].

The ion-pair compositions and the corresponding temperatures included in the mass-density external dataset are listed in Table S6, and their experimental and simulated temperature-density coverage is summarized in Fig. S2C. As for the external testing dataset of solvation free energies, the 2D chemical structures of simulated ILs are shown in Fig. S2D and the solutes under each IL-temperature condition are listed in Tables S4-S5. The distributions of the experimental solvation free energies are summarized in Fig. S2E.

**Table S4.** Solutes in the external dataset for the 298 K IL solvation free energies.

| 298 K [EMIM]$^+$ [TF$_2$N]$^-$ | | 298 K [BMIM]$^+$ [BETI]$^-$ | | 298 K [P$_{66614}$]$^+$ [L-Lact]$^-$ | | 298 K [QUIN$_8$]$^+$ [TF$_2$N]$^-$ | |
|---|---|---|---|---|---|---|---|
| CCCCO | CCc1ccccc1 | C1COCCO1 | CCOCC | CCCCC=C | C[N+]([O-])=O | C1COCCO1 | CCOC(C)=O |
| CCCCCCO | CCCCCCC | CCCCO | CC(C)OC(C)C | CCC[N+]([O-])=O | CCCCCCCCC | CCCCO | CCc1ccccc1 |
| CC1=CCCCC1 | CCCCCC | CCC=C | CCO | CC(C)CC(C)(C)C | CCCCCCCC | CCCCC=C | CCCCCCC |
| CCCCCO | CC(C)c1ccccc1 | CCCCCC#C | C=C | CCCC(C)=O | Cc1ccccc1C | CCC[N+]([O-])=O | CCCCCC |
| CCCO | C | CCCCC=C | CCc1ccccc1 | CCC(=O)CC | Cc1ccc(C)cc1 | CCCO | CO |
| CC(C)CC(C)(C)C | CO | CCCCC#C | CCCCCCC | CC#N | c1ccncc1 | CC(C)CC(C)(C)C | CC1CCCCC1 |
| CCC(C)O | COC(C)(C)C | CCC[N+]([O-])=O | CCCCCC | c1ccccc1 | s1cccc1 | CCC(C)=O | CC1CCCC1 |
| CCC(C)(C)O | Cc1cccc(C)c1 | CCCO | CO | ClC(Cl)Cl | Cc1ccccc1 | CC(C)CO | Cc1cccc(C)c1 |
| CC(C)O | CCCCCCCCC | CC(C)CC(C)(C)C | CC1CCCCC1 | C1CCCCC1 | | CCCC(C)=O | C[N+]([O-])=O |
| CC(C)=O | CCCCCCCC | CCC(C)=O | CC1CCCC1 | CCCCCCCCCC | | CC(C)O | CCCCCCCCC |
| CC#N | Cc1ccccc1C | CC(C)CO | Cc1cccc(C)c1 | ClCCl | | CCC(C)CC | CCCCCCCC |
| c1ccccc1 | Cc1ccc(C)cc1 | CCCC(C)=O | C[N+]([O-])=O | CCOCC | | CC(C)=O | Cc1ccccc1C |
| ClC(Cl)(Cl)Cl | CC(C)(C)O | CC(C)O | CCCCCCCCC | CC(C)OC(C)C | | CC#N | Cc1ccc(C)cc1 |
| ClC(Cl)Cl | CC(C)(C)c1ccccc1 | CCC(C)CC | CCCCCCCC | CCO | | c1ccccc1 | c1ccncc1 |
| C1CCCCC1 | Cc1ccccc1 | CCC(=O)CC | Cc1ccccc1C | CCc1ccccc1 | | ClC(Cl)Cl | C1CCOC1 |
| OC1CCCCC1 | | CC#N | CC=C | CCCCCCC | | C1CCCCC1 | s1cccc1 |
| C1CCC=CC1 | | c1ccccc1 | Cc1ccc(C)cc1 | CCCCCC | | CCCCCCCCCC | Cc1ccccc1 |
| CCCCCCCCCC | | ClC(Cl)Cl | c1ccncc1 | CO | | ClCCl | CCN(CC)CC |
| ClCCl | | C1CCCCC1 | s1cccc1 | CC1CCCCC1 | | CCOCC | |
| CCO | | CCCCCCCCCC | Cc1ccccc1 | CC1CCCC1 | | CC(C)OC(C)C | |
| CCOC(C)=O | | ClCCl | CCN(CC)CC | Cc1cccc(C)c1 | | CCO | |

**Table S5.** Solutes in the external dataset for the higher-than-room-temperature IL solvation free energies.

| 328 K [QUIN8]$^+$ [TF$_2$N]$^-$ | 338 K [QUIN8]$^+$ [TF$_2$N]$^-$ | 328 K [P$_{4442}$]$^+$ [DEP]$^-$ | 348 K [P$_{4442}$]$^+$ [DEP]$^-$ |
|---|---|---|---|

| | | | | | | | |
|---|---|---|---|---|---|---|---|
| C1COCCO1 | CC1CCCCC1 | C1COCCO1 | CC1CCCCC1 | C1COCCO1 | COC(C)(C)C | C1COCCO1 | CO |
| CCCCO | CC1CCCC1 | CCCCO | CC1CCCC1 | CCCCCC#C | Cc1cccc(C)c1 | CCCCCC#C | CC1CCCCC1 |
| CCCCC=C | Cc1cccc(C)c1 | CCCCC=C | Cc1cccc(C)c1 | CCCCC=C | CCCCCCCCC | CCCCC=C | COC(C)(C)C |
| CCC[N+]([O-])=O | C[N+]([O-])=O | CCC[N+]([O-])=O | C[N+]([O-])=O | CCCCC#C | CCCCC | CCCCC#C | Cc1cccc(C)c1 |
| CCCO | CCCCCCCCC | CCCO | CCCCCCCCC | CCC[N+]([O-])=O | CCCCCCCC | CCC[N+]([O-])=O | CCCCCCCCC |
| CC(C)CC(C)(C)C | CCCCCCCC | CC(C)CC(C)(C)C | CCCCCCCC | CCCCCCC=C | Cc1ccccc1C | CCCCCCC=C | CCCCC |
| CCC(C)=O | Cc1ccccc1C | CCC(C)=O | Cc1ccccc1C | CCCCCCC#C | Cc1ccc(C)cc1 | CCCCCCC#C | CCCCCCCC |
| CC(C)CO | Cc1ccc(C)cc1 | CC(C)CO | Cc1ccc(C)cc1 | CCCC=C | c1ccncc1 | CCCC=C | Cc1ccccc1C |
| CCCC(C)=O | c1ccncc1 | CCCC(C)=O | c1ccncc1 | CCCC#C | C1CCOC1 | CCCC#C | Cc1ccc(C)cc1 |
| CC(C)O | C1CCOC1 | CC(C)O | C1CCOC1 | CC(C)CC(C)(C)C | s1cccc1 | CC(C)CC(C)(C)C | c1ccncc1 |
| CCC(C)CC | s1cccc1 | CCC(C)CC | s1cccc1 | CCCC(C)=O | Cc1ccccc1 | CCCC(C)=O | C1CCOC1 |
| CC(C)=O | Cc1ccccc1 | CC(C)=O | Cc1ccccc1 | CCC(C)CC | CCCCCC=C | CCC(C)CC | s1cccc1 |
| CC#N | CCN(CC)CC | CC#N | CCN(CC)CC | CCC(=O)CC | CCOC(C)(C)C | CCC(=O)CC | Cc1ccccc1 |
| c1ccccc1 | | c1ccccc1 | | CC(C)=O | CCC(C)(C)OC | CC(C)=O | CCCCCC=C |
| ClC(Cl)Cl | | ClC(Cl)Cl | | CC#N | CCCOCCC | CC#N | CCOC(C)(C)C |
| C1CCCCC1 | | C1CCCCC1 | | c1ccccc1 | CCCCOCCCC | c1ccccc1 | CCC(C)(C)OC |
| CCCCCCCCCC | | CCCCCCCCCC | | C1CCCCC1 | CCCCC(C)=O | C1CCCCC1 | CCCOCCC |
| ClCCl | | ClCCl | | C1CCC=CC1 | CCCC(=O)CC | C1CCC=CC1 | CCCCOCCCC |
| CCOCC | | CCOCC | | C1CCCC1 | CCC(C)(C)C | C1CCCC1 | CCCCC(C)=O |
| CC(C)OC(C)C | | CC(C)OC(C)C | | CCCCCCCCCC | C1CCCCCC1 | CCCCCCCCCC | CCCC(=O)CC |
| CCO | | CCO | | CCOCC | C1CCCCCCC1 | CCOCC | CCC(C)(C)C |
| CCOC(C)=O | | CCOC(C)=O | | CC(C)OC(C)C | CCCCCCCCC=C | CC(C)OC(C)C | C1CCCCCC1 |
| CCc1ccccc1 | | CCc1ccccc1 | | CCc1ccccc1 | C=Cc1ccccc1 | CCO | C1CCCCCCC1 |
| CCCCCCC | | CCCCCCC | | CCCCCCC | | CCc1ccccc1 | CCCCCCCCC=C |
| CCCCCC | | CCCCCC | | CCCCCC | | CCCCCCC | C=Cc1ccccc1 |
| CO | | CO | | CC1CCCCC1 | | CCCCCC | |

**Table S6.** IL systems considered in the external test set of mass density.

| Ion pair | temperature (K) | Ion pair | temperature (K) |
|---|---|---|---|
| $[EMIM]^+$ $[DCA]^-$ | 298 | $[QUIN_8]^+$ $[TF_2N]^-$ | 343 |
| $[EMIM]^+$ $[TF_2N]^-$ | 298 | $[BMPyrr]^+$ $[TF_2N]^-$ | 288 |
| $[EMIM]^+$ $[OTF]^-$ | 298 | $[BMPyrr]^+$ $[TF_2N]^-$ | 298 |
| $[BMIM]^+$ $[TF_2N]^-$ | 298 | $[BMPyrr]^+$ $[TF_2N]^-$ | 308 |
| $[BMIM]^+$ $[PF_6]^-$ | 298 | $[BMPyrr]^+$ $[TF_2N]^-$ | 318 |
| $[HMIM]^+$ $[TF_2N]^-$ | 298 | $[BMPyrr]^+$ $[TF_2N]^-$ | 328 |
| $[P_{66614}]^+$ $[L\text{-}Lact]^-$ | 298 | $[BMPyrr]^+$ $[TF_2N]^-$ | 338 |
| $[P_{66614}]^+$ $[CS]^-$ | 298 | $[BMPyrr]^+$ $[TF_2N]^-$ | 348 |
| $[BMIM]^+$ $[BETI]^-$ | 298 | $[OMPyrr]^+$ $[TF_2N]^-$ | 298 |
| $[HMIM]^+$ $[FAP]^-$ | 298 | $[OMPyrr]^+$ $[TF_2N]^-$ | 308 |
| $[QUIN_6]^+$ $[TF_2N]^-$ | 303 | $[OMPyrr]^+$ $[TF_2N]^-$ | 318 |
| $[QUIN_6]^+$ $[TF_2N]^-$ | 323 | $[OMPyrr]^+$ $[TF_2N]^-$ | 328 |
| $[QUIN_6]^+$ $[TF_2N]^-$ | 343 | $[OMPyrr]^+$ $[TF_2N]^-$ | 338 |
| $[QUIN_8]^+$ $[TF_2N]^-$ | 303 | $[OMPyrr]^+$ $[TF_2N]^-$ | 348 |
| $[QUIN_8]^+$ $[TF_2N]^-$ | 323 | | |

## S6. Detailed benchmark results.

### S6.1. Chemical grounding benchmark.

We evaluated chemical grounding using a benchmark of 50 commonly used abbreviations for IL-forming ions (Tables S7-S8). Outputs were manually assessed by human experts under three increasingly lenient criteria: (i) exact canonical-SMILES recovery, requiring an exact match to the reference string; (ii) structure-level SMILES equivalence, accepting alternative SMILES that yielded the same molecular structure after manual canonicalization; and (iii) correct full-name identification, requiring either a valid systematic name or a commonly accepted trivial name for the ion. GPT-5 achieved the highest overall accuracy among the evaluated web-augmented LLMs and was therefore selected as the planner of AIonopedia. Integrating domain-specific modules further improved performance under the strictest criterion, demonstrating the importance of specialized chemical grounding for reliable downstream reasoning.

The benchmark compared leading models available at the time of evaluation, including GPT-5, Gemini 2.5 Pro, Claude Sonnet 4.5, and DeepSeek-R1-0528. The first three represented the then-current flagship models from OpenAI, Google, and Anthropic, respectively. Although Opus 4.1 was available, Sonnet 4.5 was selected because Anthropic positioned it as its most powerful model at the time (2025.9); Opus 4.5 had not yet been released. For DeepSeek, we selected R1-0528, the latest dedicated R1 checkpoint, rather than newer V3.x models to retain a fixed and reasoning-focused baseline.

**Table S7.** The model effect on the abbreviation interpretation for ions in ILs.

| Accuracy (↑) | Canonical SMILES Match | Structural Match | Full Name Match |
|---|---|---|---|
| AIonopedia | 0.920±0.028 | 0.960±0.028 | 0.993±0.009 |
| GPT-5[27] | 0.280±0.035 | 0.927±0.012 | 0.940±0.016 |
| Gemini 2.5 pro[59] | 0.207±0.012 | 0.467±0.046 | 0.533±0.050 |
| Claude 4.5 sonnet[60] | 0.100±0.035 | 0.286±0.012 | 0.433±0.009 |
| DeepSeek-R1[61,62] | 0.147±0.050 | 0.447±0.050 | 0.480±0.049 |
| GPT-5(w/o web) | 0.287±0.046 | 0.873±0.070 | 0.920±0.033 |

**Table S8.** The full name-abbreviation correspondence of 50 ions used for the evaluation of conversion between full names, abbreviations and SMILES.

| Ion Abbreviation | Full Name | Ion Abbreviation | Full Name |
|---|---|---|---|
| EMIm | 1-ethyl-3-methylimidazolium | Nip311 | Propyl(dimethyl)isopropylammonium |
| HMIm | 1-hexyl-3-Methylimidazolium | EtOHMIm | 1-ethanol-3-methylimidazolium |
| AMIm | 1-allyl-3-methylimidazolium | C4C1C1Im | 1-butyl-2,3-dimethylimidazolium |
| HEMIm | 1-ethanol-3-methylimidazolium | C22O1Im | 1-ethyl-3-(2-methoxyethyl)imidazolium |
| P4442 | Tributylethylphosphonium | MTBDH | 1,3,4,6,7,8-hexahydro-1-methyl-2H-pyrimido[1,2-a]pyrimidine / 7-methyl-1,5,7-triazabicyclo[4.4.0]dec-5-enium |
| P66614 | Trihexyltetradecylphosphonium | BMMorp | 1-butyl-1-methylmorpholinium |
| PMPyrr | 1-propyl-1-methylpyrrolidinium | 3-MeBuPy | 3-methyl-N-butyl-pyridinium |
| BMPyrr | 1-butyl-1-methylpyrrolidinium | MOEPy | N-(2-methoxyethyl)pyridinium |
| C5MPyr | 1-pentyl-1-methylpyrrolidinium | MeoeM2EAm | 2-methoxyethyl-(dimethyl)ethylammonium |
| C6MPyr | 1-hexyl-1-methylpyrrolidinium | DCA | Dicyanamide |
| C10MPyr | 1-decyl-1-methylpyrrolidinium | Trif | Trifluoromethylsulfonate |
| ChxmPyrr | 1-cyclohexylmethyl-1-methylpyrrolidinium | l-Lac | L-Lactate |
| MeoeMPyrr | 1-(2-methoxyethyl)-1-methylpyrrolidinium | PF6 | Hexafluorophosphate |
| MeoeMMorp | 1-(2-methoxyethyl)-1-methylmorpholinium | FAP | Tris(pentafluoroethyl)trifluorophosphate |
| PMPip | 1-Propyl-1-methylpiperidinium | TF2N | Bis(trifluoromethylsulfonyl)imide |
| BMPip | 1-Butyl-1-methylpiperidinium | BETI | Bis(pentafluoroethylsulfonyl)imide |
| MeoeMPip | 1-(2-methoxyethyl)-1-methylpiperidinium | DEP | Diethylphosphate |
| BzPy | 1-benzylpyridinium | TCB | tetracyanoborate |
| TDC | 1,2,3-tris(diethylamino)-cyclopropenium | SCN | thiocyanate |
| N4441 | Methyl(tributyl)ammonium | F3Ac | trifluoroacetate |
| C6TA | hexyl(trimethyl)ammonium | FSI | bis(fluorosulfonyl)imide |
| C8TA | Octyl(trimethyl)ammonium | EtSO4 | ethylsulfate |
| C10TA | Decyl(trimethyl)ammonium | BF4 | tetrafluoroborate |
| N2228 | octyltriethylammonium | TCM | tricyanomethanide |
| N8888 | Tetraoctylammonium | TDI | 4,5-dicyano-2-(trifluoromethyl)imidazolide |

## S6.2. Cross-validation benchmarks for property prediction.

Table S9-S11 report the complete numerical results underlying the model rankings, split-wise

comparisons, and strict win rates shown in Fig. 2B-D. As described in Section S3.1, AIonopedia was instantiated with Galactica, Qwen3, and Gemma3 language backbones to examine the applicability of different pretrained LLM families to IL representation learning.

The baseline suite covered both encoder- and decoder-based architectures and both unimodal and multimodal molecular representations. To ensure a fair comparison of optimization scale, most LLM-based baselines were finetuned with LoRA using their default GitHub repository settings. For PRESTO, however, the default LoRA rank was reduced to match that used in LlaSMol, avoiding a disproportionately large trainable-parameter budget relative to the other methods. The lightweight BERT[63] and RoBERTa[64] encoder-based models SPMM and ILBERT were instead fully fine-tuned. We additionally included an MLP trained directly on the physicochemical descriptors used for auxiliary supervision during modality alignment. This control was used to determine whether the performance gains of AIonopedia could be attributed solely to descriptor-based supervision.

Across the complete evaluations, AIonopedia achieved the best performance on nearly all datasets and split schemes. The Qwen3-0.6b variant ranked first in 18 comparisons, particularly across several solute-solvent thermodynamic datasets, and was therefore selected as the default model. Among the similarly sized AIonopedia variants, Gemma3-1b showed the weakest overall performance, consistent with a possible benefit from the science-oriented pretraining reported for Galactica and Qwen3. Through LoRA adaptation, AIonopedia also retained a trainable-parameter count comparable to fully fine-tuned encoder-based models and lower than several baselines built on 7~8 billion-parameter backbones.

Among the baseline methods, ILBERT was the strongest overall competitor and performed particularly well on melting-point prediction, potentially benefiting from its pretraining on ionic compounds. Nevertheless, its relative performance declined substantially on viscosity prediction. A similar but more pronounced trend was observed for SPMM, whose viscosity RMSE was nearly twice that of most other baselines. As discussed in Fig. 4, viscosity is highly sensitive to molecular size and chain length. Such dependence is likely more difficult to capture with encoder-style models using 1-D CNN aggregation or a [CLS] token than with decoder-style LLMs that read the final token. Still, this trend may also reflect differences in architecture and pretraining.

The error bars in Fig. 2C denote standard deviations across the five cross-validation folds. For the solute-solvent interaction datasets, the vast combinatorial space of solute-cation-anion systems constrains experimental coverage of individual ions. Consequently, the stringent cation-based split can yield uneven cation allocation across folds and thus relatively large variance. Therefore, overlap between error bars should not, by itself, be interpreted as evidence of statistically equivalent performance. When comparisons were aggregated across all datasets, metrics, and split schemes, the default AIonopedia model maintained strict win rates above 85% against every baseline (Fig. 2D). To facilitate comparison with previous studies, which commonly report Average Absolute Relative Deviation (AARD) as a relative-error metric, Table S12 additionally provides the average absolute relative deviation for viscosity under the cation- and IL-based splits alongside RMSE.

**Table S9.** Comparison of AIonopedia with baselines under cation-based splits; the organic solvent/water transfer ΔG dataset uses a solvent-based split.

| RMSE (↓) | solvation ΔG (kcal/mol) | IL/water transfer ΔG (kcal/mol) | melting point (K) | viscosity (mPa·s) (log scale) | surface tension (mN/m) | mass density (g/cm$^3$) | organic solvent/water transfer ΔG (kcal/mol) |
|---|---|---|---|---|---|---|---|
| AIonopedia (Galactica-125m) | 0.364±0.098 | 0.483±0.123 | 40.4±3.9 | **0.294±0.036 (0.2939)** | 3.86±0.41 | **0.0327±0.0032** | 0.530±0.039 |
| AIonopedia (Galactica-1.3b) | **0.322±0.136** | 0.457±0.108 | 40.3±3.8 | 0.298±0.038 | 3.77±0.24 | 0.0330±0.0047 | **0.529±0.051** |
| AIonopedia (Qwen3-0.6b) | 0.328±0.130 | **0.441±0.117** | 39.9±3.4 | 0.294±0.031 (0.2943) | **3.62±0.35** | 0.0333±0.0068 | 0.534±0.038 |
| AIonopedia (Gemma3-1b) | 0.325±0.136 | 0.459±0.095 | 40.6±3.2 | 0.305±0.026 | 3.66±0.33 | 0.0330±0.0060 | 0.538±0.061 |
| MLP (pretrain descriptors) | 0.417±0.106 | 0.755±0.093 | 45.5±5.3 | 0.411±0.048 | 4.48±0.48 | 0.0437±0.0067 | 0.878±0.052 |
| MolCA[65] (Galactica-1.3b) | 0.421±0.078 | 0.732±0.176 | 46.9±1.9 | 0.384±0.049 | 5.81±0.47 | 0.0525±0.0064 | 0.846±0.222 |
| T5chem[66,67] | 0.379±0.104 | 0.576±0.104 | 44.9±2.5 | 0.317±0.038 | 3.94±0.23 | 0.0357±0.0043 | 0.702±0.067 |
| Molinst (Llama3-8b)[68,69] | 0.352±0.098 | 0.510±0.159 | 42.4±3.2 | 0.310±0.036 | 4.08±0.43 | 0.0331±0.0031 | 0.678±0.076 |
| SPMM (full finetune)[70] | 0.445±0.090 | 0.531±0.118 | 46.4±3.9 | 0.588±0.018 | 4.67±0.36 | 0.0488±0.0053 | 0.670±0.098 |
| LlaSMol (Mistral-7b)[71,72] | 0.337±0.133 | 0.492±0.150 | 42.6±4.9 | 0.310±0.019 | 3.85±0.27 | 0.0340±0.0027 | 0.674±0.087 |
| PRESTO (Vicuna v1.5-7b)[73,74] | 0.400±0.144 | 0.472±0.120 | 41.2±4.6 | 0.309±0.031 | 3.97±0.46 | 0.0442±0.0046 | 0.723±0.124 |
| ILBERT (full finetune)[3] | 0.334±0.105 | 0.459±0.169 | **39.7±4.3** | 0.318±0.036 | 3.96±0.28 | 0.0377±0.0067 | 0.636±0.031 |

| Pearson r (↑) | solvation ΔG | IL/water transfer ΔG | melting point | viscosity (log scale) | surface tension | mass density | organic solvent/water transfer ΔG |
|---|---|---|---|---|---|---|---|
| AIonopedia (Galactica-125m) | 0.9540±0.0213 | 0.9798±0.0103 | 0.7186±0.0305 | **0.9098±0.0198** | 0.9123±0.0120 | **0.9845±0.0038** | **0.9754±0.0044** |
| AIonopedia (Galactica-1.3b) | **0.9592±0.0300** | 0.9819±0.0081 | 0.7201±0.0321 | 0.9065±0.0211 | 0.9123±0.0109 | 0.9839±0.0051 | 0.9751±0.0059 |
| AIonopedia (Qwen3-0.6b) | 0.9563±0.0279 | **0.9837±0.0080** | 0.7191±0.0283 | 0.9095±0.0146 | **0.9187±0.0097** | 0.9835±0.0065 | 0.9742±0.0050 |
| AIonopedia (Gemma3-1b) | 0.9581±0.0279 | 0.9830±0.0066 | 0.7113±0.0178 | 0.9022±0.0033 | 0.9162±0.0083 | 0.9840±0.0056 | 0.9734±0.0067 |
| MLP | 0.9253±0.0326 | 0.9462±0.0061 | 0.6354±0.0584 | 0.8109±0.0313 | 0.8686±0.0168 | 0.9718±0.0078 | 0.9282±0.0129 |
| MolCA | 0.9319±0.0174 | 0.9488±0.0192 | 0.5977±0.0521 | 0.8579±0.0164 | 0.7815±0.0436 | 0.9598±0.0079 | 0.9411±0.0219 |
| T5chem | 0.9510±0.0251 | 0.9813±0.0088 | 0.6824±0.0279 | 0.9023±0.0189 | 0.9059±0.0111 | 0.9830±0.0043 | 0.9632±0.0104 |
| Molinst | 0.9510±0.0204 | 0.9757±0.0136 | 0.6697±0.0469 | 0.8978±0.0179 | 0.8917±0.0251 | 0.9839±0.0031 | 0.9559±0.0123 |
| SPMM | 0.9230±0.0284 | 0.9773±0.0102 | 0.6165±0.0404 | 0.5746±0.0527 | 0.8683±0.0086 | 0.9659±0.0077 | 0.9584±0.0168 |
| LlaSMol | 0.9531±0.0284 | 0.9771±0.0117 | 0.6654±0.0395 | 0.8971±0.0150 | 0.9026±0.0187 | 0.9830±0.0035 | 0.9575±0.0126 |
| PRESTO | 0.9366±0.0348 | 0.9790±0.0090 | 0.6969±0.0482 | 0.8984±0.0072 | 0.8996±0.0128 | 0.9709±0.0055 | 0.9493±0.0193 |
| ILBERT | 0.9517±0.0277 | 0.9797±0.0136 | **0.7204±0.0381** | 0.8927±0.0127 | 0.9015±0.0177 | 0.9793±0.0062 | 0.9629±0.0060 |

| Kendall τ (↑) | solvation ΔG | IL/water transfer ΔG | melting point | viscosity (log scale) | surface tension | mass density | organic solvent/water transfer ΔG |
|---|---|---|---|---|---|---|---|
| AIonopedia (Galactica-125m) | 0.843±0.027 | 0.883±0.016 | **0.515±0.029** | **0.803±0.022** | 0.778±0.020 | 0.911±0.010 | 0.878±0.014 |
| AIonopedia (Galactica-1.3b) | **0.865±0.043** | 0.890±0.016 | 0.508±0.046 | 0.800±0.025 | 0.772±0.031 | **0.912±0.015** | **0.881±0.016** |
| AIonopedia (Qwen3-0.6b) | 0.860±0.046 | **0.893±0.011** | 0.505±0.022 | 0.800±0.021 | **0.780±0.022** | 0.908±0.015 | 0.879±0.012 |
| AIonopedia (Gemma3-1b) | 0.864±0.041 | 0.892±0.014 | 0.500±0.028 | 0.795±0.010 | 0.776±0.022 | 0.911±0.010 | 0.877±0.019 |
| MLP | 0.781±0.041 | 0.793±0.016 | 0.466±0.027 | 0.683±0.030 | 0.706±0.016 | 0.870±0.017 | 0.768±0.018 |
| MolCA | 0.802±0.017 | 0.821±0.027 | 0.395±0.056 | 0.710±0.021 | 0.624±0.039 | 0.843±0.023 | 0.807±0.033 |
| T5chem | 0.843±0.030 | 0.885±0.011 | 0.471±0.035 | 0.783±0.020 | 0.746±0.027 | 0.907±0.005 | 0.849±0.017 |
| Molinst | 0.853±0.030 | 0.874±0.013 | 0.460±0.055 | 0.775±0.022 | 0.740±0.036 | 0.904±0.013 | 0.835±0.016 |
| SPMM | 0.778±0.033 | 0.875±0.021 | 0.400±0.050 | 0.410±0.049 | 0.692±0.020 | 0.853±0.029 | 0.845±0.025 |
| LlaSMol | 0.854±0.049 | 0.868±0.018 | 0.452±0.049 | 0.774±0.014 | 0.759±0.020 | 0.903±0.023 | 0.833±0.020 |
| PRESTO | 0.825±0.057 | 0.877±0.015 | 0.484±0.051 | 0.775±0.014 | 0.749±0.012 | 0.874±0.015 | 0.825±0.029 |
| ILBERT | 0.844±0.047 | 0.883±0.019 | 0.514±0.042 | 0.770±0.021 | 0.735±0.034 | 0.889±0.008 | 0.850±0.010 |

**Table S10.** Comparison of AIonopedia with baselines under IL-based and ternary-component-based splits on the solute-solvent interaction datasets; the organic solvent/water transfer ΔG dataset uses a solute-solvent-based split.

| RMSE (↓) | solvation ΔG (kcal/mol) | IL/water transfer ΔG (kcal/mol) | solvation ΔG (ternary-component split) (kcal/mol) | IL/water transfer ΔG (ternary-component split) (kcal/mol) | organic solvent/water transfer ΔG (kcal/mol) |
|---|---|---|---|---|---|
| AIonopedia (Galactica-125m) | **0.304±0.030** | 0.473±0.100 | 0.155±0.013 | 0.373±0.043 | 0.415±0.028 |
| AIonopedia (Galactica-1.3b) | 0.321±0.029 | 0.460±0.121 | 0.126±0.012 | 0.265±0.029 | 0.399±0.030 |
| AIonopedia (Qwen3-0.6b) | 0.309±0.044 | **0.440±0.132** | **0.124±0.016** | **0.238±0.013** | **0.393±0.035 (0.3930)** |
| AIonopedia (Gemma3-1b) | 0.311±0.039 | 0.459±0.121 | 0.125±0.017 | 0.250±0.013 | 0.393±0.032 (0.3934) |
| MLP | 0.441±0.065 | 0.880±0.249 | 0.361±0.023 | 0.742±0.043 | 0.818±0.035 |
| MolCA | 0.400±0.073 | 0.630±0.051 | 0.341±0.021 | 0.644±0.056 | 0.724±0.012 |
| T5chem | 0.371±0.027 | 0.588±0.137 | 0.240±0.013 | 0.507±0.041 | 0.560±0.019 |
| Molinst | 0.352±0.038 | 0.510±0.124 | 0.178±0.019 | 0.275±0.005 | 0.468±0.032 |
| SPMM | 0.425±0.016 | 0.550±0.111 | 0.307±0.023 | 0.427±0.082 | 0.575±0.053 |
| LlaSMol | 0.315±0.037 | 0.591±0.132 | 0.160±0.016 | 0.336±0.036 | 0.513±0.033 |
| PRESTO | 0.413±0.144 | 0.530±0.088 | 0.160±0.032 | 0.259±0.022 | 0.541±0.089 |
| ILBERT | 0.323±0.051 | 0.452±0.124 | 0.147±0.024 | 0.260±0.030 | 0.404±0.042 |

| Pearson r (↑) | solvation ΔG | IL/water transfer ΔG | solvation ΔG (ternary-component split) | IL/water transfer ΔG (ternary-component split) | organic solvent/water transfer ΔG |
|---|---|---|---|---|---|
| AIonopedia (Galactica-125m) | **0.9625±0.0056** | 0.9801±0.0104 | 0.9904±0.0021 | 0.9880±0.0028 | 0.9849±0.0025 |
| AIonopedia (Galactica-1.3b) | 0.9601±0.0051 | 0.9807±0.0116 | 0.9939±0.0014 | 0.9944±0.0011 | 0.9861±0.0021 |
| AIonopedia (Qwen3-0.6b) | 0.9615±0.0099 | **0.9816±0.0113** | **0.9942±0.0018** | **0.9953±0.0006** | **0.9862±0.0024** |
| AIonopedia (Gemma3-1b) | 0.9625±0.0076 | 0.9803±0.0118 | 0.9940±0.0018 | 0.9950±0.0007 | 0.9860±0.0023 |
| MLP | 0.9185±0.0237 | 0.9200±0.0490 | 0.9444±0.0085 | 0.9465±0.0062 | 0.9360±0.0113 |
| MolCA | 0.9340±0.0203 | 0.9626±0.0062 | 0.9527±0.0056 | 0.9616±0.0069 | 0.9529±0.0038 |
| T5chem | 0.9512±0.0060 | 0.9785±0.0123 | 0.9815±0.0030 | 0.9874±0.0023 | 0.9786±0.0028 |
| Molinst | 0.9504±0.0082 | 0.9763±0.0121 | 0.9868±0.0027 | 0.9929±0.0002 | 0.9798±0.0030 |
| SPMM | 0.9272±0.0040 | 0.9789±0.0085 | 0.9633±0.0046 | 0.9892±0.0005 | 0.9713±0.0044 |
| LlaSMol | 0.9605±0.0091 | 0.9681±0.0158 | 0.9893±0.0023 | 0.9892±0.0022 | 0.9755±0.0044 |
| PRESTO | 0.9349±0.0371 | 0.9743±0.0095 | 0.9890±0.0043 | 0.9936±0.0009 | 0.9721±0.0092 |
| ILBERT | 0.9585±0.0108 | 0.9801±0.0110 | 0.9909±0.0030 | 0.9935±0.0015 | 0.9849±0.0028 |

| Kendall τ (↑) | solvation ΔG | IL/water transfer ΔG | solvation ΔG (ternary-component split) | IL/water transfer ΔG (ternary-component split) | organic solvent/water transfer ΔG |
|---|---|---|---|---|---|
| AIonopedia (Galactica-125m) | 0.870±0.019 | 0.883±0.016 | 0.944±0.002 | 0.910±0.008 | 0.906±0.005 |
| AIonopedia (Galactica-1.3b) | 0.863±0.019 | 0.888±0.018 | 0.956±0.002 | 0.938±0.005 | 0.914±0.004 |
| AIonopedia (Qwen3-0.6b) | 0.866±0.031 | **0.895±0.020** | 0.962±0.002 | **0.944±0.002** | **0.916±0.005** |
| AIonopedia (Gemma3-1b) | **0.870±0.014** | 0.885±0.020 | **0.963±0.002** | 0.943±0.003 | 0.915±0.006 |
| MLP | 0.788±0.017 | 0.772±0.040 | 0.816±0.009 | 0.798±0.011 | 0.780±0.015 |
| MolCA | 0.804±0.031 | 0.849±0.010 | 0.829±0.007 | 0.847±0.010 | 0.828±0.007 |
| T5chem | 0.848±0.016 | 0.880±0.018 | 0.919±0.003 | 0.907±0.007 | 0.884±0.004 |
| Molinst | 0.847±0.022 | 0.874±0.015 | 0.946±0.003 | 0.931±0.003 | 0.890±0.009 |
| SPMM | 0.780±0.015 | 0.878±0.022 | 0.847±0.006 | 0.913±0.002 | 0.897±0.005 |
| LlaSMol | 0.860±0.017 | 0.851±0.027 | 0.941±0.006 | 0.908±0.010 | 0.877±0.013 |
| PRESTO | 0.830±0.057 | 0.867±0.017 | 0.956±0.003 | 0.938±0.003 | 0.870±0.023 |
| ILBERT | 0.861±0.016 | 0.887±0.020 | 0.947±0.002 | 0.936±0.008 | 0.911±0.007 |

**Table S11.** Comparison of AIonopedia with baselines under IL-based splits on the bulk property datasets.

| RMSE (↓) | melting point (K) | viscosity (mPa·s) (log scale) | surface tension (mN/m) | mass density (g/cm$^3$) |
|---|---|---|---|---|
| AIonopedia (Galactica-125m) | 38.4±3.0 | 0.249±0.018 (0.2493) | 3.63±0.27 | 0.0272±0.0029 |
| AIonopedia (Galactica-1.3b) | 38.5±2.5 | **0.249±0.020 (0.2492)** | **3.42±0.27** | 0.0267±0.0030 |
| AIonopedia (Qwen3-0.6b) | 38.9±3.5 | 0.250±0.017 | 3.47±0.35 | 0.0266±0.0028 |
| AIonopedia (Gemma3-1b) | 39.6±3.0 | 0.251±0.020 | 3.44±0.32 | **0.0264±0.0032** |
| MLP | 44.4±3.6 | 0.368±0.022 | 4.33±0.31 | 0.0427±0.0036 |
| MolCA | 48.2±4.0 | 0.332±0.011 | 5.26±0.25 | 0.0499±0.0028 |
| T5chem | 43.4±3.2 | 0.288±0.025 | 3.77±0.28 | 0.0323±0.0013 |
| Molinst | 39.7±3.0 | 0.268±0.018 | 3.50±0.44 | 0.0280±0.0024 |
| SPMM | 45.5±3.3 | 0.545±0.023 | 4.59±0.37 | 0.0417±0.0026 |
| LlaSMol | 42.1±3.2 | 0.267±0.016 | 3.59±0.13 | 0.0289±0.0021 |
| PRESTO | 42.4±2.2 | 0.268±0.019 | 3.63±0.34 | 0.0468±0.0045 |
| ILBERT | **37.2±3.2** | 0.284±0.021 | 3.51±0.20 | 0.0323±0.0028 |

| Pearson r (↑) | melting point | viscosity (log scale) | surface tension | mass density |
|---|---|---|---|---|
| AIonopedia (Galactica-125m) | 0.7470±0.0380 | **0.9393±0.0080** | 0.9194±0.0160 | 0.9895±0.0021 |
| AIonopedia (Galactica-1.3b) | 0.7476±0.0471 | 0.9392±0.0105 | **0.9272±0.0127** | 0.9898±0.0021 |
| AIonopedia (Qwen3-0.6b) | 0.7408±0.0613 | 0.9389±0.0093 | 0.9242±0.0184 | 0.9897±0.0021 |
| AIonopedia (Gemma3-1b) | 0.7289±0.0533 | 0.9378±0.0104 | 0.9261±0.0162 | **0.9899±0.0024** |
| MLP | 0.6537±0.0473 | 0.8575±0.0192 | 0.8790±0.0199 | 0.9727±0.0060 |
| MolCA | 0.5728±0.0392 | 0.8949±0.0079 | 0.8187±0.0321 | 0.9641±0.0058 |
| T5chem | 0.7202±0.0401 | 0.9279±0.0103 | 0.9185±0.0131 | 0.9862±0.0022 |
| Molinst | 0.7254±0.0370 | 0.9265±0.0108 | 0.9210±0.0199 | 0.9885±0.0020 |
| SPMM | 0.6435±0.0419 | 0.6593±0.0282 | 0.8742±0.0214 | 0.9761±0.0019 |
| LlaSMol | 0.6767±0.0567 | 0.9271±0.0101 | 0.9182±0.0100 | 0.9877±0.0022 |
| PRESTO | 0.6806±0.0214 | 0.9270±0.0118 | 0.9156±0.0183 | 0.9669±0.0099 |
| ILBERT | **0.7620±0.0291** | 0.9195±0.0108 | 0.9214±0.0101 | 0.9849±0.0023 |

| Kendall τ (↑) | melting point | viscosity (log scale) | surface tension | mass density |
|---|---|---|---|---|
| AIonopedia (Galactica-125m) | 0.546±0.031 | 0.834±0.015 | 0.800±0.021 | 0.931±0.004 |
| AIonopedia (Galactica-1.3b) | 0.543±0.051 | **0.838±0.016** | **0.807±0.014** | 0.933±0.004 |
| AIonopedia (Qwen3-0.6b) | 0.548±0.048 | 0.835±0.017 | 0.804±0.022 | **0.934±0.005 (0.9337)** |
| AIonopedia (Gemma3-1b) | 0.536±0.049 | 0.833±0.015 | 0.807±0.021 | 0.934±0.005 (0.9336) |
| MLP | 0.472±0.046 | 0.708±0.018 | 0.721±0.018 | 0.876±0.009 |
| MolCA | 0.373±0.032 | 0.754±0.012 | 0.662±0.021 | 0.863±0.013 |
| T5chem | 0.516±0.046 | 0.815±0.012 | 0.781±0.018 | 0.918±0.005 |
| Molinst | 0.524±0.035 | 0.812±0.015 | 0.791±0.021 | 0.928±0.003 |
| SPMM | 0.440±0.049 | 0.462±0.022 | 0.720±0.010 | 0.884±0.004 |
| LlaSMol | 0.475±0.064 | 0.809±0.015 | 0.787±0.022 | 0.924±0.004 |
| PRESTO | 0.488±0.023 | 0.814±0.023 | 0.792±0.011 | 0.889±0.015 |
| ILBERT | **0.568±0.025** | 0.801±0.015 | 0.772±0.019 | 0.910±0.004 |

**Table S12.** AARD of viscosity

| AARD(↓) | viscosity (cation split) | viscosity (IL split) |
|---|---|---|
| AIonopedia (Galactica-125m) | 0.150±0.053 | 0.113±0.021 |
| AIonopedia (Galactica-1.3b) | **0.147±0.054** | **0.112±0.019** |
| AIonopedia (Qwen3-0.6b) | 0.153±0.077 | 0.116±0.020 |
| AIonopedia (Gemma3-1b) | 0.176±0.114 | 0.115±0.025 |
| MLP (pretrain descriptors) | 0.223±0.068 | 0.198±0.030 |
| MolCA (Galactica-1.3b) | 0.255±0.171 | 0.175±0.025 |
| T5chem | 0.153±0.044 | 0.129±0.014 |
| Molinst (Llama3-8b) | 0.195±0.120 | 0.144±0.034 |
| SPMM (full finetune) | 0.400±0.185 | 0.330±0.035 |
| LlaSMol (Mistral-7b) | 0.195±0.116 | 0.142±0.031 |
| PRESTO (Vicuna v1.5-7b) | 0.187±0.113 | 0.138±0.037 |
| ILBERT (full finetune) | 0.179±0.094 | 0.143±0.025 |

### S6.3. System-specific OOD and additional benchmarks.

Building on the cross-validation benchmarks, we further evaluated AIonopedia on five chemically diverse IL systems with temperature-dependent solvation free energy measurements (Fig. 2E and Table S13). For each system, the complete ion pair was excluded from training, providing a system-specific OOD evaluation against GROMACS-based MD simulations (Section S5) and other deep-learning models.

Overall, MD simulations yielded chemically reasonable and relatively stable errors across the five systems (RMSE = 0.524~0.897 kcal $mol^{-1}$), whereas the performance of learning-based methods varied more strongly with the chemical composition of each system. AIonopedia achieved the lowest errors in most comparisons (RMSE = 0.060~0.464 kcal $mol^{-1}$). The easiest case was $[QUIN_8]^+[TF_2N]^-$, whose ion pair was excluded from training while both constituent ions were well represented individually, enabling near-perfect predictions by almost all learning-based models. Similar trends were observed for $[BMIM]^+[BETI]^-$ and $[EMIM]^+[TF_2N]^-$, although several baselines showed reduced accuracy as the corresponding ions were represented less frequently.

The largest performance differences were observed for the phosphonium-based systems $[P_{66614}]^+$[L-Lact]$^-$ and $[P_{4442}]^+[DEP]^-$, for which one or both ions were absent from the training data. Several baseline models exhibited pronounced performance degradation on these challenging cases, while AIonopedia retained accurate predictions. In particular, ILBERT, despite being the strongest overall baseline in the cross-validation benchmarks, exceeded the commonly accepted threshold of chemical accuracy for $[P_{66614}]^+$[L-Lact]$^-$, with an RMSE above 1 kcal mol$^{-1}$. These results demonstrate the stronger generalization of AIonopedia to previously unseen ionic systems.

Additional benchmarks were conducted for mass density and hydration free energy using the protocols described in Sections S3.3 and S5. AIonopedia achieved the best or comparable performance among all evaluated approaches on both tasks (Table S14), further supporting its applicability across diverse IL-related thermodynamic properties.

**Table S13.** Comparison of AIonopedia with computational baseline and prior baselines on solvation ΔG datasets for fixed systems.

| RMSE(kcal/mol) (↓) | 298K [BMIM]$^+$ [BETI]$^-$ | 298K [EMIM]$^+$ [TF$_2$N]$^-$ | 298K $[P_{66614}]^+$ [L-Lact]$^-$ | 298K [QUIN$_8$]$^+$ [TF$_2$N]$^-$ | 328K $[P_{4442}]^+$ [DEP]$^-$ | 328K [QUIN$_8$]$^+$ [TF$_2$N]$^-$ | 338K [QUIN$_8$]$^+$ [TF$_2$N]$^-$ | 348K $[P_{4442}]^+$ [DEP]$^-$ |
|---|---|---|---|---|---|---|---|---|
| AIonopedia (Galactica-125m) | 0.387±0.018 | 0.259±0.005 | 0.320±0.091 | 0.088±0.007 | 0.441±0.042 | 0.073±0.008 | 0.075±0.006 | 0.512±0.029 |
| AIonopedia (Galactica-1.3b) | 0.380±0.030 | **0.220±0.005** | 0.350±0.045 | **0.060±0.003** | 0.369±0.058 | **0.051±0.003** | **0.051±0.003** | **0.457±0.047** |
| AIonopedia (Qwen3-0.6b) | 0.438±0.032 | 0.242±0.023 | **0.281±0.094** | 0.064±0.009 | **0.361±0.053** | 0.060±0.007 | 0.062±0.008 | 0.464±0.029 |
| AIonopedia (Gemma3-1b) | 0.405±0.008 | 0.230±0.022 | 0.305±0.050 | 0.062±0.011 | 0.433±0.060 | 0.056±0.008 | 0.054±0.010 | 0.493±0.047 |
| MD simulation[44] | 0.806±0.052 | 0.524±0.042 | 0.731±0.052 | 0.682±0.068 | 0.697±0.019 | 0.897±0.045 | 0.597±0.055 | 0.547±0.012 |
| MLP | 0.558±0.024 | 0.405±0.014 | 0.573±0.087 | 0.226±0.016 | 0.451±0.046 | 0.220±0.007 | 0.220±0.007 | 0.564±0.018 |
| MolCA | **0.280±0.017** | 0.405±0.125 | 0.769±0.039 | 0.372±0.021 | 0.534±0.023 | 0.340±0.025 | 0.336±0.026 | 0.656±0.024 |
| T5chem | 0.327±0.017 | 0.281±0.022 | 0.429±0.061 | 0.132±0.013 | 0.639±0.024 | 0.125±0.008 | 0.123±0.006 | 0.750±0.029 |
| Molinst | 0.341±0.038 | 0.227±0.039 | 0.705±0.035 | 0.068±0.013 | 0.443±0.136 | 0.063±0.010 | 0.064±0.008 | 0.581±0.139 |
| SPMM | 0.358±0.012 | 0.445±0.050 | 0.449±0.126 | 0.309±0.030 | 0.365±0.032 | 0.159±0.028 | 0.283±0.032 | 0.581±0.045 |
| LlaSMol | 0.341±0.037 | 0.277±0.085 | 0.412±0.133 | 0.077±0.014 | 0.648±0.058 | 0.069±0.014 | 0.071±0.016 | 0.746±0.061 |
| PRESTO | 0.422±0.021 | 0.809±0.024 | 0.531±0.050 | 0.075±0.006 | 0.469±0.072 | 0.065±0.009 | 0.064±0.008 | 0.562±0.074 |
| ILBERT | 0.355±0.035 | 0.261±0.048 | 1.210±0.391 | 0.086±0.009 | 0.532±0.087 | 0.062±0.007 | 0.058±0.008 | 0.640±0.079 |

| Pearson r (↑) | 298K [BMIM]$^+$ [BETI]$^-$ | 298K [EMIM]$^+$ [TF$_2$N]$^-$ | 298K [P$_{66614}$]$^+$ [L-Lact]$^-$ | 298K [QUIN$_8$]$^+$ [TF$_2$N]$^-$ | 328K [P$_{4442}$]$^+$ [DEP]$^-$ | 328K [QUIN$_8$]$^+$ [TF$_2$N]$^-$ | 338K [QUIN$_8$]$^+$ [TF$_2$N]$^-$ | 348K [P$_{4442}$]$^+$ [DEP]$^-$ |
|---|---|---|---|---|---|---|---|---|
| AIonopedia (Galactica-125m) | **0.9952±0.0005** | 0.9832±0.0022 | 0.9815±0.0047 | 0.9962±0.0009 | 0.9450±0.0065 | 0.9969±0.0006 | 0.9969±0.0005 | 0.9018±0.0030 |
| AIonopedia (Galactica-1.3b) | 0.9946±0.0004 | 0.9904±0.0012 | **0.9844±0.0023** | 0.9982±0.0002 | 0.9664±0.0056 | 0.9984±0.0002 | 0.9984±0.0002 | 0.9250±0.0039 |
| AIonopedia (Qwen3-0.6b) | 0.9943±0.0006 | **0.9908±0.0028** | 0.9718±0.0043 | **0.9987±0.0004** | 0.9685±0.0077 | 0.9985±0.0004 | 0.9984±0.0004 | **0.9300±0.0038** |
| AIonopedia (Gemma3-1b) | 0.9949±0.0005 | 0.9905±0.0013 | 0.9655±0.0090 | 0.9986±0.0005 | 0.9574±0.0049 | **0.9986±0.0004** | **0.9986±0.0004** | 0.9180±0.0053 |
| MD simulation | 0.8710±0.0171 | 0.9309±0.0115 | 0.7684±0.0248 | 0.7957±0.0418 | 0.8940±0.0089 | 0.7495±0.0441 | 0.8231±0.0311 | 0.8761±0.0056 |
| MLP | 0.9486±0.0026 | 0.9540±0.0045 | 0.8117±0.0232 | 0.9706±0.0034 | 0.9098±0.0093 | 0.9734±0.0018 | 0.9729±0.0019 | 0.8451±0.0091 |
| MolCA | 0.9876±0.0029 | 0.9602±0.0275 | 0.7441±0.0310 | 0.9369±0.0066 | 0.9053±0.0057 | 0.9474±0.0059 | 0.9471±0.0059 | 0.8294±0.0032 |
| T5chem | 0.9931±0.0008 | 0.9776±0.0047 | 0.9451±0.0195 | 0.9931±0.0009 | 0.9331±0.0037 | 0.9931±0.0007 | 0.9926±0.0009 | 0.8783±0.0032 |
| Molinst | 0.9938±0.0008 | 0.9862±0.0039 | 0.8582±0.0164 | 0.9978±0.0008 | 0.9341±0.0093 | 0.9981±0.0009 | 0.9980±0.0007 | 0.8709±0.0157 |
| SPMM | 0.9910±0.0017 | 0.9813±0.0029 | 0.9364±0.0247 | 0.9939±0.0011 | **0.9722±0.0069** | 0.9964±0.0008 | 0.9965±0.0008 | 0.9012±0.0201 |
| LlaSMol | 0.9926±0.0011 | 0.9795±0.0122 | 0.9578±0.0120 | 0.9973±0.0003 | 0.9308±0.0112 | 0.9976±0.0009 | 0.9972±0.0012 | 0.8747±0.0071 |
| PRESTO | 0.9927±0.0010 | 0.7868±0.0132 | 0.8809±0.0256 | 0.9968±0.0007 | 0.9267±0.0085 | 0.9973±0.0008 | 0.9974±0.0007 | 0.8690±0.0120 |
| ILBERT | 0.9922±0.0019 | 0.9813±0.0057 | 0.8766±0.0478 | 0.9954±0.0009 | 0.9321±0.0097 | 0.9974±0.0005 | 0.9978±0.0006 | 0.8952±0.0118 |

| Kendall τ (↑) | 298K [BMIM]$^+$ [BETI]$^-$ | 298K [EMIM]$^+$ [TF$_2$N]$^-$ | 298K [P$_{66614}$]$^+$ [L-Lact]$^-$ | 298K [QUIN$_8$]$^+$ [TF$_2$N]$^-$ | 328K [P$_{4442}$]$^+$ [DEP]$^-$ | 328K [QUIN$_8$]$^+$ [TF$_2$N]$^-$ | 338K [QUIN$_8$]$^+$ [TF$_2$N]$^-$ | 348K [P$_{4442}$]$^+$ [DEP]$^-$ |
|---|---|---|---|---|---|---|---|---|
| AIonopedia (Galactica-125m) | 0.945±0.002 | 0.935±0.004 | **0.907±0.027** | 0.950±0.006 | 0.813±0.010 | 0.952±0.007 | 0.953±0.006 | 0.797±0.007 |
| AIonopedia (Galactica-1.3b) | 0.946±0.002 | 0.922±0.003 | 0.905±0.014 | 0.970±0.004 | 0.846±0.014 | 0.966±0.006 | 0.963±0.003 | 0.831±0.010 |
| AIonopedia (Qwen3-0.6b) | 0.950±0.004 | **0.943±0.009** | 0.860±0.012 | **0.975±0.009** | 0.850±0.017 | 0.973±0.004 | 0.965±0.010 | 0.833±0.010 |
| AIonopedia (Gemma3-1b) | **0.952±0.002** | 0.933±0.009 | 0.853±0.027 | 0.970±0.007 | 0.828±0.009 | **0.975±0.001** | **0.967±0.004** | 0.817±0.006 |
| MD simulation | 0.678±0.030 | 0.733±0.027 | 0.574±0.033 | 0.620±0.053 | 0.705±0.015 | 0.555±0.045 | 0.651±0.032 | 0.702±0.011 |
| MLP | 0.849±0.004 | 0.864±0.008 | 0.617±0.025 | 0.850±0.010 | 0.753±0.007 | 0.873±0.005 | 0.860±0.011 | 0.724±0.003 |
| MolCA | 0.903±0.016 | 0.875±0.032 | 0.577±0.026 | 0.809±0.013 | 0.753±0.010 | 0.843±0.013 | 0.833±0.011 | 0.717±0.006 |
| T5chem | 0.931±0.005 | 0.900±0.006 | 0.798±0.042 | 0.933±0.008 | 0.785±0.008 | 0.935±0.005 | 0.937±0.004 | 0.762±0.005 |
| Molinst | 0.937±0.008 | 0.933±0.009 | 0.661±0.009 | 0.960±0.006 | 0.787±0.015 | 0.966±0.001 | 0.961±0.003 | 0.757±0.018 |
| SPMM | 0.932±0.004 | 0.941±0.003 | 0.810±0.036 | 0.929±0.007 | **0.876±0.014** | 0.946±0.005 | 0.954±0.005 | **0.838±0.020** |
| LlaSMol | 0.937±0.007 | 0.933±0.009 | 0.818±0.031 | 0.950±0.008 | 0.785±0.013 | 0.958±0.007 | 0.956±0.011 | 0.764±0.012 |
| PRESTO | 0.941±0.009 | 0.866±0.015 | 0.720±0.034 | 0.961±0.006 | 0.785±0.017 | 0.954±0.008 | 0.949±0.006 | 0.765±0.012 |
| ILBERT | 0.943±0.008 | 0.941±0.004 | 0.679±0.048 | 0.934±0.008 | 0.786±0.010 | 0.958±0.006 | 0.954±0.009 | 0.772±0.011 |

**Table S14.** Comparison of AIonopedia with computational baseline and prior baselines on small mass density dataset and hydration ΔG dataset.

| metric | mass density | | | hydration ΔG | | |
|---|---|---|---|---|---|---|
| | RMSE (g/cm$^3$) | Pearson r | Kendall τ | RMSE (kcal/mol) | Pearson r | Kendall τ |
| AIonopedia (Galactica-125m) | 0.0120±0.0006 | 0.9969±0.0005 | 0.956±0.006 | 0.966±0.027 | 0.9427±0.0030 | 0.909±0.008 |
| AIonopedia (Galactica-1.3b) | 0.0110±0.0006 | **0.9975±0.0003** | 0.962±0.006 | 0.802±0.016 | 0.9611±0.0012 | 0.940±0.007 |
| AIonopedia (Qwen3-0.6b) | **0.0106±0.0015** | 0.9974±0.0009 | **0.966±0.006 (0.9662)** | 0.749±0.018 | 0.9672±0.0018 | **0.948±0.002** |
| AIonopedia (Gemma3-1b) | 0.0124±0.0008 | 0.9963±0.0004 | 0.966±0.005 (0.9659) | 0.801±0.015 | 0.9627±0.0016 | 0.932±0.003 |
| MD simulation | 0.0670±0.0002 | 0.9800±0.0002 | 0.886±0.008 | 1.001±0.005 | 0.9452±0.0005 | 0.781±0.004 |
| MLP | 0.0385±0.0055 | 0.9747±0.0085 | 0.769±0.060 | 1.321±0.019 | 0.8940±0.0044 | 0.737±0.006 |
| MolCA | 0.0321±0.0039 | 0.9798±0.0022 | 0.931±0.015 | 1.035±0.191 | 0.9371±0.0214 | 0.825±0.048 |
| T5chem | 0.0132±0.0014 | 0.9963±0.0008 | 0.947±0.005 | 1.005±0.018 | 0.9382±0.0022 | 0.897±0.005 |
| Molinst | 0.0141±0.0014 | 0.9959±0.0005 | 0.930±0.016 | 0.969±0.037 | 0.9523±0.0049 | 0.901±0.009 |
| SPMM | 0.0299±0.0041 | 0.9855±0.0007 | 0.857±0.005 | 0.832±0.044 | 0.9610±0.0033 | 0.893±0.005 |
| LlaSMol | 0.0118±0.0006 | 0.9969±0.0005 | 0.944±0.009 | 0.727±0.084 | 0.9698±0.0079 | 0.935±0.008 |
| PRESTO | 0.0156±0.0014 | 0.9950±0.0012 | 0.948±0.011 | 1.035±0.0261 | 0.9340±0.0033 | 0.885±0.008 |
| ILBERT | 0.0193±0.0002 | 0.9925±0.0001 | 0.843±0.006 | **0.711±0.058** | **0.9752±0.0055** | 0.933±0.006 |

## S6.4. Ablation analysis of the multimodal property predictor.

To assess the contribution of individual components within the property predictor, we conducted an ablation analysis using the smallest AIonopedia variant with the Galactica-125m backbone. All variants were evaluated under the same IL-based split using identical training hyperparameters. The complete results are reported in Table S15 and summarized in Fig. 2F, with the full model achieving the best performance for all but one property.

We first evaluated the contribution of the training objectives. As the descriptor-based MLP baseline showed that physicochemical descriptors alone were insufficient to achieve competitive performance, we removed the descriptor-supervised objective while retaining the three BLIP-2-based self-supervised graph-text objectives: GTC, GTG and GTM. Removing physicochemical supervision increased the RMSEs of solvation free energy and IL/water transfer free energy by 22% and 38%, respectively, confirming that adding this objective does guide the model toward a deeper understanding of the system.

We next examined the contribution of the initial modality-alignment stage. Omitting this stage caused the largest overall performance degradation, increasing the RMSE on IL/water transfer free energy from 0.47 to 0.81 kcal $mol^{-1}$. Removing either the graph or language modality also reduced prediction accuracy, with a larger decline observed when the language branch was excluded.

**Table S15.** Performance metrics of the property predictor in the ablation study.

| RMSE | solvation ΔG (kcal/mol) | IL/water transfer ΔG (kcal/mol) | melting point (K) | viscosity (mPa·s) (log scale) | surface tension (mN/m) | mass density ($g/cm^3$) | organic solvent/water transfer ΔG (kcal/mol) |
|---|---|---|---|---|---|---|---|
| AIonopedia (Galactica-125m) | **0.304±0.030** | 0.473±0.100 | **38.4±3.0** | **0.249±0.018** | **3.63±0.27** | **0.0272±0.0029** | **0.415±0.028** |
| w/o supervised loss | 0.370±0.047 | 0.652±0.094 | 41.2±3.8 | 0.264±0.017 | 3.85±0.27 | 0.0287±0.0025 | 0.475±0.029 |
| w/o pretraining | 0.362±0.036 | 0.807±0.056 | 44.4±3.3 | 0.287±0.011 | 4.29±0.31 | 0.0313±0.0023 | 0.654±0.024 |
| w/o graph modality | 0.344±0.0810 | **0.463±0.128** | 39.8±3.1 | 0.257±0.017 | 3.69±0.27 | 0.0273±0.0030 | 0.426±0.031 |
| w/o language modality | 0.379±0.010 | 0.485±0.410 | 40.9±3.2 | 0.271±0.107 | 3.98±0.23 | 0.0298±0.0036 | 0.438±0.027 |

| Pearson r | solvation ΔG | IL/water transfer ΔG | melting point | viscosity | surface tension | mass density | organic solvent/water transfer ΔG |
|---|---|---|---|---|---|---|---|
| AIonopedia (Galactica-125m) | **0.9625±0.0056** | 0.9801±0.0104 | **0.7470±0.0380** | **0.9393±0.0080** | **0.9194±0.0160** | **0.9895±0.0021** | **0.9849±0.0025** |
| w/o supervised loss | 0.9502±0.0144 | 0.9692±0.0119 | 0.7101±0.0499 | 0.9334±0.0089 | 0.9118±0.0195 | 0.9887±0.0019 | 0.9808±0.0026 |
| w/o pretraining | 0.9496±0.0104 | 0.9467±0.0110 | 0.6406±0.0444 | 0.9215±0.0050 | 0.8882±0.0215 | 0.9867±0.0021 | 0.9650±0.0020 |
| w/o graph modality | 0.9574±0.0156 | **0.9807±0.0112** | 0.7256±0.0463 | 0.9358±0.0088 | 0.9134±0.0195 | 0.9892±0.0023 | 0.9844±0.0025 |
| w/o language modality | 0.9441±0.0037 | 0.9766±0.0131 | 0.7015±0.0310 | 0.9262±0.0097 | 0.9008±0.0208 | 0.9870±0.0029 | 0.9824±0.0026 |

| Kendall τ | solvation ΔG | IL/water transfer ΔG | melting point | viscosity | surface tension | mass density | organic solvent/water transfer ΔG |
|---|---|---|---|---|---|---|---|
| AIonopedia (Galactica-125m) | **0.870±0.019** | 0.883±0.016 | **0.546±0.031** | **0.834±0.015** | **0.800±0.021** | **0.931±0.004 (0.9314)** | **0.906±0.005 (0.9061)** |
| w/o supervised loss | 0.848±0.017 | 0.871±0.010 | 0.496±0.046 | 0.816±0.015 | 0.786±0.026 | 0.926±0.003 | 0.892±0.005 |
| w/o pretraining | 0.841±0.010 | 0.796±0.019 | 0.449±0.057 | 0.798±0.014 | 0.741±0.027 | 0.918±0.004 | 0.847±0.005 |
| w/o graph modality | 0.856±0.035 | **0.885±0.020** | 0.517±0.054 | 0.827±0.015 | 0.790±0.024 | 0.931±0.004 (0.9312) | 0.906±0.004 (0.9056) |
| w/o language modality | 0.809±0.013 | 0.878±0.022 | 0.500±0.040 | 0.809±0.017 | 0.758±0.024 | 0.923±0.006 | 0.898±0.004 |

**S7. Selection criteria and detailed discussion for literature case studies used for mechanistic interpretation of biphasic transitions.**

Literature case studies used for mechanistic interpretation were manually curated under expert supervision from high-impact chemistry and materials journals. The selection process was designed to ensure experimental relevance, interpretability, and independence from the training data. Studies were included only if they satisfied the following criteria:

1. Experimental focus. Studies were required to contain substantial experimental measurements or observations, rather than being predominantly computational or theoretical.
2. Relevance to model-predicted physicochemical properties. Preference was given to studies reporting physicochemical properties directly predicted by AIonopedia. When such measurements were unavailable, studies were considered if they reported quantitatively or qualitatively observable phenomena that could be mechanistically related to model-predicted thermodynamic or physicochemical quantities.
3. Chemical representability. The investigated systems were required to be representable (either exactly or with reasonable approximations) using chemical structures and identifiers supported by AIonopedia, enabling reliable model inference. Consequently, systems involving biomacromolecules, poorly defined compositions, or highly complex multicomponent mixtures were excluded.
4. Redundancy control. Studies were considered redundant and excluded when they had highly similar systems or duplicated measurements, as assessed based on the case-defining experimental variables rather than shared probe species or target analytes alone. For example, when two battery studies screened substantially overlapping cathode or anode redox-active species, only one representative case was retained. Preference was given to publications in journals with higher impact factors; if the impact factors were similar, the more recent publication was selected.
5. Independence from the training dataset. The reported measurements were essentially absent from the AIonopedia training dataset, except for negligible unavoidable overlap, thereby ensuring an independent evaluation of predictive performance.

These criteria yielded a diverse collection of independent experimental studies spanning multiple application scenarios, enabling rigorous assessment of whether the model can provide physically meaningful interpretations beyond predictive performance alone.

### S7.1. Representation of complex multiphase systems.

Because the current AIonopedia input format does not explicitly represent multicomponent solvent mixtures, each experimental phase was approximated by the component expected to dominate the relevant solvation or partitioning interactions, excluding the target solute whose free energy was evaluated. These simplified representations were intended to preserve the principal chemical contrast between phases rather than reproduce the complete mixture composition.

For the membrane-based microreactor in Fig. 3C, the IL and oil phases were represented by $[BMIM]^+[PF_6]^-$ and *n*-octane, respectively, while other dissolved species were omitted. For the first battery-related study in Fig. 3D, the NFTOS-rich anolyte and the TEGDME- or DMSO-rich catholyte were represented by NFTOS and TEGDME or DMSO, respectively. For the second battery-related study in Fig. 3E, the two phases were represented by $H_2O$ and $CH_2Cl_2$.

The iodine-extraction case in Fig. 3F required a different treatment because the objective was to determine how individual ions influenced the partitioning of iodine species. The corresponding salt solutions were therefore represented as virtual IL phases composed of their constituent cations and anions. This formulation enabled the ionic contributions to be evaluated separately and identified $[EMIM]^+$ as the primary driver of extraction over the accompanying anion.

### S7.2. Oligomeric representation of polymer phases.

Because AIonopedia operates on finite molecular sequences and does not explicitly represent polymer molecular-weight distributions, polymeric phases were modeled using fixed-length oligomeric structures. Each polymer was represented by a 10-unit oligomer, chosen as a practical compromise between limiting sequence-length OOD effects, retaining a chemically meaningful oligomeric structure, and reaching a regime in which further chain extension generally produced only minor changes in the predicted values. As different repeat units generate SMILES of different lengths, the same number of repeat units can yield different input-sequence lengths and thus affect model predictions, causing the onset of relative prediction stability to vary across polymers. Nevertheless, ten units provided a practical common setting suitable for most systems.

To reduce terminal-group effects at finite chain lengths, one additional carbon atom was appended to each oligomer terminus, including carbon-terminated structures, to ensure consistent treatment across polymers. For randomly substituted polymers such as Q-Dex, the fully substituted structure was used as a representative input, thereby avoiding the arbitrary selection of a particular substitution pattern.

Because these polymer-containing cases rely on customized structural approximations, the resulting predictions should be interpreted as evidence of trend-level transferability under severe OOD conditions rather than as exact simulations of the complete experimental systems.

**S8. Selection criteria and detailed discussion for literature case studies used for mechanistic interpretation related to bulk properties.**

To further assess the OOD generalization capability of AIonopedia, we curated an independent set of literature studies reporting experimentally measured bulk properties of ILs. The selection procedure followed principles similar to those described in the previous section, while requiring that the target properties be explicitly reported as quantitative values for predictive validation. This requirement reflects the role of bulk properties as auxiliary descriptors for characterizing IL systems and experimental conditions, rather than as standalone indicators for rationalizing wet-lab observations. Moreover, we considered only original articles and excluded review articles and secondary data compilations even when suitable experimental values were available.

Using these criteria, we identified four independent studies suitable for external validation[75-78], as shown in Fig. S3. The resulting literature-grounded test set contained 13 ILs with specially designed ions, featuring structures that differ substantially from those of conventional ILs and vary markedly across the selected studies, thereby inherently satisfying the redundancy-control criterion. Based on the original literature reports, we predicted the viscosity, density, and melting point values for these ILs. Across all available measurements, the corresponding RMSE values were 0.271 (mPa·s, Log scale), 0.0329 g/cm$^3$, and 24.1 K, respectively, comparable to those observed in the dataset-level benchmark evaluations described above. Notably, the melting-point error was substantially lower than that obtained in the corresponding benchmark, although this improvement may largely reflect sampling bias arising from the limited availability of literature-grounded validation data. In addition, the maximum absolute errors for viscosity, density, and melting point were 0.603 (mPa·s, Log scale), 0.0547 g/cm$^3$, and 43.0 K, respectively, further demonstrating the strong predictive capability of AIonopedia on structurally distinct IL systems.

However, no eligible original research articles meeting all selection criteria were identified for surface tension. To fill this gap, we treated this property as a special case and used the following result reported in a recent review[79] as an illustrative example: the experimentally reported surface tensions of $[(mPEG_2)_2IM]^+[I]^-$ and $[(mPEG_2)_2IM]^+[PF_6]^-$ are 46.7 and 45.6 mN/m at 293 K, respectively, while AIonopedia predicted corresponding values of 46.6 and 45.6 mN/m. Despite the excellent agreement, these data were not included in the external validation set, as they did not originate from an eligible primary research study.

| reported | predicted |
|---|---|

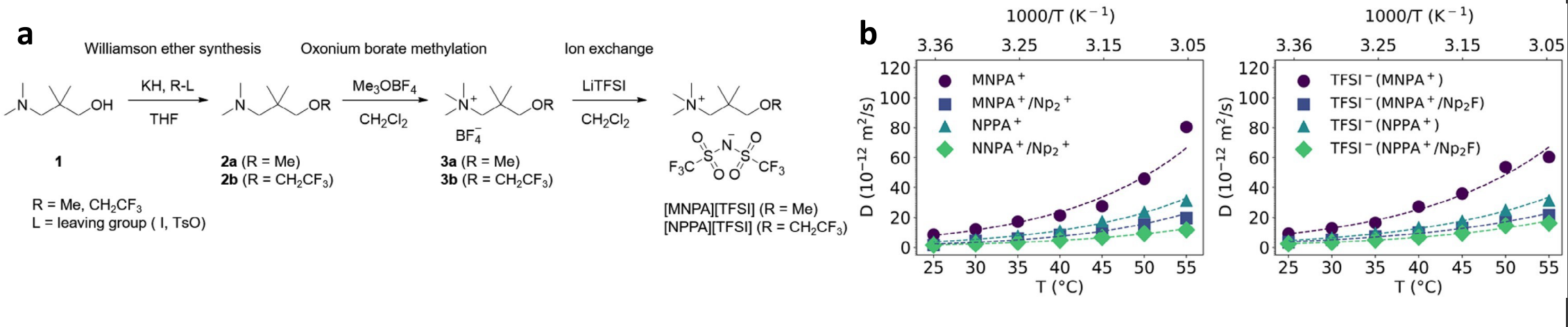

$[MNPA]^+[TFSI]^-$:

η = 210 mPa·s (Real: 235 mPa·s)

mp = 295 K (Real: 284 K)

$[NPPA]^+[TFSI]^-$:

η = 476 mPa·s (Real: 353 mPa·s)

mp = 316 K (Real: 299 K)

**(A)** (a) Synthetic route to the ionic liquids used in this study (b) Temperature dependence of the diffusion coefficients, D, for cations and anions

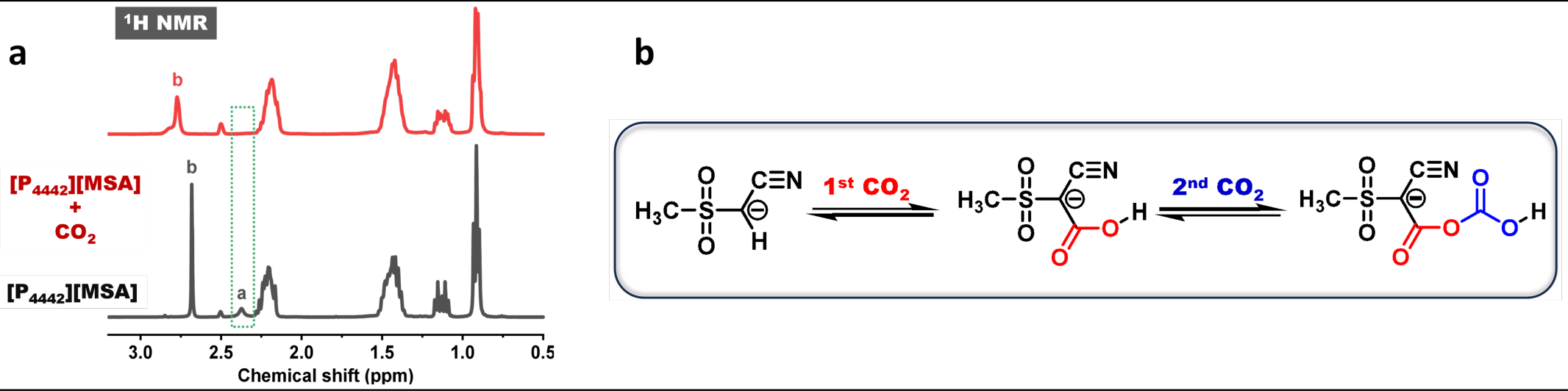

$\rho([P_{66614}]^+[MSA]^-)$ = 0.921 g/cm$^3$ (Real: 0.929 g/cm$^3$)

$\rho([P_{4442}]^+[MSA]^-)$ = 1.002 g/cm$^3$ (Real: 1.040 g/cm$^3$)

**(B)** (a) $^1$H NMR Spectra of $[P_{4442}]^+[MSA]^-$ before and after reacting with $CO_2$ (b) Proposed reaction pathway in $CO_2$ capture by $[P_{4442}]^+[MSA]^-$

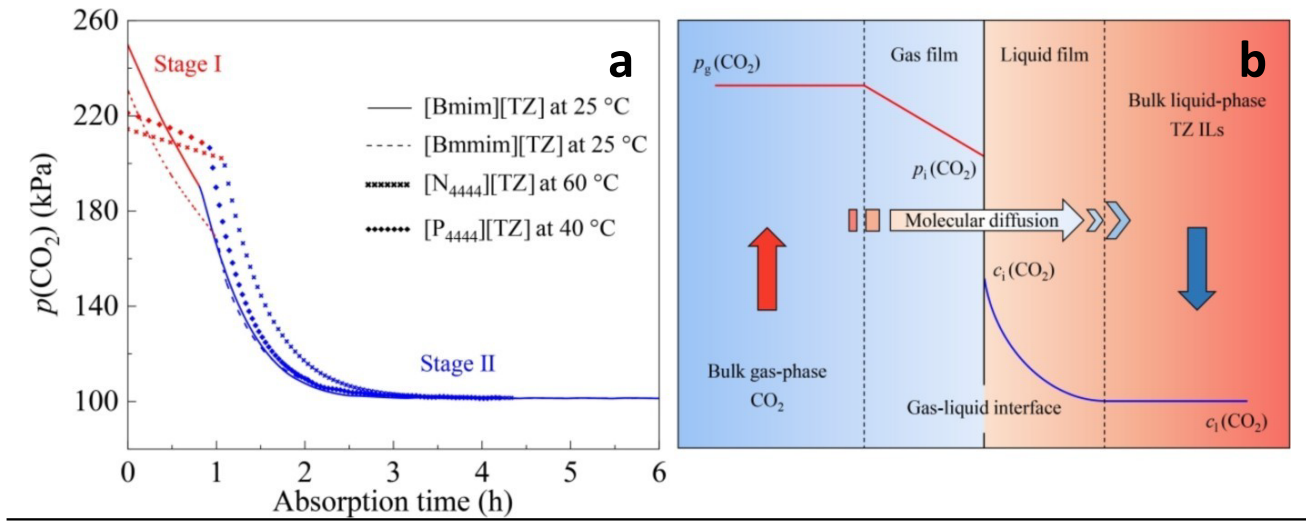

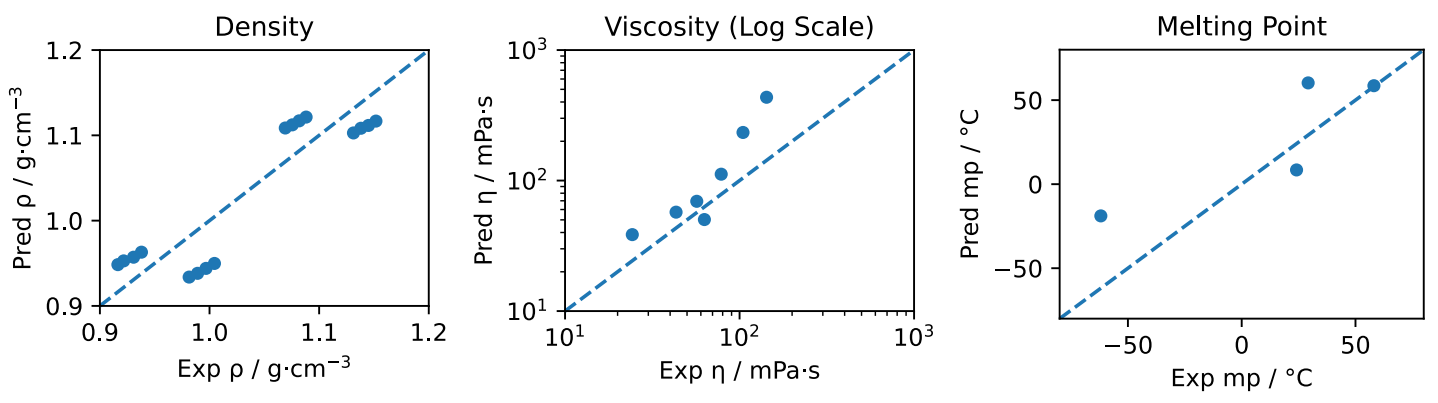

ρ RMSE: 0.0381 g/cm$^3$ η RMSE: 0.254 mp RMSE: 27.7 K

**(C)** Two-stage processes for $CO_2$ capture in $[TZ]^-$ ILs: (a) The variation of $CO_2$ pressure with absorption time and (b) a schematic diagram of the two-film theory.

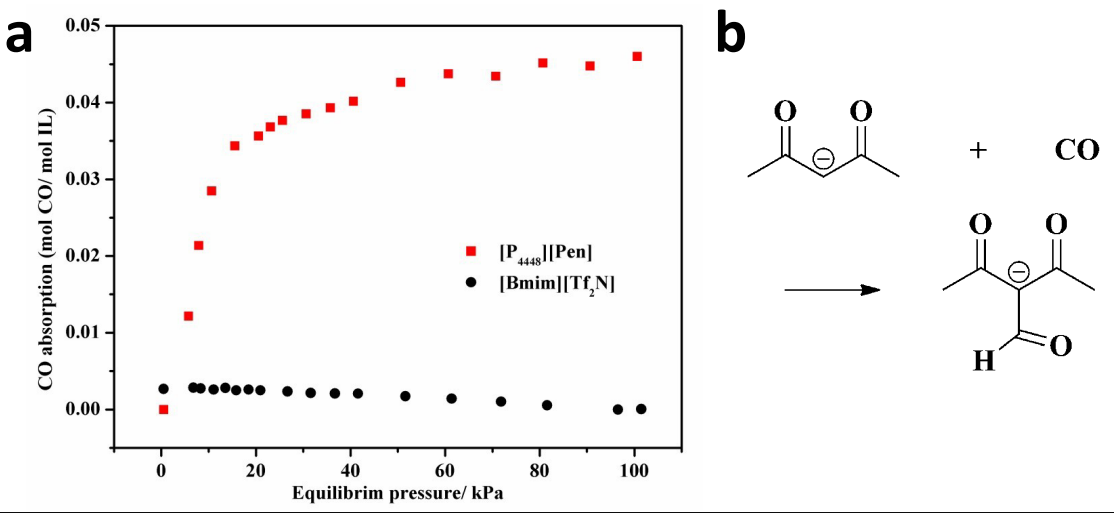

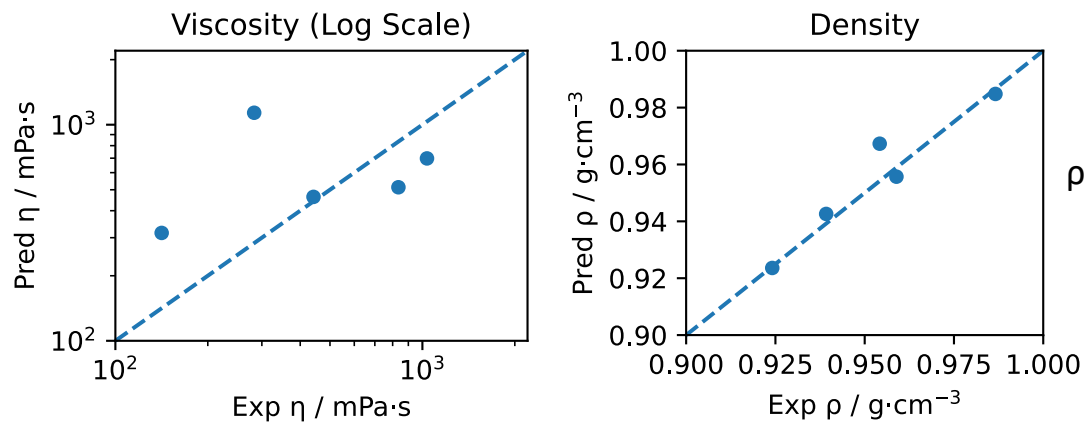

ρ RMSE: 0.0063 g/cm$^3$

η RMSE: 0.335

**(D)** (a) CO absorption in $[P_{4448}]^+[Pen]^-$ and $[BMIM]^+[Tf_2N]^-$ at 25 °C (b) The proposed mechanism of CO absorption by $[P_{4448}]^+[Pen]^-$ through C-site interaction

**Fig. S3. Computational predictions for IL bulk-property cases.** For each panel, top left: results reported in the original paper; top right: model predictions; bottom: interpretation of the observed phenomena.

## S9. Finite-difference formulation of perturbative cooperativity analysis.

Explainable analysis or XAI has become increasingly important for revealing the underlying mechanisms learned by deep learning models. In physical chemistry and molecular modelling, XAI enables the interpretation of structure-property relationships and facilitates rational molecular optimization. However, the exploration of XAI in ILs remains limited despite their complex many-body interactions and vast compositional space. In this work, we integrate the physicochemical concept of many-body expansion into an XAI framework to establish a functional-group-level attribution method for interpreting the physicochemical properties of ionic solvents, providing both mechanistic insights and design guidance.

For any chemical environment $S$, we denote by $A_p$ the perturbation corresponding to the introduction of group $A$ at position $p$, yielding a modified environment $S + A_p$. To maintain the chemical validity and closed-shell nature of individual species, all fragments involved in XAI perturbative analyses were H-capped by default according to a fixed and reproducible surrogate construction protocol. This procedure is not intended to recover physically isolated fragments, but rather to provide a unified, computable, and comparable reference representation for the model.

Under such a chemical environment, for an arbitrary observable $O$, which may represent either a mechanical or thermodynamic property, we define the finite-difference perturbation operator $\delta_{A_p}$ associated with the introduction of group $A$ at position $p$ as

$$(\delta_{A_p}O)(S) \equiv O(S + A_p) - O(S) \quad \text{(S22).}$$

Accordingly, the first-order contribution of this modification process is

$$\Delta O(A_p \mid S) \equiv (\delta_{A_p}O)(S) = O(S + A_p) - O(S) \quad \text{(S23).}$$

An interesting observation is that, for IL-related thermodynamic observables, this first-order attribution can also be interpreted from a physical perspective. When the perturbation is applied to the IL structure and $O$ represents the solvation or transfer free energy of an external solute, $\Delta O(A_p \mid S)$ corresponds to a hypothetical inter-solvent transfer free energy of the solute between the reference IL environment $S$ and the modified IL environment $S + A_p$. Although such a biphasic transfer process may not be experimentally realizable because of miscibility or other constraints, this hypothetical view enriches the physicochemical meaning of the resulting attribution.

For a disubstituted system $S + A_p + B_q$, the second-order cooperativity between the two groups, denoted as $\Delta_{\mathrm{coop}} O(A_p, B_q \mid S)$ or equivalently $\Delta^{(2)} O(A_p, B_q \mid S)$, is defined as the change in the contribution of $A_p$ upon introduction of $B_q$

$$\begin{aligned} \Delta^{(2)} O\left(A_p, B_q \mid S\right) &\equiv \Delta O\left(A_p \mid S + B_q\right) - \Delta O\left(A_p \mid S\right) \\ &= \left(\delta_{B_q} \delta_{A_p} O\right)(S) \\ &= O(S + A_p + B_q) - O(S + B_q) - O(S + A_p) + O(S) \end{aligned} \quad \text{(S24).}$$

Provided that the final chemically valid environment is independent of the order in which the two perturbations are introduced, the finite-difference operators commute, giving

$$(\delta_{B_q} \delta_{A_p} O)(S) = (\delta_{A_p} \delta_{B_q} O)(S) \quad \text{(S25).}$$

This term therefore represents the nonadditive pairwise contribution that cannot be decomposed into the two individual first-order effects.

The same finite-difference notation can be formally extended to a discrete many-body expansion. To avoid confusion with the positional subscripts $p$ and $q$, we use superscripts to label generic perturbations. Specifically, let $X^{(r)} \equiv A_{p_r}^{(r)}$ denote the $r$-th perturbation, corresponding to the introduction of group $A^{(r)}$ at position $p_r$. The corresponding finite-difference operator is denoted as $\delta^{(r)} \equiv \delta_{X^{(r)}}$. By definition,

$$(\delta^{(r)} O)(S) = O(S + X^{(r)}) - O(S) \quad \text{(S26),}$$

or equivalently,

$$O(S + X^{(r)}) = O(S) + (\delta^{(r)} O)(S) = \left[(1 + \delta^{(r)}) O\right](S) \quad \text{(S27).}$$

For multiple mutually compatible perturbations whose final environment is independent of the order of introduction, repeated application gives

$$O\left(S + \textstyle\sum_r X^{(r)}\right) = \left[\textstyle\prod_r (1 + \delta^{(r)}) O\right](S) \quad \text{(S28).}$$

Expansion of this product gives

$$\begin{aligned} O\left(S + \textstyle\sum_r X^{(r)}\right) = &\ O(S) + \textstyle\sum_r (\delta^{(r)} O)(S) + \textstyle\sum_{r<s} (\delta^{(s)} \delta^{(r)} O)(S) \\ &+ \textstyle\sum_{r<s<t} (\delta^{(t)} \delta^{(s)} \delta^{(r)} O)(S) + \cdots \end{aligned} \quad \text{(S29).}$$

The first- and second-order terms quantify individual substituent effects and pairwise non-additivity, respectively. Higher-order terms can be defined in the same manner, but the first- and second-order contributions are usually the most useful and interpretable in chemical applications.

An important special case of the above second-order analysis is the null-base limit $S = \emptyset$. In this case, since no underlying environment exists, the position labels $p$ and $q$ are omitted. Under the same surrogate construction and H-capping protocol, the null-base perturbation by $A$ directly yields the H-capped surrogate of $A$. We therefore identify $\Delta O(A \mid \emptyset) \equiv O(A)$ as an operational definition, without requiring $O(\emptyset)$ to be evaluated as an independent observable. The corresponding null-base expression therefore becomes

$$\begin{aligned}\Delta^{(2)} O(A, B \mid \emptyset) &= \Delta O(A \mid B) - \Delta O(A \mid \emptyset) \\ &= O(A + B) - O(B) - O(A)\end{aligned} \quad \text{(S30).}$$

Taken together, these definitions establish a physics-informed, many-body-expansion-based counterfactual thermodynamic attribution framework. Through this formulation, finite-difference attribution strategies widely used for protein-ligand binding affinity analysis are extended in a new context to the interpretation of IL-related properties. More broadly, this framework provides a conceptual bridge between XAI and physicochemical structure-property relationships in ILs.

Illustrative calculations of such explainable AI analysis are presented in Fig. 4, where all analyses have their correspondence in literature reports. The original references of these structure-property relationships or interaction motifs in ILs are provided here[80-94].

**S10. Interpretable symbolic correction for IL density prediction.**

The XAI analysis described above provides a chemically localized interpretation of AIonopedia predictions by decomposing property changes into first-order functional-group effects and higher-order nonadditive interactions. This fragment-level view is particularly useful for identifying which structural motifs contribute to a predicted physicochemical property and how different fragments cooperate. However, such counterfactual attributions mainly explain the model in a local and discrete manner. A complementary question is whether the information learned by AIonopedia can also be expressed in a more global, compact, and analytically usable form. Therefore, beyond fragment-level attribution analysis, we explored symbolic correction as another route toward model interpretability. Taking mass density, the property for which our model achieved the best average learning performance, as a representative example, we examined whether the descriptor space in AIonopedia could provide a compact closed-form correction to the systematic residuals of an empirical density relation.

Traditional analytical formulas for IL density typically rely on predefined group-contribution volumes or other empirical parameters.[95-98] These parameters must be obtained from literature tables, or the ionic structures must be partitioned into groups according to specific rules. As a result, their applicability is limited by parameter coverage and by how the molecular groups are defined. In addition, such methods often lack ready-to-use code implementations, making them inconvenient for practical experimental-assistance scenarios. To address these issues, we replaced the empirical ionic-volume term in the original formulas with an RDKit-derived geometric volume descriptor that can be directly computed from molecular structure. In this way, the volume input becomes more accessible and reproducible for automated density estimation.

Our work follows a subsequent development of the classic Ye and Shreeve method[96]. Since our dataset is limited to ambient pressure, the density of a given ionic solvent can be expressed as

$$\rho = \frac{M}{N_A V_0 (a+bT)} \quad \text{(S31)},$$

where $M$ is the molar mass of the ion-pair composition, $N_A$ is the Avogadro constant, $V_0$ is the total volume of the individual ions at the reference temperature, and $a$ and $b$ are empirical parameters describing the temperature dependence of the effective volume. Here, $M$ and $V_0$ are obtained by summing the masses and volumes of the constituent cation and anion, respectively. This baseline formula achieved an RMSE of 0.0629 g/cm$^3$ under the cation-based 5-fold split and 0.0626 g/cm$^3$ under the IL-based 5-fold split. Although this baseline provided reasonable density estimates, its remaining systematic deviations suggested that additional structure-derived descriptors could further improve the analytical density relation.

Following the treatment of mass and volume in previous methods, we performed an element-wise summation of the corresponding cation and anion values for the 21 descriptors used in the modality-alignment phase of AIonopedia, and additionally included molar volume as the 22nd descriptor. All features were then fed into a default MLP regressor implemented in scikit-learn for permutation analysis. Based on the more stringent cation-based split, we selected 15 descriptors whose mean permutation importance across the five folds exceeded 0.1 for the second-stage symbolic regression. The molar mass and volume terms were combined into a single baseline variable, $K_0 = \frac{M}{N_A V_0}$ , so that the original density equation can be rewritten as

$$\rho = \frac{K_0}{a+bT} \quad \text{(S32).}$$

To retain the temperature dependence of the original analytical form, we did not fit $\rho$ directly. Instead, we used $\rho(a + bT)$ as the regression target, allowing symbolic regression to learn a correction to $K_0$ while keeping the final density inversely proportional to $a + bT$. The final fitted expression can be written as:

$$\rho = \frac{K_0 + \beta \mathrm{LogP} + \gamma}{a+bT} \quad \text{(S33).}$$

As in the original Ye and Shreeve method, an additional constant term was retained here to improve accuracy. We further performed an ablation study on the two added terms, and the performance comparison is shown in Table S16. Relative to the $K_0$-only baseline, the LogP term reduced the RMSE by 8.1%, whereas the constant term alone gave a smaller improvement of 4.5%. Combining both terms achieved the lowest RMSE of 0.0530 g/cm$^3$, corresponding to a 15.7% reduction from the baseline.

To further examine whether the descriptor-augmented symbolic formula was related to the correction made by AIonopedia, we compared three baseline-relative terms, including the AIonopedia correction ($\rho_{\mathrm{AIonopedia}} - \rho_{\mathrm{base}}$), the symbolic-formula correction ($\rho_{\mathrm{symbolic\ regression}} - \rho_{\mathrm{base}}$), and the experimental residual ($\rho_{\mathrm{experiment}} - \rho_{\mathrm{base}}$), as shown in Fig. S4. Across the two 5-fold evaluation splits, the symbolic-formula correction showed moderate positive correlations with both quantities, with a slightly stronger correlation to the AIonopedia correction. By comparison, the AIonopedia correction was much more strongly correlated with the experimental residual, with $r = 0.861$ and $r = 0.905$ under the two splits.

Partial Pearson-correlation analysis was then used to account for the shared dependence on experimental density. After controlling for the experimental residual, the symbolic-formula correction still remained correlated with the AIonopedia correction ($r_{partial} = 0.175$ for cation split and $0.213$ for IL split). In contrast, its partial correlation with the experimental residual after controlling for AIonopedia was weaker ($r_{partial} = 0.172$ for cation split and $0.075$ for IL split). This asymmetry is notable because the symbolic formula was fitted directly to experimental density values.

Viewed from the training process, this result is also informative: descriptor prediction is introduced through the supervised pseudo-label objective during multimodal pretraining, whereas downstream density fine-tuning is driven only by the property target itself. If such information were not useful for density prediction, its influence in AIonopedia would be expected to diminish during fine-tuning, making this association with the descriptor-explicit symbolic formula less likely after controlling for the experimental residual. The persistence of the association therefore suggests that the supervised pseudo-label objective is not merely an auxiliary training signal, but an important route for incorporating domain prior knowledge into the learned representation, consistent with the ablation-study results that demonstrate its importance. Overall, the descriptor-augmented symbolic formula provides a compact analytical representation of the physicochemical knowledge retained by AIonopedia.

**Table S16.** Performance comparison of density calculation formulas with different correction terms.

| Metric | Cation split | | | IL split | | |
|---|---|---|---|---|---|---|
| | RMSE (g/cm$^3$) | Pearson r | Kendall τ | RMSE (g/cm$^3$) | Pearson r | Kendall τ |
| $K_0$ | 0.0629±0.0039 | 0.9484±0.0071 | 0.809±0.020 | 0.0626±0.0042 | 0.9467±0.0099 | 0.807±0.025 |
| $K_0 + \beta$LogP | 0.0578±0.0025 | 0.9577±0.0055 | 0.835±0.017 | 0.0575±0.0035 | 0.9576±0.0079 | 0.834±0.015 |
| $K_0 + \gamma$ | 0.0601±0.0046 | 0.9485±0.0071 | 0.809±0.020 | 0.0596±0.0037 | 0.9467±0.0099 | 0.807±0.025 |
| $K_0 + \beta$LogP $+ \gamma$ | **0.0530±0.0034** | **0.9581±0.0054** | **0.835±0.017** | **0.0531±0.0029** | **0.9580±0.0079** | **0.836±0.014** |

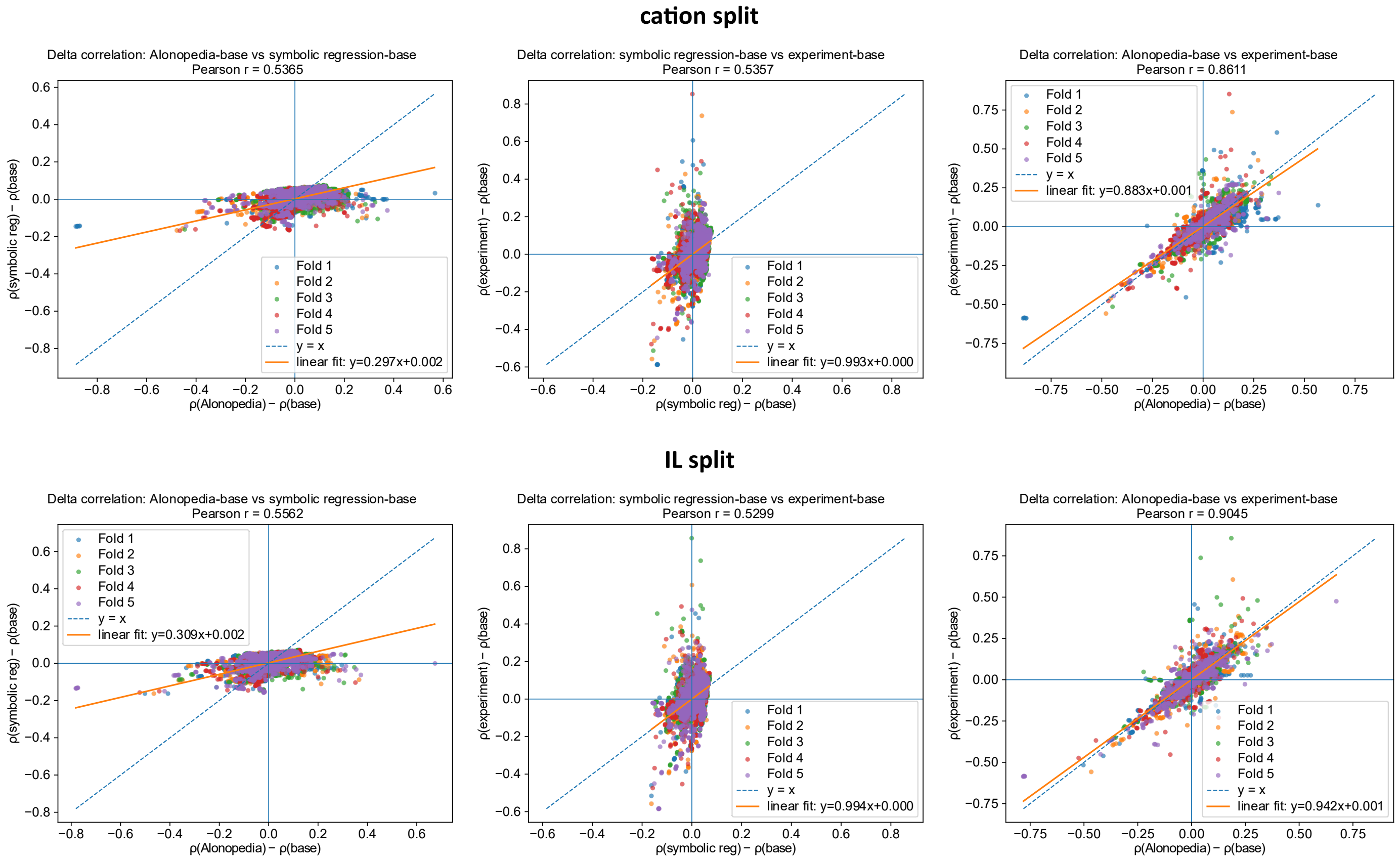

**Fig. S4. Pairwise correlations of density deviations from the empirical baseline.** Results are shown under two evaluation splits for $\rho_{\text{AIonopedia}} - \rho_{\text{base}}$, $\rho_{\text{symbolic regression}} - \rho_{\text{base}}$, and $\rho_{\text{experiment}} - \rho_{\text{base}}$.

## S11. Wet-lab experiments.

### S11.1. Materials.

We performed wet-lab experiments on the AIonopedia-designed species for VOC absorption. The ILs were purchased from Shanghai Aladdin Biochemical Technology Co., Ltd. or synthesized if the synthetic approaches were reported in the literature. Organic solvents such as acetonitrile (99.7%), toluene (99.7%), and ethanol (99.7%) were purchased from Chengdu Kelong Chemical Co., Ltd. All chemicals were directly used without further purification.

### S11.2. Synthesis of ILs.

All ILs were prepared using either the classic neutralization reaction or the metathesis reaction. The general method is as follows.[99]

Neutralization reaction: Each IL containing the target cation was first passed through a strongly basic anion exchange column to replace its anion with hydroxide ions, yielding the hydroxide-form IL. The concentration of the resulting hydroxide-form IL was determined by acid-base titration using a potassium hydrogen phthalate solution as the primary standard and phenolphthalein as the indicator. An accurately measured volume of this hydroxide-form IL was then neutralized with the acid corresponding to the target anion in a 1:1 molar ratio. Following the removal of the water generated during neutralization by drying, the desired IL was obtained.

Metathesis reaction: Salts containing the target cation and the target anion, respectively, were accurately weighed and mixed. The resulting mixture was extracted with an organic solvent. The extract was filtered, concentrated, and dried to afford the desired IL.

#### S11.2.1 Synthesis of Intermediates

**$[P_{88810}]^+[Br]^-$:** A mixture of 1-bromodecane (1 eq) and trioctylphosphine (1 eq) was heated at 120 °C overnight under an $N_2$ atmosphere. $^1$H NMR (400 MHz, DMSO): $\delta_H$ ppm: 2.46 (m, 8H), 1.56 (m, 16H), 1.26 (m, 36H), 0.87 (m, 12H).

**$[QUIN_8]^+[Br]^-$:** Quinuclidine (1 eq) and 1-bromooctane (1 eq) were dissolved in MeCN (10 mL) and heated at reflux for 12 h under an $N_2$ atmosphere. $^1$H NMR (400 MHz, $CDCl_3$): $\delta_H$ ppm: 3.41 (m, 6H), 3.10 (m, 2H), 2.16 (m, 1H), 2.01 (m, 6H), 1.60 (m, 2H), 1.38-1.10 (m, 10H), 0.81 (m, 3H).

**$[N_{88810}]^+[Br]^-$:** 1-Bromodecane (1 eq) and trioctylamine (1 eq) were heated at 120 °C for 24 h under an $N_2$

atmosphere. $^1$H NMR (400 MHz, $CDCl_3$): $\delta_H$ ppm: 3.39 (m, 8H), 1.84 (m, 2H), 1.51-1.14 (m, 50H), 0.87 (m, 12H).

**$[P_{44414}]^+[Br]^-$:** Tetradecyl bromide (1 eq) and tributylphosphine (1 eq) were dissolved in MeCN and heated at 80 °C for 12 h under an $N_2$ atmosphere. $^1$H NMR (400 MHz, DMSO) $\delta_H$ ppm: 2.11 (m, 8H), 1.51 (m, 16H), 1.27 (m, 20H), 0.97 (m, 9H), 0.88 (m, 3H).

### S11.2.2 Ionic Liquids for Ethyl acetate and Acetone Absorption

**$[EMIM]^+[TCB]^-$:** Equimolar amounts of $[EMIM]^+[Br]^-$ and potassium tetracyanoborate were dissolved in dichloromethane and stirred for 4 h. The product was extracted, and the organic phase was filtered, concentrated, and dried to obtain $[EMIM]^+[TCB]^-$. $^1$H NMR (400 MHz, DMSO) $\delta_H$ ppm: 9.10 (s, 1H), 7.68 (m, 2H), 4.19 (m, 2H), 3.84 (m, 3H), 1.41 (m, 3H)

**$[HMIM]^+[TCB]^-$:** Equimolar amounts of $[HMIM]^+[Br]^-$ and potassium tetracyanoborate were dissolved in dichloromethane and stirred for 4 h. The product was extracted, and the organic phase was filtered, concentrated, and dried to obtain $[HMIM]^+[TCB]^-$. $^1$H NMR (400 MHz, DMSO) $\delta_H$ ppm: 9.15 (s, 1H), 7.78 (t, 1H), 7.72 (t, 1H), 4.17 (t, 2H), 3.88 (s, 3H), 1.79 (m, 2H), 1.28 (m, 6H), 0.87 (m, 3H)

**$[BMpy]^+[TDI]^-$:** Equimolar amounts of $[BMpy]^+[I]^-$ and potassium 4,5-dicyano-2-trifluoromethylimidazolate were dissolved in dichloromethane and stirred for 4 h. The product was extracted, and the organic phase was filtered, concentrated, and dried to obtain $[BMpy]^+[TDI]^-$. $^1$H NMR (400 MHz, DMSO) $\delta_H$ ppm: 8.99 (s, 1H), 8.91 (d, 1H), 8.45 (d, 1H), 8.04 (t, 1H), 4.56 (t, 2H), 2.47 (s, 3H), 1.89 (m, 2H), 1.27 (m, 2H), 0.91 (m, 3H)

**$[BMIM]^+[TDI]^-$:** Equimolar amounts of $[BMIM]^+[Br]^-$ and potassium 4,5-dicyano-2-trifluoromethylimidazolate were dissolved in dichloromethane and stirred for 4 h. The product was extracted, and the organic phase was filtered, concentrated, and dried to obtain $[BMIM]^+[TDI]^-$. $^1$H NMR (400 MHz, DMSO) $\delta_H$ ppm: 9.13 (s, 1H), 7.78 (t, 1H), 7.71 (t, 1H), 4.17 (t, 2H), 3.44 (m, 3H), 1.78 (m, 2H), 1.27 (m, 2H), 0.90 (m, 3H)

**$[HMIM]^+[TF_2N]^-$:** $[HMIM]^+[I]^-$ was converted to its hydroxide form via an anion exchange column, followed by neutralization with bis(trifluoromethane)sulfonimide. $^1$H NMR (400 MHz, DMSO) $\delta_H$ ppm: 9.15 (s, 1H), 7.77 (t, 1H), 7.70 (t, 1H), 4.16 (t, 2H), 3.85 (s, 3H), 1.78 (m, 2H), 1.27 (m, 6H), 0.86 (m, 3H)

**$[EMIM]^+[TF_2N]^-$:** $[EMIM]^+[Br]^-$ was converted to its hydroxide form via an anion exchange column, followed by neutralization with bis(trifluoromethane)sulfonimide. $^1$H NMR (400 MHz, $CDCl_3$): $\delta_H$ ppm: 8.64 (s, 1H), 7.76 (t, 1H), 7.69 (t, 1H), 4.16 (q, 2H), 3.80 (s, 3H), 1.41 (m, 3H).

### S11.2.3 Ionic Liquids for Toluene and o-Xylene Absorption

**$[P_{88810}]^+[2\text{-CNPyr}]^-$:** $[P_{88810}]^+[Br]^-$ was converted to its hydroxide form via an anion exchange column, followed by neutralization with pyrrole-2-carbonitrile. $^1H$ NMR (400 MHz, $CDCl_3$): $\delta_H$ ppm: 6.97 (m, 1H), 6.71 (m, 1H), 6.08 (m, 1H), 2.27 (m, 8H), 1.42 (m, 16H), 1.19 (m, 36H), 0.83 (m, 12H).

**$[QUIN_8]^+[TF_2N]^-$:** $[QUIN_8]^+[Br]^-$ was converted to its hydroxide form via an anion exchange column, followed by neutralization with bis(trifluoromethane)sulfonimide. $^1H$ NMR (400 MHz, $CDCl_3$): $\delta_H$ ppm: 3.42 (m, 6H), 3.11 (m, 2H), 2.18 (m, 1H), 2.01 (m, 6H), 1.67 (m, 2H), 1.40-1.13 (m, 10H), 0.87 (m, 3H).

**$[P_{88810}]^+[HX]^-$:** Equimolar amounts of $[P_{88810}]^+[Br]^-$ and sodium hypoxanthinate (HX) were dissolved in dichloromethane and stirred for 4 h. The product was extracted, and the organic phase was filtered, concentrated, and dried to obtain $[P_{88810}]^+[HX]^-$. $^1H$ NMR (400 MHz, $CDCl_3$): $\delta_H$ ppm: 8.85 (s, 1H), 8.08 (s, 1H), 7.88 (s, 1H), 2.26 (m, 8H), 1.44 (m, 16H), 1.24 (m, 36H), 0.87 (m, 12H).

**$[P_{44414}]^+[DCI]^-$:** $[P_{44414}]^+[Br]^-$ was converted to its hydroxide form via an anion exchange column, followed by neutralization with 4,5-dicyanoimidazole. $^1H$ NMR (400 MHz, $CDCl_3$): $\delta_H$ ppm: 3.93 (s, 1H), 2.32 (m, 8H), 1.52 (m, 16H), 1.25 (m, 20H), 0.97-0.87 (m, 12H).

**$[N_{88810}]^+[2\text{-CNPyr}]^-$:** $[N_{88810}]^+[Br]^-$ was converted to its hydroxide form via an anion exchange column, followed by neutralization with pyrrole-2-carbonitrile. $^1H$ NMR (400 MHz, $CDCl_3$): $\delta_H$ ppm:6.95 (m, 1H), 6.75 (m, 1H), 6.16 (m, 1H), 3.39 (m, 8H), 1.84 (m, 2H), 1.74-1.01 (m, 50H), 0.88 (m, 12H).

**$[P_{88810}]^+[DCI]^-$:** $[P_{88810}]^+[Br]^-$ was converted to its hydroxide form via an anion exchange column, followed by neutralization with 4,5-dicyanoimidazole. $^1H$ NMR (400 MHz, $CDCl_3$): $\delta_H$ ppm: 7.45 (s, 1H), 2.18 (m, 8H), 1.75-1.09 (m, 52H), 0.96 (m, 12H).

**$[N_{8888}]^+[TF_2N]^-$:** $[N_{8888}]^+[Br]^-$ was converted to its hydroxide form via an anion exchange column, followed by neutralization with bis(trifluoromethane)sulfonimide. $^1H$ NMR (400 MHz, $CDCl_3$): $\delta_H$ ppm: 3.06 (m, 8H), 1.96-0.98 (m, 48H), 0.96 (m, 12H).

### S11.3. Volatile Vapor Absorption Test.

The absorption capacity and absorption curve were collected from a gravimetric absorption instrument (Beishide, China). The IL sample was first dried in a vacuum oven at 70 °C for 24 hours. Subsequently, approximately 100 mg of the dried IL was placed into a quartz crucible inside the test chamber, forming a thin layer that covered the bottom of the crucible. An ion fan was used to eliminate electrostatic charges, and the test chamber was then sealed. Under atmospheric pressure at 100 °C, the system was purged with argon at a flow rate of 50 sccm for about 8 hours. During this period, the sample was weighed repeatedly until a constant mass was achieved, and the mass of the activated sample was recorded. High-purity argon was used as the carrier gas for the vapor generator to prepare adsorbate vapors at various volume fractions for the absorption tests. The structural schematic diagram of the absorption device is shown in Fig. S5.

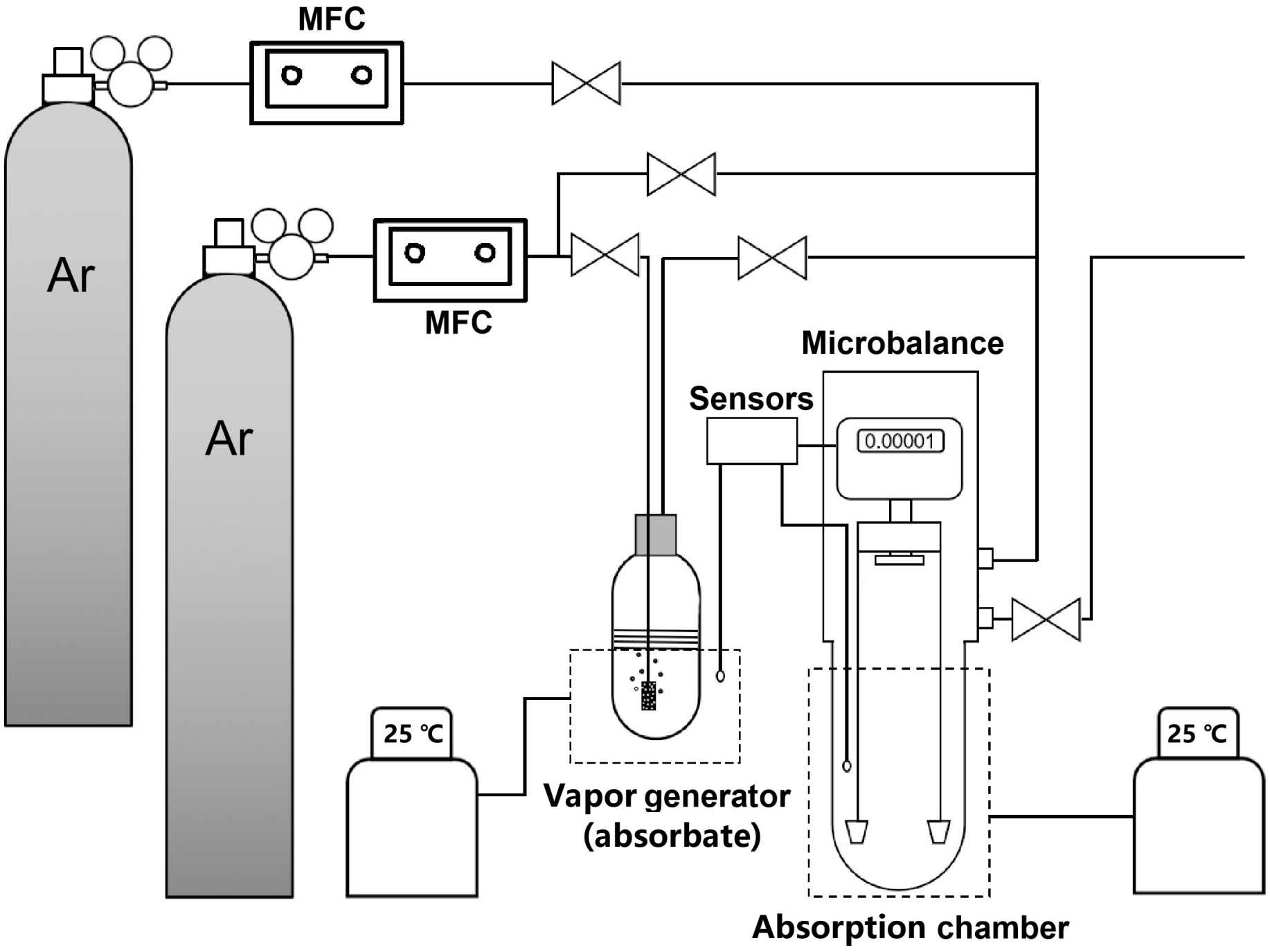

**Fig. S5. Schematic diagram of the vapor absorption device.**

**S11.4. Results of wet-lab experiments.**

The structures of all ions used in the experiments are shown in Table S17. The results of a series of absorption experiments are presented in Tables S18-S20 and S22 and Fig. S6, while the predicted CO2 solvation free energies are provided in Table S21. In addition, VOC absorption performance under controlled-variable conditions is compared in Tables S23 and S24.

**Table S17.** Abbreviations and Structures of the Ionic Liquids Used

| Abbreviation Name | Structure |
|---|---|
| $P_{CaCbCcCd}$ | 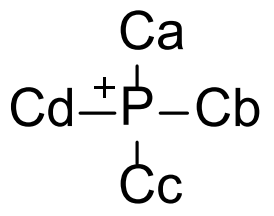 |
| **$N_{CaCbCcCd}$** | 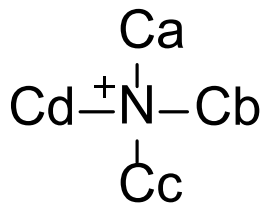 |
| **Camim** | 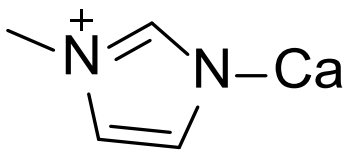 |
| **CaCbim** | 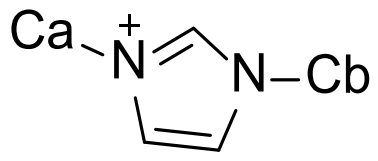 |
| **EMIM** | 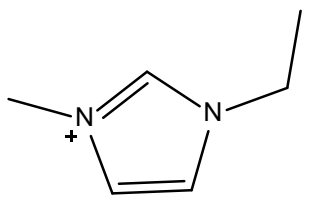 |
| **BMIM** | 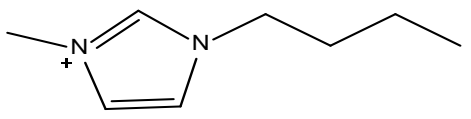 |
| **HMIM** | 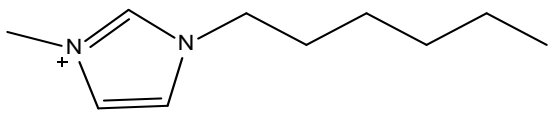 |
| **QUIN$_8$** | 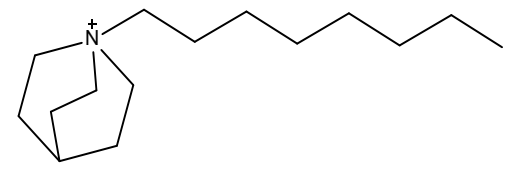 |
| **HMpy** | 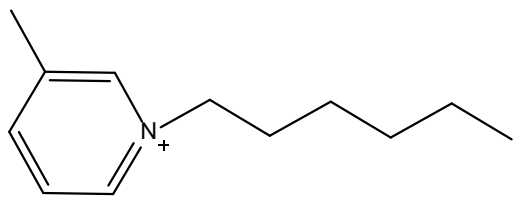 |
| **BMpy** | 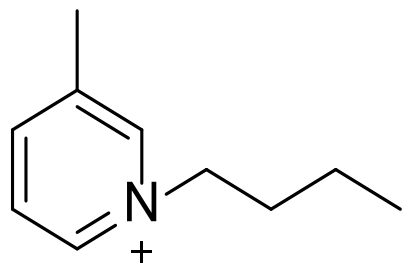 |
| **2-CNPyr** | 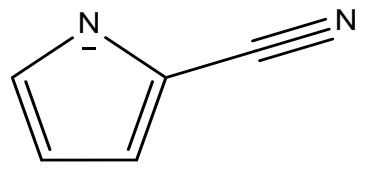 |
| **HX** | 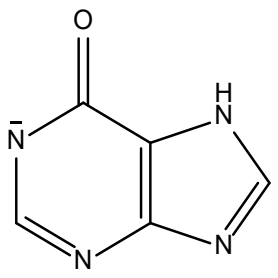 |
| **DCA** | 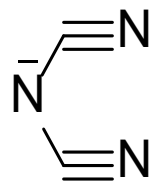 |
| **TCB** | 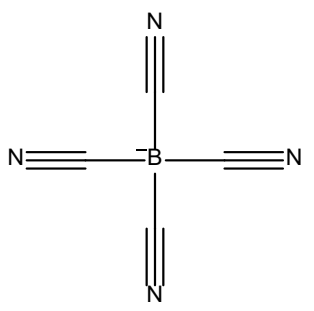 |
| **AC** | 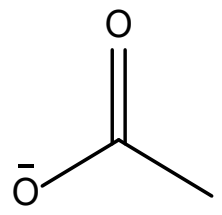 |
| **$TF_2N$** | 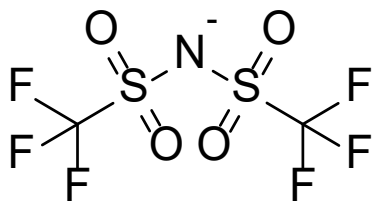 |
| **DCI** | 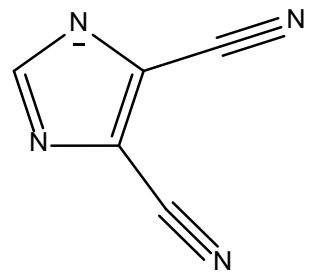 |
| **TDI** | 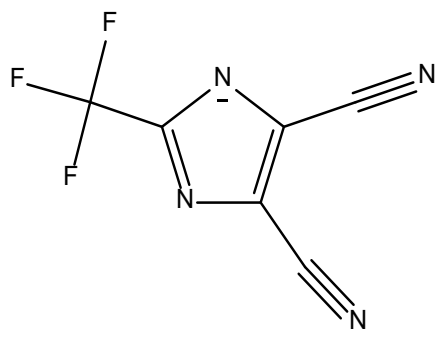 |
| **$PF_6$** | 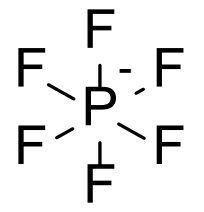 |

**Table S18.** Ethyl acetate absorption performance of designed ILs and the other materials reported in the literature. (Ethyl acetate Saturated vapor pressure $p^0$: 15 °C 7.60 kPa; 20 °C 9.84 kPa; 25 °C 12.62 kPa; 30 °C 16.01 kPa; 35 °C 20.12 kPa.)[100]

| No. | Name | Material Type | Absorption (mg $g^{-1}$) | Concentration | References |
|---|---|---|---|---|---|
| 1 | $[EMIM]^+[TCB]^-$ | ILs | 643 | $p/p^0$=0.8, 25°C | This work |
| 2 | $[HMIM]^+[TCB]^-$ | ILs | 938 | $p/p^0$=0.8, 25°C | This work |
| 3 | $[BMPy]^+[TDI]^-$ | ILs | 1978 | $p/p^0$=0.8, 25°C | This work |
| 4 | $[BMIM]^+[TDI]^-$ | ILs | 875 | $p/p^0$=0.8, 25°C | This work |
| 5 | $[HMIM]^+[TF_2N]^-$ | ILs | 665 | $p/p^0$=0.8, 25°C | This work |
| 6 | $[EMIM]^+[TCB]^-$ | ILs | 1483 | $p/p^0$=0.95, 25°C | This work |
| 7 | $[HMIM]^+[TCB]^-$ | ILs | 2045 | $p/p^0$=0.95, 25°C | This work |
| 8 | $[BMPy]^+[TDI]^-$ | ILs | 3501 | $p/p^0$=0.95, 25°C | This work |
| 9 | $[BMIM]^+[TDI]^-$ | ILs | 1942 | $p/p^0$=0.95, 25°C | This work |
| 10 | $[HMIM]^+[TF_2N]^-$ | ILs | 1486 | $p/p^0$=0.95, 25°C | This work |
| 11 | HCCP2-1.25 | HCPs | 1605 | saturated, 30°C (static vapor) | [101] |
| 12 | MIL-101(Cr) | MOFs | 925 | 54 mbar, 15°C (static vapor) | [102] |
| 13 | PHCP-3 | HCPs | 685 | saturated, 25°C (static vapor) | [103] |
| 14 | activated carbon | carbon | 730 | saturated, 30°C (static vapor) | [104] |
| 15 | silica gel | silica gel | 378 | 11.5 kPa, 25°C (static vapor) | [105] |
| 16 | SLC | ACPs | 222 | 120 ppm, 35°C | [106] |
| 17 | ZSM-5/MCM-41（C12） | Composite Porous Materials | 209 | 1600 mg/$m^3$, 30°C | [107] |
| 18 | Na-HZSM-5-750 | Composite Porous Materials | 161 | saturated, T not specified (static vapor) | 108 |
| 19 | commercial activated carbon | carbon | 444 | 500 mg/m$^3$, 25°C | 109 |
| 20 | NDA-88 | HCPs | 152 | 2500mg/m$^3$, 20°C | 110 |

**Table S19.** Acetone absorption performance of designed ILs and the other materials reported in the literature. (Acetone Saturated vapor pressure $p^0$: 15 °C 19.61 kPa; 20 °C 24.60 kPa; 25 °C 30.60 kPa; 30 °C 37.75 kPa; 35 °C 46.22 kPa.)

| No. | Name | Material Type | Absorption (mg g$^{-1}$) | Concentration | References |
|---|---|---|---|---|---|
| 1 | $[EMIM]^+[TCB]^-$ | ILs | 878 | $p/p^0$=0.8, 25°C | This work |
| 2 | $[HMIM]^+[TCB]^-$ | ILs | 1227 | $p/p^0$=0.8, 25°C | This work |
| 3 | $[BMIM]^+[TDI]^-$ | ILs | 1397 | $p/p^0$=0.8, 25°C | This work |
| 4 | $[EMIM]^+[TF_2N]^-$ | ILs | 865 | $p/p^0$=0.8, 25°C | This work |
| 5 | $[EMIM]^+[TCB]^-$ | ILs | 2581 | $p/p^0$=0.95, 25°C | This work |
| 6 | $[HMIM]^+[TCB]^-$ | ILs | 3540 | $p/p^0$=0.95, 25°C | This work |
| 7 | $[BMIM]^+[TDI]^-$ | ILs | 3560 | $p/p^0$=0.95, 25°C | This work |
| 8 | $[EMIM]^+[TF_2N]^-$ | ILs | 2746 | $p/p^0$=0.95, 25°C | This work |
| 9 | UC5 | carbon | 2607 | $p/p^0$=0.98, 25°C (static vapor) | 111 |
| 10 | NaOCs | carbon | 2010 | 18 kPa, 25°C (static vapor) | 112 |
| 11 | KAC | carbon | 1874 | $p/p^0$=0.99, 25°C | 113 |
| 12 | OCF-N-0.5h | carbon | 1208 | $p/p^0$=0.98, 25°C (static vapor) | 114 |
| 13 | OCF-N-1h | carbon | 1206 | $p/p^0$=0.98, 25°C (static vapor) | 114 |
| 14 | UC800 | carbon | 1522 | 18 kPa, 25°C (static vapor) | 115 |
| 15 | HPC800 | carbon | 1440 | 18 kPa, 25°C (static vapor) | 116 |
| 16 | 20%Mg@MIL-101/PC | hybrid materials | 1309 | 18 kPa, 15°C (static vapor) | 117 |
| 17 | MIL-101 | MOFs | 1291 | $p/p^0$=0.55, 25°C (static vapor) | 118 |
| 18 | GrO@MIL-101 | hybrid materials | 1167 | 16.18 kPa, 15°C (static vapor) | 119 |
| 19 | NC900 | carbon | 1221 | 18 kPa, 25°C (static vapor) | 120 |
| 20 | MIL-101/TC-40 | hybrid materials | 1137 | 18.9 kPa, 15°C (static vapor) | 121 |

**Table S20.** o-Xylene absorption performance of designed ILs and the other materials reported in the literature. (o-Xylene Saturated vapor pressure $p^0$: 15 ℃ 0.475 kPa; 20 ℃ 0.650 kPa; 25 ℃ 0.879 kPa; 30 ℃ 1.176 kPa; 35 ℃ 1.557 kPa.)

| No. | Name | Material Type | Absorption (mg $g^{-1}$) | Concentration | References |
|---|---|---|---|---|---|
| 1 | $[P_{88810}]^+[DCI]^-$ | ILs | 746 | $p/p^0$=0.8, 25℃ | This work |
| 2 | $[QUIN_8]^+[TF_2N]^-$ | ILs | 501 | $p/p^0$=0.8, 25℃ | This work |
| 3 | $[N_{8888}]^+[TF_2N]^-$ | ILs | 716 | $p/p^0$=0.8, 25℃ | This work |
| 4 | $[P_{88810}]^+[DCI]^-$ | ILs | 3184 | $p/p^0$=0.95, 25℃ | This work |
| 5 | $[QUIN_8]^+[TF_2N]^-$ | ILs | 1755 | $p/p^0$=0.95, 25℃ | This work |
| 6 | $[N_{8888}]^+[TF_2N]^-$ | ILs | 2024 | $p/p^0$=0.95, 25℃ | This work |
| 7 | $[P_{8883}]^+[BF_4]^-$ | ILs | 1462 | saturated, 25℃ (static vapor) | [122] |
| 8 | $[P_{8883}]^+[TF_2N]^-$ | ILs | 1225 | saturated, 25℃ (static vapor) | [122] |
| 9 | sql-1-Co-NCS | MOFs | 849 | $p/p^0$=0.95, 25℃ (static vapor) | [123] |
| 10 | Zn-ETTOB | MOFs | 691 | 0.8 kPa, 25℃ (static vapor) | [124] |
| 11 | HDTMAC | organoclays | 604 | $p/p^0$=0.7, 20℃ (static vapor) | [125] |
| 12 | silica gel | silica gel | 325 | 790 Pa, 25℃ (static vapor) | [126] |
| 13 | X-ddi-2-Ni | coordination networks | 514 | $p/p^0$=0.95, 25℃ (static vapor) | [127] |
| 14 | ClCTF-1-400 | COFs | 500 | $p/p^0$=1, 25℃ (static vapor) | [128] |
| 15 | SAMM-3-Cu-OTf | Werner complexes | 420 | $p/p^0$=0.9, 25℃ (static vapor) | [129] |
| 16 | SA−0.1 | silica-supported acids | 946 | 7.5 mg/L, 150℃ (reactive absorption) | [130] |
| 17 | SSSA-20 | silica supported acids | 414 | 7000 mg/$m^3$, 140℃ (reactive absorption) | [131] |
| 18 | CTF-2-400 | COFs | 358 | $p/p^0$=1, 25℃ (static vapor) | [132] |

**Table S21.** Predicted $CO_2$ solvation free energies in $[EMIM]^+$-based ILs with anions containing different numbers of cyano groups.

| Cyano-group count | Name | Predicted solvation ΔG (kcal/mol) | References |
|---|---|---|---|
| 1 | $[SCN]^-$ | −0.596 | [133] |
| 2 | $[DCA]^-$ | −0.734 | [133,134] |
| 3 | $[TCM]^-$ | −1.369 | [133,134] |
| - | $[TF_2N]^-$ | −1.635 | [134,135] |
| 4 | $[TCB]^-$ | −1.720 | [135] |

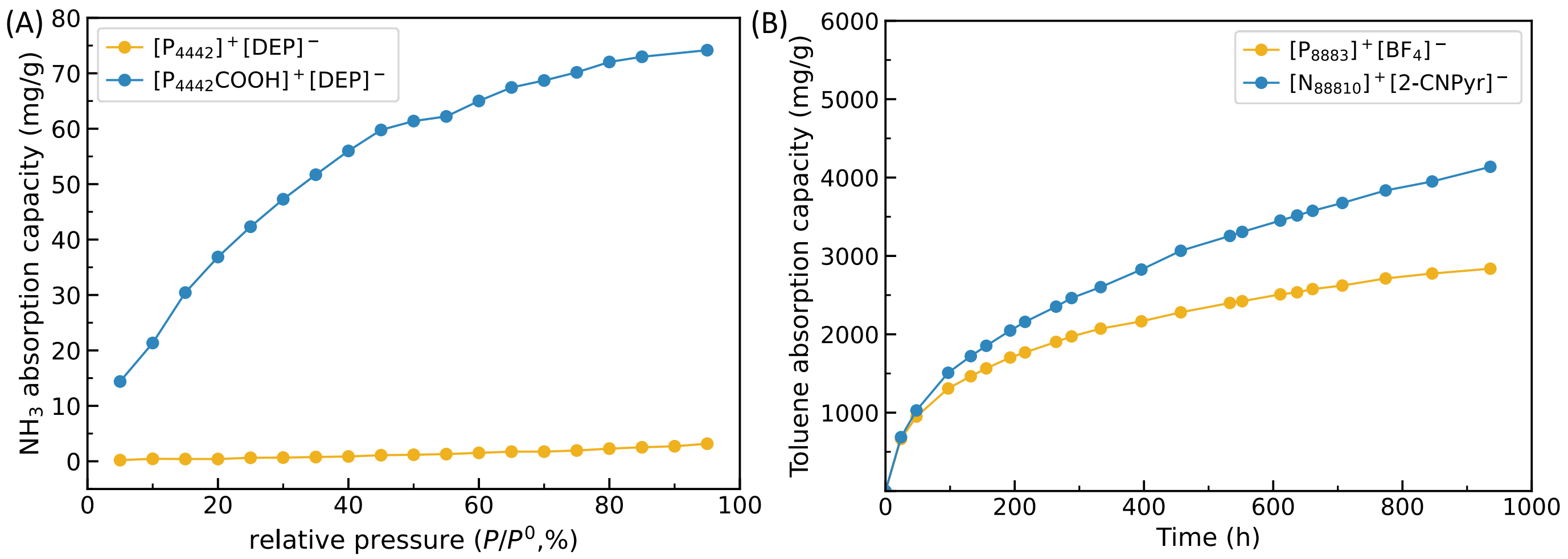

**Fig. S6. Gas/vapor absorption curves.** (A) Dynamic absorption curves of ammonia in ILs as a function of partial pressure. (B) Static saturated absorption of toluene at $p/p^0=1$. This experiment reproduced the original experimental conditions reported in the literature. When $[P_{8883}]^+[BF_4]^-$ approached saturation[136], with an absorption capacity of 2837 mg $g^{-1}$, our measured absorption capacity reached 4137 mg $g^{-1}$ and still showed no clear indication of saturation. Due to time constraints, the ultimate saturated absorption capacity could not be determined.

**Table S22.** Toluene absorption performance of designed ILs and the other ILs reported in the literature. (Toluene Saturated vapor pressure $p^0$: 15 °C 2.21 kPa; 20 °C 2.91 kPa; 25 °C 3.79 kPa; 30 °C 4.89 kPa; 35 °C 6.23 kPa.)

| No. | Name | Material Type | Absorption (mg $g^{-1}$) | test conditions | References |
|---|---|---|---|---|---|
| 1 | $[P_{88810}]^+[2\text{-}CNPyr]^-$ | ILs | 1309.6 | $p/p^0=0.95$, 25°C | this work (round1) |
| 2 | $[P_{88810}]^+[DCI]^-$ | ILs | 992 | $p/p^0=0.95$, 25°C | this work (round1) |
| 3 | $[P_{44414}]^+[DCI]^-$ | ILs | 871.4 | $p/p^0=0.95$, 25°C | this work (round1) |
| 4 | $[P_{88810}]^+[HX]^-$ | ILs | 851.8 | $p/p^0=0.95$, 25°C | this work (round1) |
| 5 | $[QUIN_8]^+[TF_2N]^-$ | ILs | 659.1 | $p/p^0=0.95$, 25°C | this work (round1) |
| 6 | $[N_{88810}]^+[2\text{-}CNPyr]^-$ | ILs | 2349.6 | $p/p^0=0.95$, 25°C | this work (round2) |
| 7 | $[P_{66614}]^+[Cl]^-$ | ILs | 1418 | $p/p^0=0.95$, 25°C | this work |
| 8 | $[N_{8881}]^+[HSO_4]^-$ | ILs | 149.2 | $p/p^0=0.95$, 25°C | this work |
| 9 | $[P_{4446}]^+[Br]^-$ | ILs | 876.8 | $p/p^0=0.95$, 25°C | this work |
| 10 | $[N_{1111}]^+[Su]^-$ | ILs | 1.85 | $p/p^0=0.95$, 25°C | this work |
| 11 | $[N_{1112OH}]^+[Cl]^-$ | ILs | 1.28 | $p/p^0=0.95$, 25°C | this work |
| 12 | $[N_{2222}]^+[Cl]^-$ | ILs | 8.53 | $p/p^0$=0.95, 25°C | this work |
| 13 | $[C_{10}MIM]^+[DCA]^-$ | ILs | 868.5 | $p/p^0$=0.95, 25°C | this work |
| 14 | $[C_8MIM]^+[PF_6]^-$ | ILs | 507.7 | $p/p^0$=0.95, 25°C | this work |
| 15 | $[BMIM]^+[TF_2N]^-$ | ILs | 342.3 | $p/p^0$=0.95, 25°C | this work |
| 16 | $[EMIM]^+[Ac]^-$ | ILs | 93.4 | $p/p^0$=0.95, 25°C | this work |
| 17 | $[P_{8883}]^+[BF_4]^-$ | ILs | 1802.4 | $p/p^0$=0.95, 25°C | this work (reproduced)[136] |
| 18 | $[P_{8883}]^+[TF_2N]^-$ | ILs | 847.6 | $p/p^0$=0.95, 25°C | this work (reproduced)[136] |
| 19 | $[P_{4443}]^+[BF_4]^-$ | ILs | 718.9 | $p/p^0$=0.95, 25°C | this work (reproduced)[136] |
| 20 | $[EMIM]^+[BF_4]^-$ | ILs | 120 | $p/p^0$=0.9, 25°C | [137] |
| 21 | $[BMIM]^+[BF_4]^-$ | ILs | 240 | $p/p^0$=0.9, 25°C | [137] |
| 22 | $[HMIM]^+[BF_4]^-$ | ILs | 380 | $p/p^0$=0.9, 25°C | [137] |
| 23 | $[BMIM]^+[I]^-$ | ILs | 170 | $p/p^0$=0.9, 25°C | [137] |
| 24 | $[BMIM]^+[PF_6]^-$ | ILs | 290 | $p/p^0$=0.9, 25°C | [137] |
| 25 | $[BMIM]^+[HSO_4]^-$ | ILs | 70 | $p/p^0$=0.9, 25°C | [137] |
| 26 | $[BMIM]^+[Ac]^-$ | ILs | 400 | $p/p^0$=0.9, 25°C | [137] |
| 27 | $[C_{10}MIM]^+[TF_2N]^-$ | ILs | 330 | $p/p^0$=0.7, 20°C | [138] |
| 28 | $[C_8MIM]^+[TF_2N]^-$ | ILs | 311 | $p/p^0$=0.7, 20°C | [138] |
| 29 | $[HMIM]^+[TF_2N]^-$ | ILs | 286 | $p/p^0$=0.7, 20°C | [138] |
| 30 | $[BMIM]^+[TF_2N]^-$ | ILs | 222 | $p/p^0$=0.7, 20°C | [138] |
| 31 | $[EMIM]^+[TF_2N]^-$ | ILs | 174 | $p/p^0$=0.7, 20°C | [138] |
| 32 | $[C_{10}MIM]^+[BF_4]^-$ | ILs | 249 | $p/p^0$=0.7, 20°C | [138] |
| 33 | $[EMIM]^+[EtSO_4]^-$ | ILs | 82 | $p/p^0$=0.7, 20°C | [138] |
| 34 | $[BMIM]^+[BF_4]^-$ | ILs | 109 | $p/p^0$=0.7, 20°C | [138] |
| 35 | $[EMIM]^+[MeSO_3]^-$ | ILs | 25.1 | $p/p^0$=1, 30°C | [139] |
| 36 | $[EMIM]^+[MeSO_4]^-$ | ILs | 33.9 | $p/p^0$=1, 30°C | [139] |
| 37 | $[EMIM]^+[BF_4]^-$ | ILs | 37.9 | $p/p^0$=1, 30°C | [139] |
| 38 | $[EMIM]^+[EtSO_4]^-$ | ILs | 40.1 | $p/p^0$=1, 30°C | [139] |
| 39 | $[EMIM]^+[Phenoxy]^-$ | ILs | 63.7 | $p/p^0$=1, 30°C | [139] |
| 40 | $[EMIM]^+[SCN]^-$ | ILs | 65.5 | $p/p^0$=1, 30°C | [139] |
| 41 | $[EMIM]^+[Va]^-$ | ILs | 80.1 | $p/p^0$=1, 30°C | [139] |
| 42 | $[EMIM]^+[DCA]^-$ | ILs | 91.9 | $p/p^0$=1, 30°C | [139] |
| 43 | $[EMIM]^+[DEtPO_4]^-$ | ILs | 102.8 | $p/p^0$=1, 30°C | [139] |
| 44 | $[EMIM]^+[TF_2N]^-$ | ILs | 118.6 | $p/p^0$=1, 30°C | [139] |
| 45 | $[EMIM]^+[Ca]^-$ | ILs | 143.9 | $p/p^0$=1, 30°C | [139] |
| 46 | $[EMIM]^+[Pe]^-$ | ILs | 152.7 | $p/p^0=1$, 30°C | 139 |
| 47 | $[EMIM]^+[De]^-$ | ILs | 173.3 | $p/p^0=1$, 30°C | 139 |

**Table S23.** toluene absorption performance of ILs with $[2\text{-CNPyr}]^-$ as the anion.

| No. | Name | Material Type | Absorption (mg $g^{-1}$) | test conditions | References |
|---|---|---|---|---|---|
| 1 | $[P_{88810}]^+[2\text{-CNPyr}]^-$ | ILs | 1309.6 | $p/p^0=0.95$, 25°C | this work (round1) |
| 2 | $[N_{88810}]^+[2\text{-CNPyr}]^-$ | ILs | 2349.6 | $p/p^0=0.95$, 25°C | this work (round2) |
| 3 | $[P_{8883}]^+[2\text{-CNPyr}]^-$ | ILs | 1672.6 | $p/p^0=0.95$, 25°C | this work |
| 4 | $[C_{10}MIM]^+[2\text{-CNPyr}]^-$ | ILs | 1029.1 | $p/p^0=0.95$, 25°C | this work |

**Table S24.** performance of ILs screened from o-xylene absorption for p-xylene absorption.

| No. | Name | Material Type | Absorption (mg $g^{-1}$) | Concentration | References |
|---|---|---|---|---|---|
| 1 | $[P_{88810}]^+[DCI]^-$ | ILs | 693 | $p/p^0=0.8$, 25°C | This work |
| 2 | $[QUIN_8]^+[TF_2N]^-$ | ILs | 480 | $p/p^0=0.8$, 25°C | This work |
| 3 | $[N_{8888}]^+[TF_2N]^-$ | ILs | 718 | $p/p^0=0.8$, 25°C | This work |

## S12. Physical origin of selective VOC absorption.

To elucidate the microscopic interactions underlying the experimentally observed absorption trends, nine AIonopedia-prioritized ionic solvents were selected from the candidate set in Fig. 5C after excluding the relatively common $[TF_2N]^-$-based candidates. These ionic solvents were paired with the corresponding VOC analogue series to construct 90 IL-VOC systems for atomistic molecular simulations and quantum-chemical calculations. Specifically, four ILs were selected from the ethyl acetate and acetone discovery tasks: $[HMIM]^+[TCB]^-$, $[BMPy]^+[TDI]^-$, $[BMIM]^+[TDI]^-$ and $[EMIM]^+[TCB]^-$. Each IL was paired with five ethyl acetate analogues and five acetone analogues, yielding $4 \times (5 + 5) = 40$ carbonyl-associated systems. Likewise, five ILs were selected from the o-xylene and toluene discovery tasks: $[P_{88810}]^+[DCI]^-$, $[P_{44414}]^+[DCI]^-$, $[P_{88810}]^+[2\text{-CNPyr}]^-$, $[P_{88810}]^+[HX]^-$ and $[N_{88810}]^+[2\text{-CNPyr}]^-$. Each IL was paired with five o-xylene analogues and five toluene analogues, yielding $5 \times (5 + 5) = 50$ aromatic-associated systems. Together, these two subsets comprised 90 systems, as summarized in Table S25.

**Table S25.** The 90 IL-VOC systems included in the atomistic molecular simulations and quantum-chemical calculations: four carbonyl-associated IL candidates paired with ten carbonyl analogues, and five aromatic-associated candidates paired with ten aromatic analogues.

| Category | VOC SMILES | IL Name | IL SMILES |
|---|---|---|---|
| carbonyl-VOC candidates | ethyl acetate: CCOC(C)=O analogues | $[HMIM]^+[TCB]^-$ | CCCCCC[n+]1ccn(C)c1.N#C[B-](C#N)(C#N)C#N |
| | | $[BMPy]^+[TDI]^-$ | CCCC[n+]1cccc(C)c1.N#Cc1nc(C(F)(F)F)[n-]c1C#N |
| | acetone: CC(C)=O analogues | $[BMIM]^+[TDI]^-$ | CCCC[n+]1ccn(C)c1.N#Cc1nc(C(F)(F)F)[n-]c1C#N |
| | | $[EMIM]^+[TCB]^-$ | CC[n+]1ccn(C)c1.N#C[B-](C#N)(C#N)C#N |
| aromatic-VOC candidates | o-xylene: Cc1ccccc1C analogues | $[P_{88810}]^+[DCI]^-$ | CCCCCCCCCC[P+](CCCCCCCC)(CCCCCCCC)CCCCCCCC.N#Cc1nc[n-]c1C#N |
| | | $[P_{44414}]^+[DCI]^-$ | CCCCCCCCCCCCCC[P+](CCCC)(CCCC)CCCC.N#Cc1nc[n-]c1C#N |
| | | $[P_{88810}]^+[2\text{-CNPyr}]^-$ | CCCCCCCCCC[P+](CCCCCCCC)(CCCCCCCC)CCCCCCCC.N#Cc1ccc[n-]1 |
| | toluene: Cc1ccccc1 analogues | $[P_{88810}]^+[HX]^-$ | CCCCCCCCCC[P+](CCCCCCCC)(CCCCCCCC)CCCCCCCC.O=c1[n-]cnc2nc[nH]c12 |
| | | $[N_{88810}]^+[2\text{-CNPyr}]^-$ | CCCCCCCCCC[N+](CCCCCCCC)(CCCCCCCC)CCCCCCCC.N#Cc1ccc[n-]1 |

### S12.1. Atomistic Modelling.

As local solvent organization around the solute critically influences intermolecular interactions, equilibrated solvation structures were generated through a nonequilibrium growth protocol rather than conventional insertion. We construct all-atom models with fixed-charge force fields similar to the solvation free energy calculation protocol detailed in Section S5. Specifically, we employed RESP charges and GAFF2 bonded and vdW parameters for the solute and ion pairs, used a charge scaling factor 0.8 for IL-forming ions, and packed the IL simulation cell containing approximately 11000 atoms. The IL cell is then relaxed for 50 ns to get an equilibrated configuration. We then use a strategy similar to the nonequilibrium solute creation procedure to get the equilibrated solute-solvent configurations. Specifically, the solute molecule is put at the center of the IL cell with no solute-solvent interaction. Then, the decoupled solute-solvent vdW interaction is switched on in 2 ns, after which an additional 1 ns Hamiltonian-switching simulation is conducted to turn on the solute-solvent electrostatics. After that, an additional 1 ns equilibration is conducted to reach an equilibrated state. This solute growth procedure is repeated 5 times, generating 5 independent configurations of the solute-solvent system. This procedure substantially accelerates relaxation of the first solvation shell while avoiding unfavorable initial contacts that frequently occur in direct insertion simulations.

### S12.2. Construction of representative IL-solute clusters

For each IL-solute system, five MD snapshots were used to construct representative solute-centered clusters. Ion-solute distances were evaluated between heavy atoms using the minimum-image convention. Pairwise contacts were weighted using the PLUMED RATIONAL switching function with $D_0 = 0$, $R_0 = D_{\max} = 6.0$ Å, $n = 6$, and $m = 12$, smoothly truncated at $D_{\max}$.[140]

Let $H_{\text{sol}}$ and $H_i$ denote the heavy-atom sets of the central solute and ion $i$, respectively, and let $r_{\alpha,ab}$ denote the minimum-image distance between solute atom $a$ and ion atom $b$ in snapshot $\alpha$. Denoting the corresponding RATIONAL contact weight by $s(r_{\alpha,ab})$, the unnormalized ion-solute contact score was defined as

$$C_{\alpha,i} = \sum_{a\in H_{\text{sol}}} \sum_{b\in H_i} s(r_{\alpha,ab}) \quad \text{(S34).}$$

Because additive atom-pair sums favor larger molecular ions, ions were ranked using a per-heavy-atom score analogous to size-normalized screening metrics[141]

$$Q_{\alpha,i} = \frac{C_{\alpha,i}}{|H_i|} \quad \text{(S35),}$$

where $|H_i|$ is the number of heavy atoms in ion $i$.

For a given cluster size $N$, all cations and anions were ranked together by $Q_{\alpha,i}$, and the unconstrained Top-$N$ selection was obtained independently for each snapshot. To maintain a common cation:anion composition across the five snapshots, let $k_\alpha$ denote the number of cations in the unconstrained selection and $K = \{k_\alpha \mid \alpha = 1, \ldots, 5\}$ the set of observed cation counts. For each $k \in K$, $A_{\alpha,k}$ comprised the $k$ highest-ranked cations and $N - k$ highest-ranked anions. The retained fraction of the total unnormalized contact score was

$$F_{\alpha,k} = \frac{\sum_{i \in A_{\alpha,k}} C_{\alpha,i}}{\sum_{j \in I_\alpha} C_{\alpha,j}} \qquad \text{(S36)},$$

where $I_\alpha$ denotes all candidate ions in snapshot $\alpha$. The common cation count was chosen to maximize the mean coverage across the five snapshots

$$k^* = \arg\max_{k \in K} \left( \frac{1}{5} \sum_{\alpha=1}^{5} F_{\alpha,k} \right) \qquad \text{(S37)}.$$

As a precaution against potential convergence difficulties in strongly charge-imbalanced clusters and to reduce the computational cost of subsequent Density-Functional-Theory (DFT)-based Energy Decomposition Analysis (EDA), the preliminary composition of $k^*$ cations and $N - k^*$ anions was further required to satisfy $|2k^* - N| \leq 2$. This restricted the magnitude of the cluster net charge to at most $2e$ while preventing excessive enrichment of the generally larger cations, thereby limiting the total number of atoms. If this limit was exceeded, the cation count was clipped to the nearest allowed value, after which the highest-ranked cations and anions were reselected and the contact coverage was recalculated. For the Top-8 clusters, the allowed cation:anion compositions were therefore 3:5, 4:4, and 5:3.

The complete selection procedure, including the composition constraint, was applied independently at each $N = 2, 4, \ldots, 16$, yielding an overall mean coverage curve with a geometric elbow at $N = 8$, corresponding to 90.1% coverage (Fig. S7 A-C). The effects of the composition constraint on ion composition and total cluster atom counts are summarized in Fig. S7 D-F. For output, each selected ion was rigidly translated by a periodic lattice vector to the image that minimized its closest heavy-atom distance to the central solute, while its internal coordinates remained unchanged.

Overall, the cluster-selection protocol was designed to balance structural fidelity to the original solvation environment with computational tractability, while ensuring consistent and comparable treatment across all IL-solute systems.

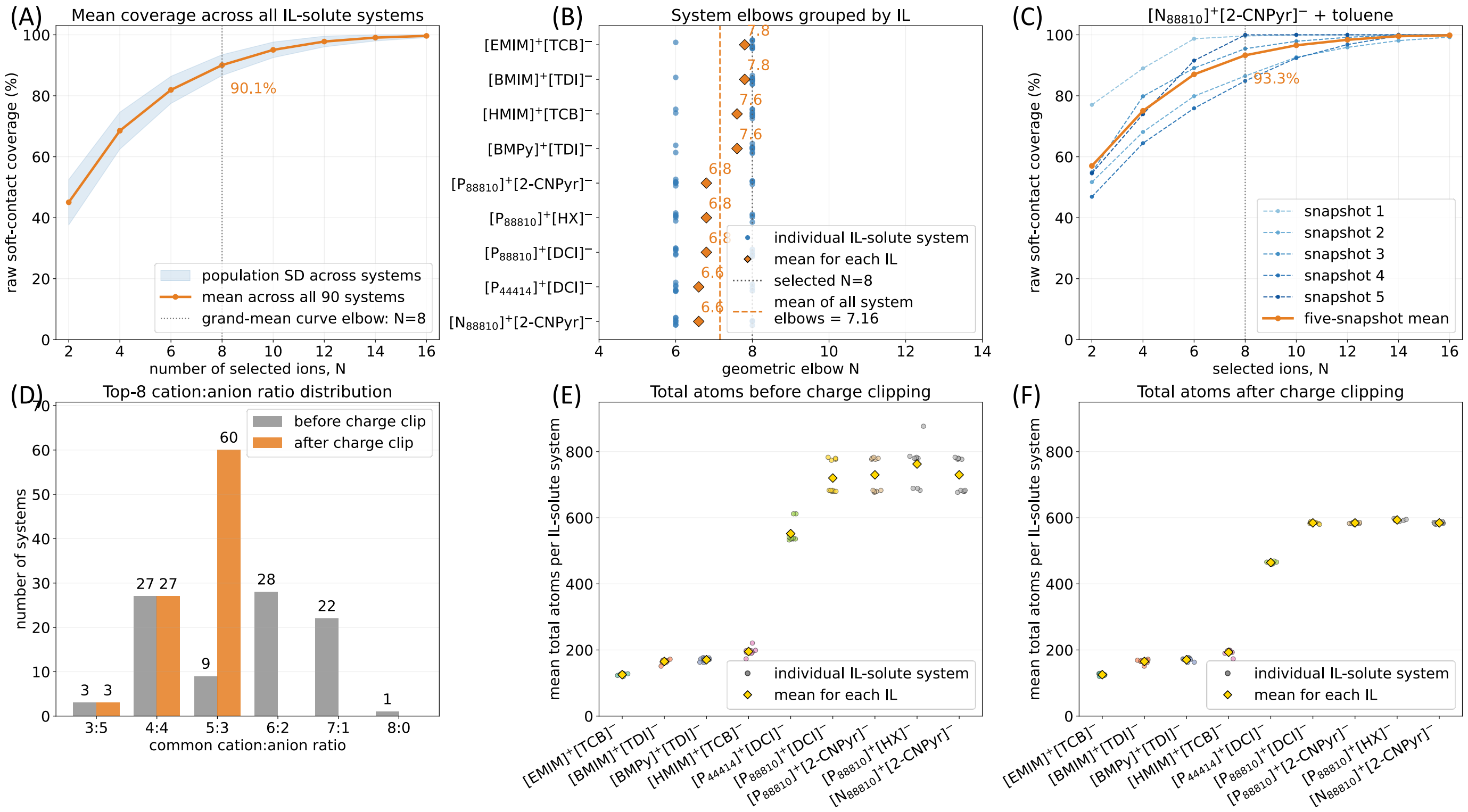

**Fig. S7. Selection of cluster size and ion composition for IL-solute systems.** Coverage curves were calculated for $N = 2,4,\ldots,16$, using a common ion composition across the five snapshots of each system and imposing $|n_{cation} - n_{anion}| \leq 2$ at each $N$. The geometric elbow was defined as the point of maximum departure from the chord connecting the two endpoints after min-max scaling of both axes.[142] (A) Mean soft-contact coverage across 90 IL-solute systems, with shading indicating ±1 population SD across the 90 system means, each averaged over five snapshots. The dotted line marks the elbow at $N = 8$. (B) System-level elbows grouped by IL. Circles and diamonds denote individual systems and IL means, respectively. The dashed line indicates the mean of all 90 system elbows (7.16). (C) Coverage curves for $[N_{88810}]^+[2\text{-CNPyr}]^-$ with toluene. Dashed curves represent the five snapshots, and the solid curve represents their mean. (D) Numbers of systems with each common Top-8 cation:anion ratio before and after charge clipping. (E,F) Total cluster atom counts before and after charge clipping, respectively, plotted on the same vertical scale. Circles and diamonds denote individual IL-solute systems and IL means, respectively.

### S12.3. DFT-based energy decomposition calculations

For each IL-VOC system, single-point DFT calculations were performed on the five clusters described in Section S12.2. All atomic coordinates were retained at their MD geometries without further optimization.

Each cluster was divided into two fragments: the central VOC molecule (fragment A) and the complete eight-ion solvation cage (fragment B). This partition defines the interaction between the VOC and the surrounding ionic cage, with ion-ion interactions remaining internal to fragment B. The VOC fragment was neutral, and the ionic cage had a net charge of -2, 0 or +2, depending on its cation-to-anion composition. Restricted closed-shell singlet calculations were used for all fragments and complete clusters.

Electronic-structure calculations were performed with Gaussian at the B3LYP-D3(BJ)/6-31+G(d,p) level. Diffuse functions were included to describe the spatially extended electron density associated with the anionic constituents of the ionic cage.

Energy decomposition was performed using sobEDAw, a variant of the sobEDA method described below.[143] CounterPoise (CP) correction was applied by evaluating both fragment energies in the full basis set of the complete cluster.

The CP-corrected interaction energy was calculated as

$$\Delta E_{int}^{CP} = E_{AB}^{AB} - E_{A}^{AB} - E_{B}^{AB} \quad \text{(S38)}.$$

Here, $E_{AB}^{AB}$ denotes the complete-cluster energy, and $E_{A}^{AB}$ and $E_{B}^{AB}$ denote the corresponding fragment energies. All three energies were calculated at the same fixed geometry. This treatment reduces basis-set superposition error. The CP superscript is omitted below because all reported interaction energies include this correction.

The sobEDA method partitions the interaction energy into electrostatic, exchange, Pauli-repulsion, orbital-relaxation, DFT-correlation, and empirical dispersion-correction contributions. For the noncovalent interactions between each VOC and its ionic cage, we adopted sobEDAw, a variant specifically developed for weakly interacting systems. By redistributing the DFT-correlation contribution between the exchange-repulsion and dispersion terms, sobEDAw yields a compact four-component decomposition whose dispersion-to-electrostatics ratio closely agrees with that obtained using the substantially more expensive Symmetry-Adapted Perturbation Theory (SAPT).[143] This transformation requires no additional electronic-structure calculations, and the resulting components provide a physically interpretable basis for the interaction fingerprints used in the subsequent analyses. Using the redistribution coefficient $\omega$ defined below, sobEDA and sobEDAw are related by

$$\begin{aligned} \Delta E_{int} &= \Delta E_{els} + \Delta E_{x} + \Delta E_{rep} + \Delta E_{orb} + \Delta E_{DFTc} + \Delta E_{dc} \\ &= \Delta E_{els} + \left[\Delta E_{x} + \Delta E_{rep} + (1-\omega)\Delta E_{DFTc}\right] + \Delta E_{orb} + (\Delta E_{dc} + \omega\Delta E_{DFTc}) \\ &= \Delta E_{els} + \Delta E_{xrep}^{\omega} + \Delta E_{orb} + \Delta E_{disp}^{\omega} \quad \text{(S39)}. \end{aligned}$$

Here, $\Delta E_{els}$ describes the quasi-classical electrostatic interaction between the unperturbed fragment charge distributions. The exchange contribution, $\Delta E_x$ , arises from anti-symmetrization of the fragment wavefunctions, whereas $\Delta E_{rep}$ represents the energy penalty associated with orbital overlap and orthogonalization. The orbital contribution, $\Delta E_{orb}$, describes electronic relaxation from the frozen state to the self-consistent state of the complete cluster. It includes both intra-fragment polarization and inter-fragment charge transfer, which were not separated in this analysis. The remaining terms, $\Delta E_{DFTc}$ and $\Delta E_{dc}$, represent the DFT-correlation contribution and empirical D3(BJ) dispersion-correction contribution, respectively.

In the final line of Eq. (S39), $\Delta E^{\omega}_{xrep}$ and $\Delta E^{\omega}_{disp}$ denote the sobEDAw exchange-repulsion and dispersion components, respectively. The redistribution leaves the total interaction energy, electrostatic contribution and orbital contribution unchanged. For each cluster, the coefficient $\omega$ was calculated as

$$\omega = \min(1, c + (1-c)exp\left[-r(\frac{\Delta E_{dc}}{\Delta E_{els}} - a)\right]) \quad \text{(S40).}$$

The published parameters for B3LYP-D3(BJ)/6-31+G(d,p) with CP correction were used: $c = 0.575$, $a = 0.071$ , $r = 2.571$ .[143] Negative energy contributions indicate stabilization, whereas positive contributions indicate destabilization. The resulting fixed-geometry interaction energies describe local electronic stabilization and do not represent absorption free energies, which also depend on entropy and the surrounding bulk solvent.

### S12.4. Physical interpretation of AIonopedia-derived solvent recommendations using complementary interaction analyses

As described in Sections S12.1-S12.2, five snapshots were generated for each screened system and used to construct five corresponding conformational clusters. For the main analyses, one representative conformation was selected for each system by ranking the snapshot-derived clusters according to their coverage, with higher priority assigned to conformations whose coverage values were closer to the five-snapshot mean. As a robustness check, we repeated the analyses using quantities averaged over all five conformations for each system, yielding patterns consistent with those from the main analysis.

Using these representative systems, we first examined the energy distributions underlying the sobEDAw interaction fingerprints across the four VOC analogue series. The full distributions were consistent with the mean patterns presented in Fig. 5D and showed that these differences extended across individual systems (Fig. S8A). The total interaction-energy distributions were broadly similar, although the ethyl acetate series was shifted toward less negative values. The largest component-level differences occurred in dispersion and electrostatics, which showed opposing trends: dispersion was more stabilizing for the aromatic-associated systems, whereas electrostatics was more stabilizing for the carbonyl-associated systems. Exchange-repulsion and orbital-interaction energies also varied, but their distributions remained comparatively similar across the four series.

To assess whether the block-diagonal organization observed in Fig. 5E depended on normalization, we recalculated the pairwise distances using both raw and feature-wise z-score-standardized sobEDAw interaction fingerprints (Fig. S8B, top and middle). Both representations retained the broad block-diagonal separation between carbonyl- and aromatic-associated systems, although the blocks were less sharply resolved than in the L2-normalized map in Fig. 5E. The persistence of this organization across raw, z-score-standardized, and L2-normalized fingerprints indicates that it was not an artifact of a particular normalization scheme or the dominance of the larger-magnitude components. Notably, the L2-normalized final-token embeddings extracted immediately before the prediction FFN displayed an even clearer block-diagonal structure with substantially less within-block noise (Fig. S8B, bottom), suggesting that AIonopedia consolidates the family-level organization captured by the sobEDAw interaction fingerprints into a cleaner learned representation.

The corresponding Principal Component Analysis (PCA) projections provided a complementary view of the sobEDAw interaction fingerprints (Fig. S8C). The first two principal components together explained 93.7% and 97.1% of the variance in the z-score-standardized and raw fingerprints, respectively. Both projections preserved the broad separation between carbonyl- and aromatic-associated systems observed in Fig. 5F. Aromatic-associated systems still tended to form more compact groupings than carbonyl-associated systems, although this difference in grouping compactness was less pronounced than in the L2-normalized projection.

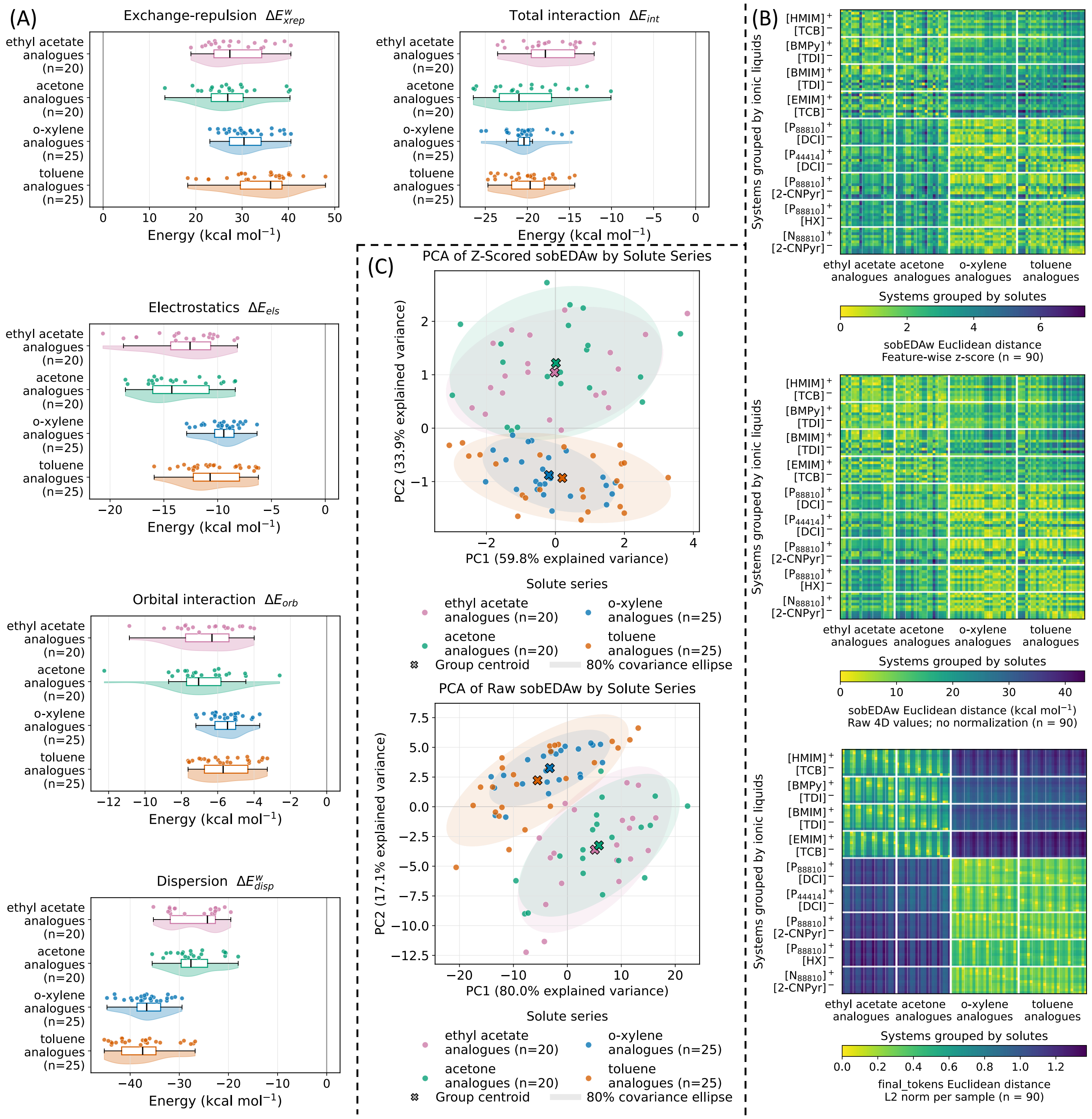

**Fig. S8. Energy distributions and complementary analyses of sobEDAw interaction fingerprints across VOC analogue series.** (A) Raincloud distributions of the total interaction energy and four sobEDAw components across the 90 IL-VOC systems, grouped by VOC analogue series. Points represent individual systems, whereas box plots and half violins summarize the distributions. (B) Pairwise Euclidean-distance heatmaps calculated using feature-wise z-score-standardized four-component sobEDAw interaction fingerprints (top), raw unnormalized fingerprints (middle), and per-system L2-normalized 512-dimensional AIonopedia final-token embeddings (bottom). Systems are grouped by ILs along rows and VOC analogue

series along columns; lower values indicate more similar fingerprints or embeddings. (C) PCA projections of the four-component sobEDAw interaction fingerprints after feature-wise z-score standardization (top) and without normalization (bottom), colored by VOC analogue series. Crosses and shaded regions indicate group centroids and 80% covariance ellipses, respectively.

Because sobEDAw quantifies the energetic components of solute-cage interactions without directly resolving their spatial distribution, we applied the Independent Gradient Model based on Hirshfeld partition (IGMH)[144] to two representative clusters corresponding to the carbonyl- and aromatic-associated systems, respectively, using the single-point wavefunctions described in Section S12.3. In the resulting maps, the $\delta g_{inter}$ isosurfaces identify interaction regions between the solute and the surrounding ionic cage, while the $\text{sign}(\lambda_2)\rho$ coloring qualitatively represents a progression from stronger attractive interactions (blue), through weak vdW interactions (green), to steric repulsion (red) (Fig. S9).

Under the same visualization settings, acetone in $[EMIM]^+[TCB]^-$ exhibits more prominent blue regions, with the principal iso-surfaces forming continuous patches (Fig. S9A). The blue region near the carbonyl oxygen lies between this oxygen and the imidazolium C2-H group, whose carbon is flanked by the two ring nitrogen atoms. The corresponding H···O distance is approximately 2.26 Å, and the C-H···O angle is approximately 155.5°. Together with the localized attractive region, this geometry supports the assignment of a C-H···O hydrogen-bonding contact.

By comparison, the iso-surfaces for toluene in $[N_{88810}]^+[2\text{-CNPyr}]^-$ are predominantly green, with more numerous, disconnected patches around the aromatic ring and methyl group, indicating spatially distributed weak vdW contacts (Fig. S9B). This diffuse interaction pattern is consistent with the greater dispersion contribution identified by sobEDAw for aromatic-associated systems.

Overall, the chemically intuitive sobEDAw component distributions and representative IGMH maps provide complementary energetic and spatial views of solute-ion interactions. The family-level correspondence between sobEDAw interaction fingerprints and AIonopedia-generated final-token embeddings further supports the conclusion that the IL discovery process is grounded in physically meaningful intermolecular interactions, linking data-driven candidate prioritization to the distinct solvation requirements of different VOC families.

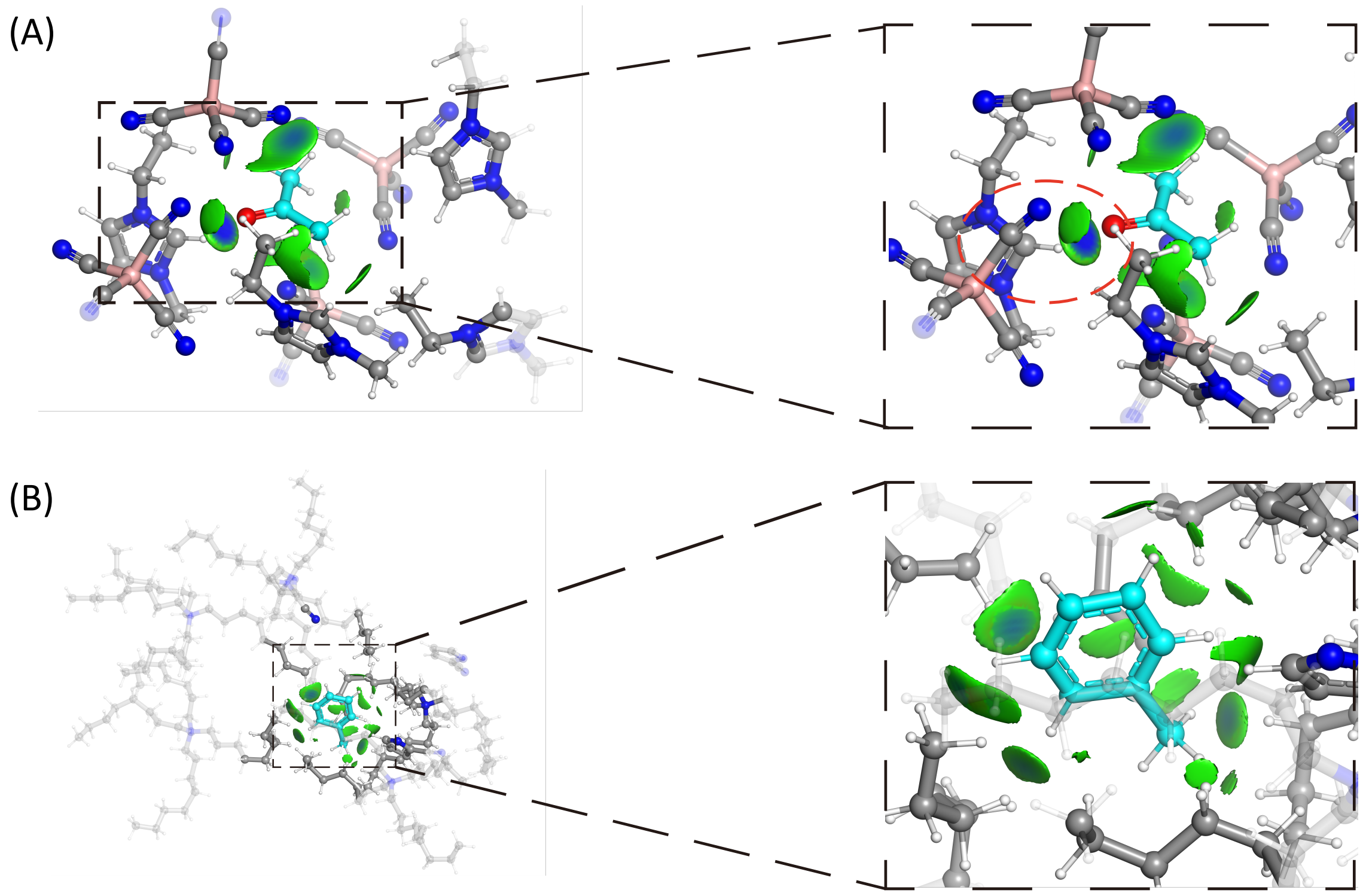

**Fig. S9. IGMH visualization of solute-ion interactions in representative solvation clusters.** (A) Acetone in $[EMIM]^+[TCB]^-$ and (B) toluene in $[N_{88810}]^+[2\text{-}CNPyr]^-$. Each panel shows the full cluster on the left and an enlarged view of the boxed region on the right. Solute carbon atoms are shown in cyan, and peripheral ions are rendered partially transparent. For the IGMH analysis, the solute constitutes one fragment, while all surrounding ions collectively constitute the other. The $\delta g_{inter} = 0.004$ a.u. isosurfaces are colored by $\text{sign}(\lambda_2)\rho$ over the range of -0.015 to +0.015 a.u., where $\rho$ is the electron density and $\lambda_2$ is the second eigenvalue of its Hessian matrix, ordered as $\lambda_1 \leq \lambda_2 \leq \lambda_3$. Blue, green, and red represent negative, near-zero, and positive values, respectively, corresponding qualitatively to stronger attractive interactions, weak vdW interactions, and steric repulsion. The red dashed ellipse marks the contact region between the acetone carbonyl oxygen and a neighboring imidazolium cation discussed in the text.

## S13. Example demonstrations of AIonopedia agent task execution and online deployment.

The complete agentic workflow employed in this study is illustrated through a series of demonstrations encompassing information retrieval and processing, property prediction, high-throughput screening, molecular refinement, and chemically grounded image-text reasoning (Figs. S10-S15). To further facilitate practical use of AIonopedia, we deployed a web-based interface that provides direct access to the trained property-prediction models (Fig. S16).

**(Prompt):** Please tell me the solvation free energy of 1-hexene in Nip311 and DEP at 298.15 K, as well as the transfer free energy of the system.

**Thought:** I need to identify the exact ions corresponding to "Nip311" and "DEP" and obtain their SMILES, as well as the solute "1-hexene", in order to compute solvation and transfer free energies at 298.15 K using the property calculator. I will first use the literature/web search tool to verify the meanings/abbreviations of "Nip311" and "DEP". Then I will convert names to SMILES with PubChem and run the property calculator.
**ToolCall:** web_searcher({'query': 'What is the abbreviation Nip311 in ionic liquids and what ion does DEP refer to?'})
**Observation[web_searcher]:** In ionic liquids, [Nip311]+ denotes the quaternary ammonium cation N-isopropyl-N-propyl-N,N-dimethylammonium, and DEP refers to the diethyl phosphate anion. Reference: Investigation of the Thermostability and Activity of Biomolecules in Bio-Friendly Deep Eutectic Solvents

**Thought:** I verified what Nip311 and DEP stand for. Next, I will fetch SMILES for the cation [Nip311]+ (N-isopropyl-N-propyl-N,N-dimethylammonium), the anion DEP (diethyl phosphate), and the solute 1-hexene using PubChem. Then I will canonicalize them and compute solvation and transfer free energies at 298.15 K using the property calculator.
**ToolCall:** pubchem_searcher({'query': 'N-isopropyl-N-propyl-N,N-dimethylammonium'})
**Observation[pubchem_searcher]:** [N+](C)(C)(CCC)C(C)C

**Thought:** I will retrieve SMILES for diethyl phosphate (DEP) and 1-hexene from PubChem in parallel to speed up, then canonicalize all SMILES before computing solvation and transfer free energies at 298.15 K.
**ToolCall:** pubchem_searcher({'query': 'diethyl phosphate anion'})
**Observation[pubchem_searcher]:** CCOP(=O)([O-])OCC
**ToolCall:** pubchem_searcher({'query': '1-hexene'})
**Observation[pubchem_searcher]:** C=CCCCC

**Thought:** I will compute the solvation free energy of 1-hexene in the ionic liquid [Nip311][DEP] at 298.15 K using the property calculator with the verified SMILES, then compute the transfer free energy (IL to water) for the same system at the same temperature.
**ToolCall:** property_predictor({'property_string': 'solvation', 'cation_SMILES': '[N+](C)(C)(CCC)C(C)C', 'anion_SMILES': 'CCOP(=O)([O-])OCC', 'solute_SMILES': 'C=CCCCC', 'solvent_SMILES': '', 'Temp': '298.15'})
**Observation[property_predictor]:** -1.7461 kcal/mol
**ToolCall:** property_predictor({'property_string': 'transfer', 'cation_SMILES': '[N+](C)(C)(CCC)C(C)C', 'anion_SMILES': 'CCOP(=O)([O-])OCC', 'solute_SMILES': 'C=CCCCC', 'solvent_SMILES': '', 'Temp': '298.15'})
**Observation[property_predictor]:** 4.0166 kcal/mol

**Final Answer:**
- Definitions verified via literature:
  - Nip311 denotes the quaternary ammonium cation N-isopropyl-N-propyl-N,N-dimethylammonium; SMILES: [N+](C)(C)(CCC)C(C)C.
  - DEP denotes the diethyl phosphate anion; SMILES: CCOP(=O)([O-])OCC.
  - Solute 1-hexene; SMILES: C=CCCCC.

- Predicted free energies at 298.15 K using the property calculator:
  - Solvation free energy of 1-hexene in [Nip311][DEP]: −1.7461 kcal/mol (gas → IL).
  - Transfer free energy (ionic liquid → water) for 1-hexene: +4.0166 kcal/mol.

**Fig. S10. Example demonstration of AIonopedia executing an IL-related molecular search and calculation task.** Starting from the abbreviation $[N_{ip311}]^+[DEP]^-$, AIonopedia searches for the full chemical name and corresponding SMILES representation, and then calculates the solvation free energy (gas to IL) of 1-hexene in this IL, as well as the corresponding transfer free energy (IL to water).

**(Prompt):** I have a specific solute, toluene, and I am looking for ionic liquids that offer the highest solubility and compatibility in a two-phase system of ionic liquid and air. Could you recommend a few optimal ionic liquids that meet these criteria?

**Thought:** I need the SMILES for the solute toluene to run the optimal ionic liquid search for maximum solubility (minimize solvation free energy from air to IL). I will first convert "toluene" to SMILES using PubChem. Then I will use the molecule_searcher with property_string="solvation", solute_SMILES of toluene, Max_or_Min="Min", from_solvation_dataset="split" to get the Top5 lists.
**ToolCall:** pubchem_searcher({'query': 'toluene'})
**Observation[**pubchem_searcher**]:** Cc1ccccc1
**ToolCall:** molecule_searcher({'property_string': 'solvation', 'cation_SMILES': '', 'anion_SMILES': '', 'solute_SMILES': 'Cc1ccccc1', 'Max_or_Min': 'Min', 'from_solvation_dataset': 'split'})
**Observation[**molecule_searcher**]:** Demo-style optimal ionic liquid search results. Preserve all three Top5 sections when reporting recommendations.

PubChem search results Top5:
1. cation: CCCCCCCCCC[P+](CCCCCCCC)(CCCCCCCC)CCCCCCCC anion: N#Cc1nc[n-]c1C#N prediction: -5.041kcal/mol
2. cation: CCCCCCCCCCCCCC[P+](CCCC)(CCCC)CCCC anion: N#Cc1nc[n-]c1C#N prediction: -4.998kcal/mol
3. cation: CCCCCCCCCC[P+](CCCCCCCC)(CCCCCCCC)CCCCCCCC anion: N#Cc1ccc[n-]1 prediction: -4.972kcal/mol
4. cation: CCCCCCCCCCCCCC[P+](CCCC)(CCCC)CCCC anion: O=c1[n-]cnc2nc[nH]c12 prediction: -4.947kcal/mol
5. cation: CCCCCCCCCC[P+](CCCCCCCC)(CCCCCCCC)CCCCCCCC anion: O=c1[n-]cnc2nc[nH]c12 prediction: -4.938kcal/mol

Original dataset Top5:
1. cation: CCCCCCCCCC[P+](CCCCCCCC)(CCCCCCCC)CCCCCCCC anion: N#Cc1nc[n-]c1C#N prediction: -5.041kcal/mol
2. cation: CCCCCCCCCCCCCC[P+](CCCC)(CCCC)CCCC anion: N#Cc1nc[n-]c1C#N prediction: -4.998kcal/mol
3. cation: CCCCCCCCCC[P+](CCCCCCCC)(CCCCCCCC)CCCCCCCC anion: N#Cc1ccc[n-]1 prediction: -4.972kcal/mol
4. cation: CCCCCCCCCCCCCC[P+](CCCC)(CCCC)CCCC anion: O=c1[n-]cnc2nc[nH]c12 prediction: -4.947kcal/mol
5. cation: CCCCCCCCCC[P+](CCCCCCCC)(CCCCCCCC)CCCCCCCC anion: O=c1[n-]cnc2nc[nH]c12 prediction: -4.938kcal/mol

Solvation dataset Top5:
1. cation: CCCC[n+]1cccc(C)c1 anion: N#Cc1nc(C(F)(F)F)[n-]c1C#N prediction: -4.809kcal/mol
2. cation: COCC[N+]1(C)CCCCC1 anion: FC(F)(F)C(F)(F)[P-](F)(F)(F)(C(F)(F)C(F)(F)F)C(F)(F)C(F)(F)F prediction: -4.732kcal/mol
3. cation: CCCCCCCC[N+](CC)(CC)CC anion: O=S(=O)(F)[N-]S(=O)(=O)F prediction: -4.732kcal/mol
4. cation: CCCCCCCC[N+]12CCC(CC1)CC2 anion: O=S(=O)([N-]S(=O)(=O)C(F)(F)F)C(F)(F)F prediction: -4.689kcal/mol
5. cation: CCCCCC[n+]1ccn(C)c1 anion: FC(F)(F)C(F)(F)[P-](F)(F)(F)(C(F)(F)C(F)(F)F)C(F)(F)C(F)(F)F prediction: -4.672kcal/mol

**Final Answer:**
Here are ionic liquid recommendations for maximizing toluene solubility from air into the IL phase (i.e., most negative solvation free energy). SMILES are provided for unambiguous identification, and lists are preserved as returned by the optimization tool.

PubChem search results Top5:
1) Cation: CCCCCCCCCC[P+](CCCCCCCC)(CCCCCCCC)CCCCCCCC | Anion: N#Cc1nc[n-]c1C#N | Predicted ΔG_solv (air → IL): -5.041 kcal/mol
(Omitted here for brevity)

Original dataset Top5:
1) Cation: CCCCCCCCCC[P+](CCCCCCCC)(CCCCCCCC)CCCCCCCC | Anion: N#Cc1nc[n-]c1C#N | Predicted ΔG_solv (air → IL): -5.041 kcal/mol
(Omitted here for brevity)

Solvation dataset Top5:
1) Cation: CCCC[n+]1cccc(C)c1 | Anion: N#Cc1nc(C(F)(F)F)[n-]c1C#N | Predicted ΔG_solv (air → IL): -4.809 kcal/mol
(Omitted here for brevity)

**Fig. S11. Example demonstration of AIonopedia executing an IL screening task.** AIonopedia searches for promising IL candidates for gas-phase toluene absorption.

**(Prompt):** I have an ionic liquid tributylethylphosphonium diethylphosphate (P4442 DEP) with the SMILES CCCC[P+](CC)(CCCC)CCCC.CCOP(=O)([O-])OCC.

Please help me increase its ammonia adsorption exclusively by modifying or replacing the cation, without modifying the anion. No web searching or lookup is needed. Do not use PubChem or any lookup tool for ammonia; use solute_SMILES='N' directly. Before choosing any cation-modification strategy, proposing any candidate, or calling any tool, you must explicitly write a reasoning preamble in text. The preamble must include: the solute interaction profile, including polarity, Bronsted/Lewis acidity or basicity, hydrogen-bond donor and/or acceptor behavior, hydrophilicity/hydrophobicity, and target complementary interactions; the chosen cation-modification strategy or strategies; the functional-group list if side-chain functionalization is used; and a candidate-planning table. Only after this preamble is written may you call property_predictor.

You may use either or both of the following common cation-modification strategies; they are not sequential steps.

1. One strategy is replacing the ion family with a common structurally related ion family, for example a piperidinium cation can be replaced by a morpholinium analogue such as CC[N+]1(C)CCCCC1 -> CC[N+]1(C)CCOCC1. Avoid replacing the cation with a structurally distant family, for example piperidinium and imidazolium should usually not be treated as close analogues.

2. Another strategy is functionalizing the cation side chains. For example, CC[n+]1ccn(C)c1 can be modified to C[n+]1ccn(CCO)c1 or C[n+]1ccn(CCC#N)c1. Using the solute interaction profile above, explicitly write a functional-group list before proposing any modified SMILES: list 8-10 common, simple, single functional groups matching the solute type, prioritizing the most broadly used organic groups over highly specialized, multifunctional, or overly strong-binding groups. For each group, briefly explain its interaction rationale by considering Bronsted/Lewis acidity or basicity, hydrogen-bond donor and/or acceptor behavior, polarity, and hydrophilicity/hydrophobicity together. After writing the list, select side-chain candidates only from it, using a diverse subset of interaction drivers when applicable rather than several groups with the same rationale. Do not exclude common acidic or basic functional groups solely because acid-base interactions or proton transfer may be possible. The substitution must be single substitution rather than multiple substitution. If the cation has multiple side chains, modify the most structurally distinctive unmodified carbon chain. During side-chain modification, usually preserve the original alkyl spacer length, but count any carbon atoms in the added functional group separately; for example, adding C#N to an ethyl spacer gives CC-C#N (CCC#N), not CC#N.

All examples above are illustrative only, not templates to copy. Do not default to the same groups or structures; choose modifications by reasoning from the specific solute and cation. Use several chemically reasonable cation candidates, calculate ammonia solvation free energy for each with the fixed DEP anion, compare them against the original cation, and choose the best improvement. You must arrive at the final answer within 10 calculation rounds.

**Thought:** I will first provide the required reasoning preamble to ground the cation-design choices in ammonia's interaction profile and then begin calculations. After the preamble, I will compute the baseline ammonia solvation free energy for the original IL to set a reference, followed by calculations on a diverse set of cation modifications that target complementary interactions with NH3 while keeping the DEP anion fixed.

Reasoning preamble

1) Solute interaction profile (NH3)
- Polarity: Highly polar, permanent dipole (~1.47 D), small, compact.
- Brønsted/Lewis character: Strong Brønsted base (pKb ~4.7) and good Lewis base (lone pair on N).
- Hydrogen-bonding: H-bond acceptor (strong, via N lone pair); H-bond donor (weak–moderate, via N–H).
- Hydrophilicity/hydrophobicity: Strongly hydrophilic; prefers polar, H-bonding environments.
- Target complementary interactions to enhance adsorption:
  - Provide H-bond donors on the cation side chain to bind NH3 as an H-bond acceptor.
  - Provide strong H-bond acceptors (carbonyl, nitrile, ether) to accept H from NH3.
  - Increase local polarity near the cation to better solvate a small polar base.
  - Maintain reasonable sterics so NH3 can access the interaction site near the charge.

2) Chosen cation-modification strategies (non-sequential)
- Strategy A: Replace the ion family with a closely related family while preserving overall topology (e.g., quaternary phosphonium → quaternary ammonium with the same alkyl set) to increase polarity and reduce hydrophobicity without introducing distant ring systems.
- Strategy B: Single functional-group substitution on the most structurally distinctive side chain (the ethyl arm of P4442+), preserving its two-carbon spacer and appending a simple, common functional group that complements NH3's basicity/H-bonding.

3) Functional-group list for side-chain functionalization (8–10 simple, common groups)
- Hydroxyl (–OH): Polar; H-bond donor and acceptor. Donor H can bind NH3 as acceptor; acceptor O can receive H from NH3. Hydrophilic.
- Amide (–C(=O)NH2): Strong H-bond acceptor (C=O) and moderate donor (NH). Polar, hydrophilic; dual-mode H-bonding with NH3.
- Nitrile (–C≡N): Strong H-bond acceptor via N; polar and compact; enhances dipole interactions with NH3's donor H.
- Carboxylic acid (–C(=O)OH): Strong donor and acceptor; very polar and hydrophilic; can H-bond strongly with NH3 and engage acid–base interactions.
- Ether (–O–): Moderate H-bond acceptor; increases polarity and flexibility; can accept H from NH3.
- Primary amine (–NH2): Donor and acceptor; basic; can form strong H-bonds and acid–base pairs with NH3.
- Ester (–C(=O)OR): H-bond acceptor (carbonyl); polar; tunable hydrophilicity; softer interaction than acids/amides.
- Sulfone (–SO2R): Very strong H-bond acceptor, highly polar; strengthens binding to NH3's donor H.

4) Candidate-planning table (fixed anion: diethyl phosphate, CCOP(=O)([O-])OCC; modify the ethyl arm on the cation; preserve 2-carbon spacer; single substitution)
- Baseline (reference): P4442+ cation unchanged. Purpose: establish reference solvation for NH3.
- A1 (Strategy B, hydroxyl): 2-hydroxyethyl substituent on the ethyl arm. Rationale: add strong donor/acceptor near charge for dual H-bonding with NH3.
- A2 (Strategy B, amide): 2-acetamido tail (…–CH2–CH2–C(=O)NH2). Rationale: strong acceptor (C=O) plus donor (NH) near charge.
- A3 (Strategy B, nitrile): 2-cyanoethyl tail (…–CH2–CH2–C≡N). Rationale: compact, strong acceptor, high polarity.
- A4 (Strategy B, carboxylic acid): 2-carboxyethyl tail (…–CH2–CH2–C(=O)OH). Rationale: very polar, donor/acceptor; strong specific interactions with NH3.
- A5 (Strategy B, ether): 2-methoxyethyl tail (…–CH2–CH2–O–CH3). Rationale: moderate acceptor; increases local polarity with minimal sterics.
- A6 (Strategy B, primary amine): 2-aminoethyl tail (…–CH2–CH2–NH2). Rationale: donor/acceptor; basic; can form strong H-bonds with NH3.
- B1 (Strategy A, family replacement): Quaternary ammonium analogue retaining tri-n-butyl + ethyl. Rationale: related family with greater polarity and potentially better NH3 solvation while avoiding distant ring systems.

I will now compute the baseline solvation free energy for NH3 in the original IL (P4442+ | DEP) to benchmark improvements.
**ToolCall:** property_predictor({'property_string': 'solvation', 'cation_SMILES': 'CCCC[P+](CC)(CCCC)CCCC', 'anion_SMILES': 'CCOP(=O)([O-])OCC', 'solute_SMILES': 'N', 'solvent_SMILES': '', 'Temp': '298'})
**Observation[property_predictor]:** -4.4575 kcal/mol
**(Omitted here for brevity)**

**Final Answer:**
Reasoning preamble

1) Solute interaction profile (ammonia, solute_SMILES='N')
- Polarity: Highly polar, small, with a sizable permanent dipole.
- Brønsted/Lewis character: Strong Brønsted base and good Lewis base (lone pair on N).
- Hydrogen-bonding: Strong H-bond acceptor via the N lone pair; weak-to-moderate H-bond donor via N–H.
- Hydrophilicity/hydrophobicity: Strongly hydrophilic; benefits from polar, H-bonding environments.
- Target complementary interactions at the cation:
  - Add H-bond donors to bind NH3 as an H-bond acceptor.
  - Add strong H-bond acceptors (e.g., carbonyl, sulfone, ether) to accept H from NH3.
  - Increase local polarity near the cation while keeping sterics open for NH3 access.

2) Chosen cation-modification strategies (not sequential)
- Strategy A (family replacement): Replace quaternary phosphonium with the closely related quaternary ammonium analogue (same side chains) to slightly increase polarity without a large topological change.
- Strategy B (single side-chain functionalization): Modify only the most distinctive short arm (the ethyl substituent) on P4442+, preserving a two-carbon spacer (–CH2–CH2–) and appending a single, simple functional group to drive complementary H-bonding/polar interactions with NH3.

3) Functional-group list for side-chain functionalization (simple, common, single groups; diverse interaction drivers)
- Hydroxyl (–OH): Polar; H-bond donor/acceptor; increases hydrophilicity and enables dual H-bonding with NH3.
- Amide (–C(=O)NH2): Strong acceptor (C=O) plus donor (NH); highly polar/hydrophilic; dual-mode H-bonding with NH3.
- Nitrile (–C≡N): Strong acceptor (nitrile N); compact/polar; enhances dipolar and H-bonding to NH3's donor H.
- Carboxylic acid (–C(=O)OH): Strong donor/acceptor; very polar; can form strong specific H-bonds and acid–base interactions with NH3.
- Ether (–O–): Moderate acceptor; increases polarity/flexibility; can accept H from NH3.
- Primary amine (–NH2): Donor/acceptor; basic; can form strong H-bonds and acid–base pairs with NH3.
- Ester (–C(=O)OR): Carbonyl acceptor; polar; softer interaction than acids/amides.
- Sulfone (–SO2R): Very strong H-bond acceptor; highly polar; strengthens binding to NH3's donor H.

4) Candidate-planning table (fixed anion: diethyl phosphate, CCOP(=O)([O-])OCC; modify only the ethyl arm; preserve a two-carbon spacer; single substitution)
- Baseline: P4442+ (tributylethylphosphonium), cation SMILES CCCC[P+](CC)(CCCC)CCCC.
- A1 (hydroxyl): 2-hydroxyethyl arm → CCCC[P+](CCO)(CCCC)CCCC.
- A2 (amide): 2-acylamido arm (–CH2–CH2–C(=O)NH2) → CCCC[P+](CCC(=O)N)(CCCC)CCCC.
- A3 (nitrile): 2-cyanoethyl arm → CCCC[P+](CCC#N)(CCCC)CCCC.
- A4 (carboxylic acid): 2-carboxyethyl arm → CCCC[P+](CCC(=O)O)(CCCC)CCCC.
- A5 (ether): 2-methoxyethyl arm → CCCC[P+](CCOC)(CCCC)CCCC.
- A6 (primary amine): 2-aminoethyl arm → CCCC[P+](CCN)(CCCC)CCCC.
- A7 (sulfone): 2-(methylsulfonyl)ethyl arm → CCCC[P+](CCS(=O)(=O)C)(CCCC)CCCC.
- B1 (family replacement): Quaternary ammonium analogue (tri-n-butyl + ethyl) → CCCC[N+](CC)(CCCC)CCCC.

Computed ammonia solvation free energies (direction: gas → IL; more negative = stronger adsorption), all with fixed DEP anion CCOP(=O)([O-])OCC at 298 K
- Baseline P4442+ | DEP: -4.4575 kcal/mol
- A1 (hydroxyl): -4.3717 kcal/mol
- A2 (amide): -4.5433 kcal/mol ← best improvement
- A3 (nitrile): -4.0757 kcal/mol
- A4 (carboxylic acid): -4.5004 kcal/mol
- A5 (ether): -4.3545 kcal/mol
- A6 (primary amine): -4.4318 kcal/mol
- A7 (sulfone): -4.3374 kcal/mol
- B1 (ammonium family): -4.3159 kcal/mol

Recommendation
- Best-performing cation modification: A2 amide side chain (single substitution on the ethyl arm, preserving a two-carbon spacer), cation SMILES CCCC[P+](CCC(=O)N)(CCCC)CCCC.
- Predicted ammonia solvation free energy with DEP: -4.5433 kcal/mol vs baseline -4.4575 kcal/mol (improvement of ~0.086 kcal/mol, i.e., stronger adsorption).
- If a second option is needed, A4 (carboxylic acid) also improves adsorption to -4.5004 kcal/mol but is less effective than A2.

Final recommended IL (keeping the anion unchanged):
- Cation: CCCC[P+](CCC(=O)N)(CCCC)CCCC
- Anion: CCOP(=O)([O-])OCC
- Predicted ammonia solvation free energy (gas → IL): -4.5433 kcal/mol

**Fig. S12. Example demonstration of AIonopedia executing an IL structure-modification task for ammonia absorption.** Starting from $[P_{4442}]^+[DEP]^-$, AIonopedia attempts to improve the ammonia absorption capacity through cation modification. Guided by a preamble specifying the desired IL characteristics, the agent performs multi-step reasoning and identifies the target structure $[P_{4442COOH}]^+[DEP]^-$ within ten modification rounds.

**(Prompt):** I have an ionic liquid trioctyldecylphosphonium 2-cyanopyrrolide (P88810 2-CNPyr) with the SMILES CCCCCCCCCC[P+](CCCCCCCC)(CCCCCCCC)CCCCCCCC.N#Cc1ccc[n-]1.

Please help me increase its toluene adsorption exclusively by modifying or replacing the cation, without modifying the anion. No web searching or lookup is needed. Do not use PubChem or any lookup tool for toluene; use solute_SMILES='Cc1ccccc1' directly. Before choosing any cation-modification strategy, proposing any candidate, or calling any tool, you must explicitly write a reasoning preamble in text. The preamble must include: the solute interaction profile, including polarity, Bronsted/Lewis acidity or basicity, hydrogen-bond donor and/or acceptor behavior, hydrophilicity/hydrophobicity, and target complementary interactions; the chosen cation-modification strategy or strategies; the functional-group list if side-chain functionalization is used; and a candidate-planning table. Only after this preamble is written may you call property_predictor.

You may use either or both of the following common cation-modification strategies; they are not sequential steps.

1. One strategy is replacing the ion family with a common structurally related ion family, for example a piperidinium cation can be replaced by a morpholinium analogue such as CC[N+]1(C)CCCCC1 -> CC[N+]1(C)CCOCC1. Avoid replacing the cation with a structurally distant family, for example piperidinium and imidazolium should usually not be treated as close analogues.

2. Another strategy is functionalizing the cation side chains. For example, CC[n+]1ccn(C)c1 can be modified to C[n+]1ccn(CCO)c1 or C[n+]1ccn(CCC#N)c1. Using the solute interaction profile above, explicitly write a functional-group list before proposing any modified SMILES: list 8-10 common, simple, single functional groups matching the solute type, prioritizing the most broadly used organic groups over highly specialized, multifunctional, or overly strong-binding groups. For each group, briefly explain its interaction rationale by considering Bronsted/Lewis acidity or basicity, hydrogen-bond donor and/or acceptor behavior, polarity, and hydrophilicity/hydrophobicity together. After writing the list, select side-chain candidates only from it, using a diverse subset of interaction drivers when applicable rather than several groups with the same rationale. Do not exclude common acidic or basic functional groups solely because acid-base interactions or proton transfer may be possible. The substitution must be single substitution rather than multiple substitution. If the cation has multiple side chains, modify the most structurally distinctive unmodified carbon chain. During side-chain modification, usually preserve the original alkyl spacer length, but count any carbon atoms in the added functional group separately; for example, adding C#N to an ethyl spacer gives CC-C#N (CCC#N), not CC#N.

All examples above are illustrative only, not templates to copy. Do not default to the same groups or structures; choose modifications by reasoning from the specific solute and cation. Use several chemically reasonable cation candidates, calculate toluene solvation free energy for each with the fixed 2-CNPyr anion, compare them against the original cation, and choose the best improvement. You must arrive at the final answer within 10 calculation rounds.

**Thought:** I will first craft the required preamble before making any calculations. I need to reason from the solute's interaction profile to choose cation-modification strategies and a focused set of side-chain functional groups, then plan several cation candidates (including structurally related family replacements and single side-chain substitutions on the decyl arm) to test with the fixed 2-CNPyr anion. After that, I will compute toluene solvation free energies for the baseline and each candidate and compare them to select the best improvement.

Preamble

1) Solute interaction profile (toluene; SMILES: Cc1ccccc1)
- Polarity: Very low overall polarity (small dipole), large quadrupole associated with the aromatic ring. Highly polarizable.
- Bronsted/Lewis acidity/basicity: Not a Brønsted acid or base; very weak Lewis base via the π cloud; no meaningful Lewis acidity.
- Hydrogen-bonding: Neither a hydrogen-bond donor nor a good acceptor.
- Hydrophilicity/hydrophobicity: Strongly hydrophobic; prefers nonpolar, polarizable environments.
- Target complementary interactions in the IL cation region: Maximize dispersion and π–π/π–π(T-shape) contacts, provide soft/polarizable local environment (S > O; heavy atoms), allow cation–π or induced-dipole interactions without introducing overly strong specific H-bond sites that would exclude aromatics, and maintain/highlight hydrophobic domains to host toluene.

2) Chosen cation-modification strategies
- Strategy A (ion-family replacement with a closely related family): Replace tetraalkylphosphonium [P88810]+ with structurally related onium cations of the same alkyl pattern:
  - Quaternary ammonium [N88810]+ to probe slightly "harder," less polarizable headgroup while keeping hydrophobic shell.
  - Trialkylsulfonium [S88810]+ to probe a softer, more polarizable headgroup that can enhance cation–π/dispersion with aromatics.
- Strategy B (single side-chain functionalization on the most distinctive chain): Modify only the decyl arm (the most distinctive among 10/8/8/8) at its ω-terminus with one functional group, preserving the 10-carbon spacer and counting added-group carbons separately. Aim to introduce π-surface, softness/polarizability, or controlled free volume while keeping the cation hydrophobic.

3) Functional-group list for side-chain functionalization (8–10 simple, common groups, with rationale)
- Phenyl (–CH2–Ph): Hydrophobic, highly polarizable π-surface enabling π–π/T-shape interactions with toluene; no HBD/HBA; complements aromatic solute via dispersion/π-stacking.
- Cyclohexyl (–CH2–cHex): Bulky, hydrophobic, high dispersion surface without π; can provide shape complement and CH···π contacts with toluene's ring.
- tert-Butyl (–CH2–C(CH3)3): Very hydrophobic, creates free volume and strong dispersion; can lower cohesive polarity locally and favor uptake of bulky aromatics.
- Nitrile (–CH2–C≡N): Small, π-accepting, weak HBA; increases polarizability and engages quadrupole/induced-dipole interactions while retaining compactness.
- Ether (–CH2–O–CH3): Weak HBA, moderate polarity; can tune cohesion and flexibility, enabling induced-dipole/dispersion with toluene without strong H-bonding.
- Thioether (–CH2–S–CH3): Softer, more polarizable than ether; hydrophobic with weak basicity; enhances dispersion/induction with aromatics.
- Fluoroalkyl (–CH2–CF3): Hydrophobic, highly electronegative shell with moderate polarizability; tunes surface energy and can sometimes improve aromatic partitioning.
- Halogen (–CH2–Cl): Increases polarizability and hydrophobicity; potential weak σ-hole interactions; usually mild effect but common and simple.
- Hydroxyl (–CH2–OH): HBD/HBA; increases polarity and cohesion—often reduces aromatic uptake, but included for breadth and tunability in nanosegregation contexts.
- Isopropyl (–CH(CH3)2): Hydrophobic branching increases dispersion and free volume relative to linear chain without adding strong polarity.

4) Candidate-planning table (fixed anion: 2-CNPyr, SMILES N#Cc1ccc[n-]1; solute: toluene, SMILES Cc1ccccc1)
- Baseline: [P88810]+ (trioctyldecylphosphonium), cation SMILES: CCCCCCCCCC[P+](CCCCCCCC)(CCCCCCCC)CCCCCCCC. Purpose: reference solvation free energy.
- Family replacement 1: [N88810]+ quaternary ammonium; cation SMILES: CCCCCCCCCC[N+](CCCCCCCC)(CCCCCCCC)CCCCCCCC. Rationale: test “harder,” less polarizable headgroup.
- Family replacement 2: [S88810]+ trialkylsulfonium; cation SMILES: CCCCCCCCCC[S+](CCCCCCCC)(CCCCCCCC)CCCCCCCC. Rationale: test “softer,” more polarizable headgroup for aromatics.
- Side-chain FG 1 (π-surface): ω-benzyl on decyl arm; cation SMILES: c1ccccc1CCCCCCCCCCC[P+](CCCCCCCC)(CCCCCCCC)CCCCCCCC. Rationale: add π–π/dispersion site for toluene.
- Side-chain FG 2 (softness): ω-(methylthio) on decyl arm; cation SMILES: CSCCCCCCCCCCC[P+](CCCCCCCC)(CCCCCCCC)CCCCCCCC. Rationale: increase softness/polarizability with minimal polarity.
- Side-chain FG 3 (quadrupole/compact): ω-cyano on decyl arm; cation SMILES: CC#NCCCCCCCCCC[P+](CCCCCCCC)(CCCCCCCC)CCCCCCCC. Rationale: small π-acceptor to engage induced-dipole/quadrupole.
- Side-chain FG 4 (free-volume/dispersion): ω-tert-butyl on decyl arm; cation SMILES: CC(C)(C)CCCCCCCCCC[P+](CCCCCCCC)(CCCCCCCC)CCCCCCCC. Rationale: bulky hydrophobe to create insertion pockets for toluene.

Plan: Compute solvation free energies for toluene in the baseline IL and each candidate IL; more negative values indicate stronger toluene solvation/adsorption from gas phase. Compare and select the best improvement while keeping the anion unchanged.
**ToolCall:** property_predictor({'property_string': 'solvation', 'cation_SMILES': 'CCCCCCCCCC[P+](CCCCCCCC)(CCCCCCCC)CCCCCCCC', 'anion_SMILES': 'N#Cc1ccc[n-]1', 'solute_SMILES': 'Cc1ccccc1', 'solvent_SMILES': '', 'Temp': '298'})
**Observation[property_predictor]:** -4.9723 kcal/mol
Thought: Compute toluene solvation free energy for the first cation-family replacement: quaternary ammonium [N88810]+ with the fixed 2-CNPyr anion, to compare against baseline and assess whether a “harder” less polarizable headgroup reduces or improves toluene uptake. I will keep solute_SMILES='Cc1ccccc1' as instructed.
**ToolCall:** property_predictor({'property_string': 'solvation', 'cation_SMILES': 'CCCCCCCCCC[N+](CCCCCCCC)(CCCCCCCC)CCCCCCCC', 'anion_SMILES': 'N#Cc1ccc[n-]1', 'solute_SMILES': 'Cc1ccccc1', 'solvent_SMILES': '', 'Temp': '298'})
**Observation[property_predictor]:** -5.0152 kcal/mol
**(Omitted here for brevity)**

**Final Answer:**
Computed objective: solvation free energy for moving toluene (solute_SMILES='Cc1ccccc1') from gas phase to the ionic liquid at 298 K. More negative values indicate stronger toluene affinity/adsorption.

Fixed anion: 2-cyanopyrrolide (anion_SMILES='N#Cc1ccc[n-]1').

Results
- Baseline cation [P88810]+ (CCCCCCCCCC[P+](CCCCCCCC)(CCCCCCCC)CCCCCCCC): −4.9723 kcal/mol
- Family replacement [N88810]+ (CCCCCCCCCC[N+](CCCCCCCC)(CCCCCCCC)CCCCCCCC): −5.0152 kcal/mol ← best (Δ = −0.0429 vs baseline; improved adsorption)
- Family replacement [S88810]+ (CCCCCCCCCC[S+](CCCCCCCC)(CCCCCCCC)CCCCCCCC): −4.9637 kcal/mol (slightly worse than baseline)
- Side-chain ω-benzyl on decyl arm (two placements tried):
  - [P+](CCCCCCCC)(CCCCCCCC)(CCCCCCCC)CCCCCCCCCCc1ccccc1: −4.7407 kcal/mol
  - [P+](CCCCCCCC)(CCCCCCCC)(CCCCCCCC)CCCCCCCCCCCc1ccccc1: −4.7235 kcal/mol
- Side-chain ω-(methylthio) on decyl arm: [P+](CCCCCCCC)(CCCCCCCC)(CCCCCCCC)CCCCCCCCCCSC: −4.8951 kcal/mol
- Side-chain ω-cyano on decyl arm:
  - [P+](CCCCCCCC)(CCCCCCCC)(CCCCCCCC)CCCCCCCCCCCC#N: −4.8093 kcal/mol
  - [P+](CCCCCCCC)(CCCCCCCC)(CCCCCCCC)CCCCCCCCCCC#N: −4.8265 kcal/mol
- Side-chain ω-tert-butyl on decyl arm:
  - [P+](CCCCCCCC)(CCCCCCCC)(CCCCCCCC)CCCCCCCCCCCC(C)(C)C: −4.8865 kcal/mol
  - [P+](CCCCCCCC)(CCCCCCCC)(CCCCCCCC)CCCCCCCCCCC(C)(C)C: −4.9123 kcal/mol

Recommendation
- To increase toluene adsorption without modifying the anion, replace the cation with the closely related quaternary ammonium analogue:
  - Recommended IL: [N88810][2-CNPyr]
  - Cation SMILES: CCCCCCCCCC[N+](CCCCCCCC)(CCCCCCCC)CCCCCCCC
  - Achieved improvement: solvation free energy = −5.0152 kcal/mol (more negative than the baseline −4.9723 kcal/mol).
- Within the single side-chain modifications tested on the decyl arm (benzyl, methylthio, cyano, tert-butyl), none outperformed the original phosphonium system for toluene; avoid these specific substitutions if maximizing toluene uptake is the goal under this model.

**Fig. S13. Example demonstration of AIonopedia executing an IL structure-modification task for toluene absorption.** Starting from $[P_{88810}]^{+}$[2-CNPyr]$^{-}$, AIonopedia attempts to improve the toluene absorption capacity through cation modification. Guided by a preamble specifying the desired IL characteristics, the agent performs multi-step reasoning and identifies the target structure $[N_{88810}]^{+}$[2-CNPyr]$^{-}$ within ten modification rounds.

**(Prompt):** I have an ionic liquid 1-ethyl-3-methylimidazolium thiocyanate (EMIM SCN) with the SMILES CC[n+]1ccn(C)c1.N#C[S-].

Please help me increase its carbon dioxide adsorption exclusively by replacing the anion, without modifying the cation. The solute is carbon dioxide with solute_SMILES='O=C=O'. No web searching is needed. Use chemical reasoning to propose replacement anions first. Before using PubChem, any lookup tool, property_predictor, or any other tool, you must first explicitly output a reasoning preamble in text. The preamble must include the original anion's key structural motif, common structurally related anion families, a complete enumeration of common neighboring anions within each selected family, including variants with different substitution numbers or central atoms when applicable, and a candidate-planning table. Only after this preamble is written may you use PubChem or lookup tools, and only to obtain or verify SMILES for anions already listed in the candidate-planning table. Do not use lookup tools to discover candidate anions or adsorption-performance information. Only after the candidate table has anion SMILES may you call property_predictor.

When proposing replacement anions, avoid overly complex or highly specialized anions; prioritize common ionic-liquid anions. First identify the original anion's key structural motif and enumerate common structurally related anion families. For any selected family, list the common members together, including variants with different substitution numbers, coordination numbers, or central atoms when applicable; do not list only one common member and skip its close neighbors. After the enumeration, select candidates only from the listed anions, based on structural similarity, common ionic-liquid usage, and expected CO2 affinity.

The following examples are only meant to illustrate what structural relatedness can mean; they are not required candidates and should not be copied mechanically. Structural relatedness does not require the same number of substituents: for example, BF4- and PF6- can be considered related inorganic fluorinated anions, and triflate O=S(=O)([O-])C(F)(F)F and TFSI O=S(=O)([N-]S(=O)(=O)C(F)(F)F)C(F)(F)F can be considered related sulfonyl/fluorinated anions.

Before any property_predictor call, output a candidate-planning table with columns: anion family, structural relation to the original anion, substitution/coordination count if applicable, anion SMILES, and reason for selection. Every calculated anion must appear in this table first; do not calculate table-external candidates.

Use several chemically reasonable anion candidates, calculate carbon dioxide solvation free energy for each with the fixed EMIM cation, compare them against the original SCN anion, and choose the best improvement. You must arrive at the final answer within 10 calculation rounds.

**Thought:** I will first provide the required chemical-reasoning preamble and enumerate related anion families, then present a candidate-planning table (with SMILES left blank). After this preamble, I will use PubChem only to obtain/verify SMILES for the listed anions, and then proceed to CO2 solvation-energy calculations.

Reasoning preamble (no tools)

1) Original anion’s key structural motif (SCN−)
- Structural motif: linear pseudohalide N≡C–S− with ambidentate character (resonance between −S–C≡N and −N≡C–S−). Features a soft, polarizable chalcogen (S) and a π-accepting nitrile unit.
- CO2 interaction logic: CO2 is quadrupolar/Lewis acidic at carbon; polarizable soft sites (S) and π systems (−C≡N) can stabilize via dipole–quadrupole and Lewis acid–base contacts. IL nanostructure/viscosity also modulate uptake and transport.

2) Structurally related anion families and enumeration of common neighbors
A) Pseudohalides and close analogs (linear, π-rich, ambidentate; varying chalcogen/heteroatom)
- Cyanide: CN−
- Cyanate / isocyanate: OCN− / NCO−
- Thiocyanate / isothiocyanate: SCN− / NCS−
- Selenocyanate: SeCN−
- Tellurocyanate: TeCN−
- Azide: N3−

B) Cyano-rich “multinitrile” anions (increasing nitrile count and charge delocalization; varying central atoms)
- Dicyanamide: [N(CN)2]− (DCA−) — 2 nitriles
- Tricyanomethanide: [C(CN)3]− (TCM−) — 3 nitriles
- Tetracyanoborate: [B(CN)4]− — 4 nitriles; central atom B
- Tetracyanoaluminate: [Al(CN)4]− — 4 nitriles; central atom Al

C) Perfluoro(sulfonate/sulfonyl imide) and related sulfonate families (varying number of sulfonyls and perfluoroalkyl size)
- Sulfonates (1 sulfonyl): triflate CF3SO3− (OTf−); nonaflate C3F7SO3−; perfluorobutanesulfonate C4F9SO3−; tosylate p-CH3C6H4SO3− (aryl, non-fluorinated analog)
- Sulfonyl imides (2 sulfonyls): bis(trifluoromethanesulfonyl)imide [N(SO2CF3)2]− (TFSI−); bis(fluorosulfonyl)imide [N(SO2F)2]− (FSI−); bis(pentafluoroethanesulfonyl)imide [N(SO2C2F5)2]− (BETI−)

D) Inorganic fluorinated anions (compact, weakly coordinating; differing coordination numbers)
- Tetrafluoroborate: BF4− (tetra-coordination)
- Hexafluorophosphate: PF6− (octa-coordination)
- Hexafluoroarsenate: AsF6− (octa)
- Hexafluoroantimonate: SbF6− (octa)

E) Carboxylates (small/basic H-bond acceptors; varying alkyl/fluoro substitution)
- Formate: HCOO−
- Acetate: CH3COO−
- Propionate: C2H5COO−
- Trifluoroacetate: CF3COO−
- Lactate: CH3CH(OH)COO−
- Benzoate: C6H5COO−

3) Candidate selection (only from enumerated lists)
- Pseudohalide analog: SeCN− (heavier chalcogen analog of SCN−; higher polarizability)
- Cyano-rich series: DCA−, TCM−, B(CN)4− (2/3/4 nitriles; strong delocalization)
- Perfluoro(sulfonyl/sulfonate): TFSI−, FSI−, triflate (OTf−) (common IL anions; tune polarity/viscosity; CO2-philic tendencies)
- Carboxylate benchmark: acetate (OAc−) (for contrast)

4) Candidate-planning table (pre-SMILES; SMILES to be filled via PubChem next)
- Columns: anion family | structural relation to SCN− | substitution/coordination count | anion SMILES | reason for selection

- Pseudohalides
  - Selenocyanate (SeCN−) | Heavier chalcogen analog of SCN−; same linear pseudohalide motif | N/A | [to be filled] | Higher polarizability expected to strengthen CO2 interactions

- Cyano-rich multinitriles
  - Dicyanamide (DCA−) | Adds a second nitrile and delocalized anionic center | 2 nitriles | [to be filled] | Common IL anion; nitrile-rich; good CO2 affinity
  - Tricyanomethanide (TCM−) | Three nitriles with strong delocalization | 3 nitriles | [to be filled] | π-accepting, often high CO2 solubility
  - Tetracyanoborate (B(CN)4−) | Four nitriles on B center | 4 nitriles | [to be filled] | Very nitrile-rich; strong physical absorption

- Perfluoro(sulfonyl/sulfonate)
  - TFSI− | Sulfonyl-imide family; highly delocalized | 2 sulfonyls | [to be filled] | Very common IL anion; often good CO2 uptake and low viscosity
  - FSI− | Same family with −SO2F groups | 2 sulfonyls | [to be filled] | Lower viscosity; can favor CO2 transport/uptake
  - Triflate (OTf−) | Sulfonate analog with CF3 group | 1 sulfonyl | [to be filled] | Common IL anion; balances polarity/hydrophobicity

- Carboxylates
  - Acetate (OAc−) | Small/basic anion (different donor character) | N/A | [to be filled] | Benchmark; informative contrast

Next, I will use PubChem only to obtain/verify SMILES for these listed anions.
**ToolCall:** pubchem_searcher({'query': 'What is the smiles of selenocyanate anion?'})
**Observation[pubchem_searcher]:** N#C[Se-]
**(Omitted here for brevity)**

**Thought:** I have now obtained SMILES for all pre-selected anions using PubChem. Before any calculations, I will present the completed candidate-planning table with SMILES, then begin computing CO2 solvation free energies (gas → IL) for EMIM with each anion, starting with the original SCN− baseline.

Candidate-planning table:
- Pseudohalides
  - Selenocyanate (SeCN−) | Heavier chalcogen analog of SCN−; same linear pseudohalide motif | N/A | N#C[Se-] | Higher polarizability expected to strengthen CO2 interactions
- Cyano-rich multinitriles
  - Dicyanamide (DCA−) | Adds a second nitrile and delocalized anionic center | 2 nitriles | N#CN=C=[N-] | Common IL anion; nitrile-rich; good CO2 affinity
  - Tricyanomethanide (TCM−) | Three nitriles with strong delocalization | 3 nitriles | [C-](C#N)(C#N)C#N | π-accepting, often high CO2 solubility
  - Tetracyanoborate (B(CN)4−) | Four nitriles on B center | 4 nitriles | N#C[B-](C#N)(C#N)C#N | Very nitrile-rich; strong physical absorption
- Perfluoro(sulfonyl/sulfonate)
  - TFSI− | Sulfonyl-imide family; highly delocalized | 2 sulfonyls | O=S(=O)(C(F)(F)F)[N-]S(=O)(=O)C(F)(F)F | Very common IL anion; often good CO2 uptake and low viscosity
  - FSI− | Same family with −SO2F groups | 2 sulfonyls | O=S(=O)(F)[N-]S(=O)(=O)F | Lower viscosity; can favor CO2 transport/uptake
  - Triflate (OTf−) | Sulfonate analog with CF3 group | 1 sulfonyl | O=S(=O)([O-])C(F)(F)F | Common IL anion; balances polarity/hydrophobicity
- Carboxylates
  - Acetate (OAc−) | Small/basic anion (different donor character) | N/A | CC(=O)[O-] | Benchmark; informative contrast

I will now compute the baseline CO2 solvation free energy for EMIM SCN.
**ToolCall:** property_predictor({'property_string': 'solvation', 'cation_SMILES': 'CC[n+]1ccn(C)c1', 'anion_SMILES': 'N#C[S-]', 'solute_SMILES': 'O=C=O', 'solvent_SMILES': '', 'Temp': '298'})
**Observation[property_predictor]:** -0.5964 kcal/mol
**(Omitted here for brevity)**

**Final Answer:**
Reasoning preamble (no tools)
(Omitted here for brevity)

Solvation-energy calculations at 298 K (direction: gas → IL; more negative means stronger CO2 absorption):
- Baseline EMIM SCN: −0.5964 kcal/mol
- EMIM SeCN: −0.5620 kcal/mol (slightly weaker than SCN)
- EMIM DCA: −0.8709 kcal/mol (improved vs SCN)
- EMIM TCM: −1.3686 kcal/mol (substantial improvement)
- EMIM B(CN)4: −1.7204 kcal/mol (largest improvement among nitrile-rich)
- EMIM TFSI: −1.6346 kcal/mol (strong improvement)
- EMIM FSI: −1.3857 kcal/mol (strong improvement)
- EMIM OTf: −1.0597 kcal/mol (improvement)
- EMIM OAc: −0.6135 kcal/mol (similar to or slightly worse than SCN)

Recommendation:
- Best predicted improvement comes from replacing SCN− with tetracyanoborate B(CN)4−, giving a CO2 solvation free energy of −1.7204 kcal/mol (vs −0.5964 for SCN).
- Strong alternatives with excellent predicted performance are TFSI− (−1.6346 kcal/mol) and TCM− (−1.3686 kcal/mol); FSI− also improves (−1.3857 kcal/mol).
- If prioritizing maximum CO2 uptake: EMIM B(CN)4 is the top choice among tested anions. If balancing uptake and viscosity/availability: EMIM TFSI or EMIM FSI are strong, widely used options.

Fixed cation and solute used in all calculations:
- Cation (EMIM): CC[n+]1ccn(C)c1
- Solute (CO2): O=C=O
- Temperature: 298 K

**Fig. S14. Example demonstration of AIonopedia executing an IL structure-modification task for carbon dioxide absorption.** Starting from $[EMIM]^+[SCN]^-$, AIonopedia attempts to improve the carbon dioxide absorption capacity through anion substitution. Guided by a preamble specifying the desired IL characteristics, the agent performs multi-step reasoning and successfully identifies the target structure $[EMIM]^+[TCB]^-$ within ten modification rounds.

**(Prompt):** I have a structural image of an ionic liquid: ions.png, and a corresponding comparison plot of toluene adsorption capacities in several ionic liquids: graph.png. Please answer the following questions:

1. At what relative pressure p/p0 does this ionic liquid reach its maximum toluene adsorption capacity in the second figure?
2. What is the toluene adsorption capacity of this ionic liquid at this relative pressure in the second figure?
3. Convert this adsorption capacity at this relative pressure into units of mol/mol.
4. How many mL of this ionic liquid are needed to adsorb 1 mol of toluene at this relative pressure?

Use the available tools and codes as needed.
Do not use web search.

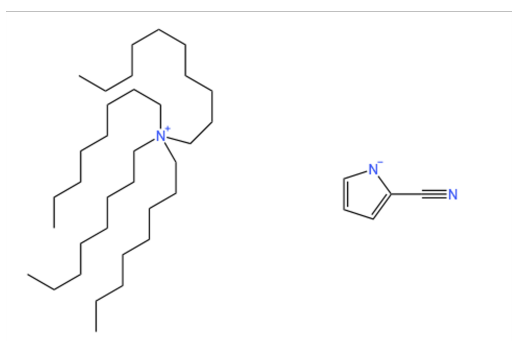

ions.png

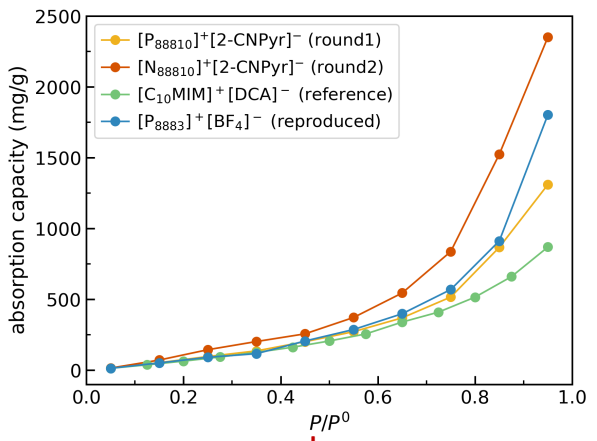

graph.png

**Thought:** I need the identities (SMILES) of the ions from the provided structural image so I can later compute molecular weights and use the property calculator for density. I will extract canonical SMILES from ions.png using the molecule processor.
**ToolCall:** molecule_processor({'molecule_input': 'path/ions.png', 'input_type': 'image'})
**Observation[molecule_processor]:** CCCCCCCCCC[N+](CCCCCCCC)(CCCCCCCC)CCCCCCCC.N#Cc1ccc[n-]1

**Thought:** I identified the cation and anion SMILES from ions.png. Next I will estimate the ionic liquid's density using the property_predictor, which requires the cation and anion SMILES. This will later allow me to convert mass of IL into volume for question 4.
**ToolCall:** property_predictor({'property_string': 'density', 'cation_SMILES': 'CCCCCCCCCC[N+](CCCCCCCC)(CCCCCCCC)CCCCCCCC', 'anion_SMILES': 'N#Cc1ccc[n-]1', 'solute_SMILES': '', 'solvent_SMILES': '', 'Temp': '298'})
**Observation[property_predictor]:** 0.8741 g/cm^3
**ToolCall:** code_processor

```
from rdkit import Chem
from rdkit.Chem import Descriptors

smiles_cation = 'CCCCCCCCCC[N+](CCCCCCCC)(CCCCCCCC)CCCCCCCC'
smiles_anion = 'N#Cc1ccc[n-]1'
smiles_toluene = 'Cc1ccccc1'

mol_cation = Chem.MolFromSmiles(smiles_cation)
mol_anion = Chem.MolFromSmiles(smiles_anion)
mol_tol = Chem.MolFromSmiles(smiles_toluene)

mw_cation = Descriptors.MolWt(mol_cation)
mw_anion = Descriptors.MolWt(mol_anion)
mw_tol = Descriptors.MolWt(mol_tol)

mw_il = mw_cation + mw_anion

print('MW cation:', mw_cation)
print('MW anion:', mw_anion)
print('MW IL:', mw_il)
print('MW toluene:', mw_tol)

# Use capacity 2400 mg/g -> 2.4 g/g
cap_mg_per_g = 2400
cap_g_per_g = cap_mg_per_g/1000.0

# mol toluene per mol IL
mol_per_mol = cap_g_per_g * mw_il / mw_tol
print('mol/mol at 2400 mg/g:', mol_per_mol)

# IL mass needed for 1 mol toluene
mass_tol = mw_tol # g per mol
mass_il_needed = mass_tol / cap_g_per_g
print('IL mass needed for 1 mol toluene (g):', mass_il_needed)

# Volume using density 0.8741 g/mL
rho = 0.8741
vol_ml = mass_il_needed / rho
print('Volume IL needed (mL):', vol_ml)
```

**Observation[code_processor]:** MW cation: 494.95700000000056
MW anion: 91.09299999999999
MW IL: 586.0500000000005
MW toluene: 92.14099999999999
mol/mol at 2400 mg/g: 15.264865803496829
IL mass needed for 1 mol toluene (g): 38.39208333333333
Volume IL needed (mL): 43.921843419898565

**Final Answer:**
1) The ionic liquid shown in ions.png corresponds to the [N88810]+[2-CNPyr]− curve in the plot. Its maximum toluene adsorption capacity occurs at the highest plotted relative pressure, p/p0 ≈ 0.95.

2) The adsorption capacity at p/p0 ≈ 0.95 for this IL is ≈ 2400 mg/g.

3) Converting 2400 mg/g to mol of toluene per mol of IL:
- Canonical SMILES from ions.png: cation CCCCCCCCCC[N+](CCCCCCCC)(CCCCCCCC)CCCCCCCC; anion N#Cc1ccc[n-]1.
- Molecular weights (RDKit): IL = 586.05 g/mol; toluene = 92.141 g/mol.
- 2400 mg/g = 2.4 g toluene per g IL → mol/mol = (2.4 g toluene/g IL) × (586.05 g IL/mol IL) / (92.141 g toluene/mol) ≈ 15.27 mol toluene per mol IL.

4) mL of IL needed to adsorb 1 mol toluene at this relative pressure:
- Mass of toluene to capture = 1 mol × 92.141 g/mol = 92.141 g.
- With 2.4 g toluene per g IL, required IL mass = 92.141/2.4 ≈ 38.39 g.
- IL density at 298 K from the calculator: 0.8741 g/mL.
- Volume = 38.39 g / 0.8741 g/mL ≈ 43.9 mL.

Answers:
1) p/p0 ≈ 0.95
2) ≈ 2400 mg/g
3) ≈ 15.3 mol/mol
4) ≈ 43.9 mL

**Fig. S15. Example demonstration of AIonopedia executing an image-text reasoning task.** Given a line-angle molecular structure image and the corresponding absorption-curve chart, AIonopedia answers a series of user questions by leveraging external tools, code execution, and the intrinsic multimodal reasoning capability of GPT-5.

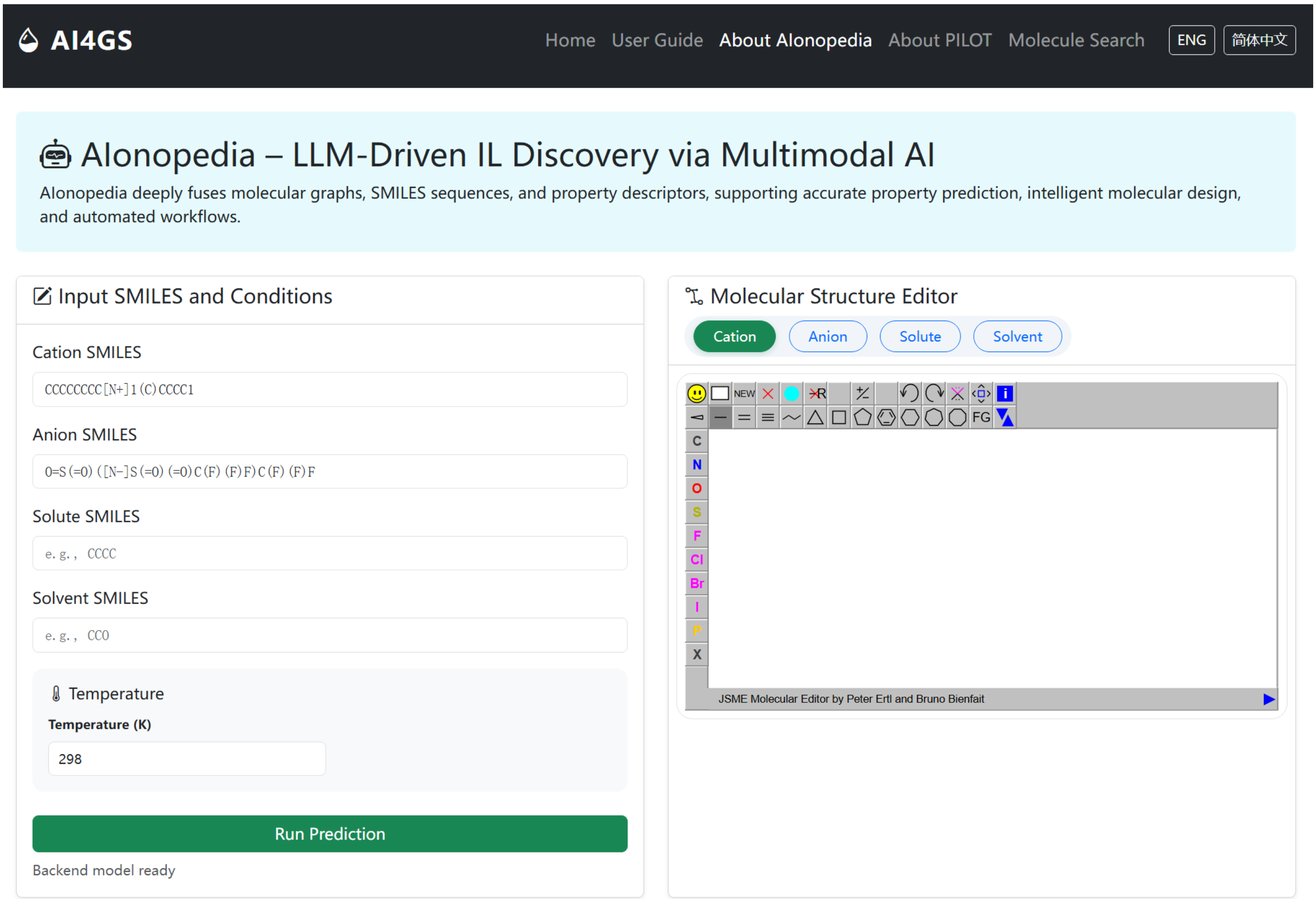

**Fig. S16. Web interface of the AIonopedia online service.** Users can input cation, anion, solute, and solvent structures using SMILES representations or the integrated molecular structure editor, specify the temperature, and perform property prediction through the deployed AIonopedia backend. The online service is freely accessible at https://ai4gs.suat-sz.edu.cn/zh.